\documentclass{article}
\usepackage{iclr2027_conference,times}
\usepackage{amsmath,amssymb}
\usepackage{booktabs,tabularx,array,multirow}
\usepackage{xcolor}
\usepackage{float}
\usepackage{wrapfig}
\usepackage{hyperref,url}
\usepackage{graphicx}
\hypersetup{
  hidelinks,
  pdftitle={HiSTOR: Structured Treatment-outcome Representation Learning for High-Dimensional Longitudinal Data},
  pdfauthor={Chuanqi Lao, Junzhou Huang, Qiao Liu}
}

\makeatletter
\newcounter{algorithm}
\renewcommand{\thealgorithm}{\arabic{algorithm}}
\newcommand{\fps@algorithm}{tbp}
\newcommand{\ftype@algorithm}{4}
\newcommand{\ext@algorithm}{loa}
\newcommand{\fnum@algorithm}{Algorithm~\thealgorithm}

\makeatother

\title{HiSTOR: Structured Treatment-outcome Representation Learning for High-Dimensional Longitudinal Data}
\author{%
Chuanqi Lao$^{1}$ \quad Junzhou Huang$^{1}$ \quad Qiao Liu$^{2,*}$\\
$^{1}$Department of Computer Science and Engineering, The University of Texas at Arlington\\
$^{2}$Department of Biostatistics, Yale School of Public Health, Yale University\\
$^{*}$Corresponding author: \texttt{qiao.liu@yale.edu}%
}
\iclrfinalcopy

\begin{document}
\maketitle
\fancyhead{}

\begin{abstract}
Estimating potential outcomes under alternative treatment sequences is essential for personalized decision-making, but remains challenging with high-dimensional covariates and time-varying confounding. In this study, we propose \textbf{HiSTOR} (\textbf{Hi}gh-dimensional \textbf{S}tructured \textbf{T}reatment-\textbf{O}utcome \textbf{R}epresentation), an end-to-end framework that integrates nonlinear dimension reduction with longitudinal counterfactual prediction. The central idea is to learn structured low-dimensional representations of both current patient states and evolving histories, according to their shared and distinct relationships with treatment assignment and outcomes. This extends latent-space covariate adjustment to longitudinal settings, jointly addressing covariate dimensionality and historical complexity. The resulting framework supports one-step and multi-step outcome prediction under specified treatment sequences. Experiments on CVSim and CancerSim demonstrate the superiority of HiSTOR across varying covariate dimensions, confounding strengths, and prediction horizons compared to state-of-the-art methods. To sum up, HiSTOR offers a unified approach to connecting high-dimensional covariate adjustment with longitudinal potential-outcome prediction. The code is available at \url{https://github.com/liuq-lab/HiSTOR}.
\end{abstract}

\section{Introduction}\label{sec:intro}

Personalized treatment planning requires predicting outcomes under alternative treatment sequences from longitudinal observations. This problem is fundamentally challenging because treatment decisions may depend on time-varying covariates that also affect future outcomes and may themselves be influenced by prior treatments. The difficulty is further amplified in modern biomedical applications, where each patient is characterized by high-dimensional and strongly correlated measurements accumulated over time~\citep{allam2021analyzing,bica2021real,battalio2021sense2stop,murray2016evaluating}. As the covariate dimension and observation history grow, effective counterfactual prediction therefore requires methods that can effectively compress patient information while retaining what is relevant to both treatment assignment and outcome prediction.

\begin{figure}[!htbp]
\centering
\includegraphics[width=\linewidth]{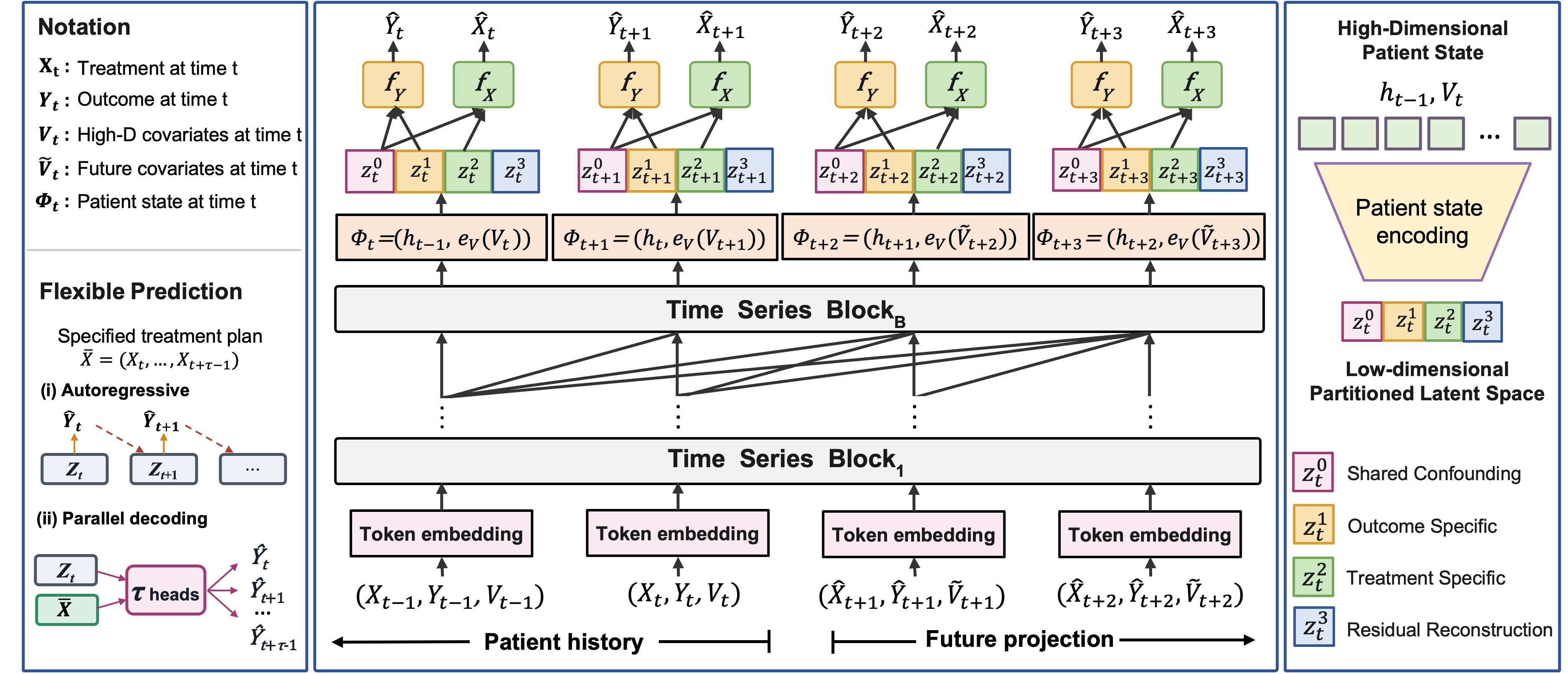}
\caption{Overview of HiSTOR. The model adopts a temporal Transformer backbone and combines a high-dimensional current patient state
with a learned representation of longitudinal history,
maps their fusion into a structured latent space,
and supports treatment-conditioned flexible one-step
and multi-step outcome prediction.}
\label{fig:architecture}
\end{figure}

Recent modern AI approaches to longitudinal counterfactual prediction address time-varying confounding through
reweighting, representation balancing, sequential outcome modeling, or recursive regression \citep{lim2018forecasting,bica2020estimating,melnychuk2022causal,li2021g,xiong2024g,hess2026igc}. These methods primarily focus on how longitudinal histories should be adjusted, balanced, or propagated over time. Less attention has been paid to how high-dimensional covariates and accumulated histories should be jointly compressed and represented for treatment-conditioned outcome prediction. 

To address the high-dimensional covariate and confounding challenges, we propose \textbf{HiSTOR} (\textbf{Hi}gh-dimensional \textbf{S}tructured \textbf{T}reatment-\textbf{O}utcome \textbf{R}epresentation),
an end-to-end framework for structured representation learning in high-dimensional longitudinal settings (Figure~\ref{fig:architecture}). The key idea of HiSTOR is to separately represent the current high-dimensional patient state and the preceding longitudinal history, and then organize their fusion into a task-structured low-dimensional latent space. This representation preserves information relevant to treatment assignment and outcome prediction. By explicitly separating current-state compression from historical representation learning, HiSTOR provides a unified representation
for longitudinal covariate adjustment and treatment-conditioned outcome prediction. We summarize our main contributions as follows:
\begin{itemize}
    \item
    We introduce a structured representation-learning framework
    for high-dimensional longitudinal causal inference
    that encodes the current patient state
    and preceding history by their roles in the treatment and outcome mechanisms.

    \item
    We extend structured latent covariate adjustment
    from cross-sectional to longitudinal settings
    and develop flexible prediction mechanisms
    for both one-step and multi-step
    treatment-conditioned potential outcomes.

    \item
    We demonstrate strong empirical performance of HiSTOR across varying confounding strengths,
    covariate dimensions, and prediction horizons.
    HiSTOR achieves the lowest mean
    one-step and five-horizon-average errors
    across all tested CancerSim confounding strengths,
    and the lowest five-horizon-average errors
    across all tested CVSim covariate dimensions.
\end{itemize}

\section{Related Work}\label{sec:related}

\textbf{Traditional longitudinal causal inference.} Static estimators learn potential outcomes from covariate-treatment-outcome observations \citep{johansson2016learning,alaa2018limits,wager2018estimation,yoon2018ganite,curth2021nonparametric}, whereas longitudinal methods condition on evolving patient histories. Classical approaches include marginal structural models and G-computation \citep{robins1986new,robins2000marginal,robins2008estimation}. These approaches provide the statistical foundation for longitudinal causal inference, but their direct application can become challenging with complex, high-dimensional patient histories.

\textbf{Modern AI for longitudinal counterfactual prediction.} 
Recent work integrates these causal principles with flexible neural models for learning treatment effects from longitudinal data. RMSN combines recurrent outcome models with stabilized inverse-probability weights \citep{lim2018forecasting}. CRN uses treatment-adversarial training to balance recurrent histories, and CT uses a multi-input Transformer with domain confusion \citep{bica2020estimating,melnychuk2022causal}. Causal CPC combines recurrent encoding, contrastive predictive coding, and representation balancing \citep{bouchattaoui2024causal}; CTD-NKO uses treatment-dependent neural Koopman dynamics and distributional weighting \citep{wang2025enhancing}. G-Net and G-Transformer estimate future-variable distributions for G-computation \citep{li2021g,xiong2024g}, while IGC-Net recursively regresses pseudo-outcomes without modeling complete future-covariate distributions \citep{hess2026igc}. These approaches advance complementary aspects of confounding adjustment, temporal representation learning, and treatment-conditioned forecasting.

\paragraph{Structured representation learning for causal inference.}
In cross-sectional settings, DR-CFR learns treatment-related, outcome-related, and shared representations \citep{hassanpour2019learning}, while TEDVAE separates instrumental, confounding, and outcome risk factors \citep{zhang2021treatment}. CausalEGM, CausalBGM, and BGM-IV combine structured latent modeling with high-dimensional covariate adjustment through encoding-generative or Bayesian formulations \citep{liu2024encoding,liu2026ai,luo2026bgm}. These methods do not model repeated treatments or evolving patient histories.

\paragraph{Positioning of HiSTOR.}
HiSTOR addresses this longitudinal prediction task with a focus on high-dimensional patient covariates. Its contribution is to couple explicit covariate compression with a latent-space partition designed for both prediction and reconstruction. The partition combines shared and task-specific predictive components with a dedicated residual reconstruction component, linking the organization of treatment-outcome information with information retention under compression. This design brings together structured covariate adjustment and longitudinal representation learning for patient histories containing many correlated covariates.


\section{Problem Formulation}\label{sec:problem}

Let $\mathcal D=\{\{V_{it},X_{it},Y_{it}\}_{t=1}^{T_i}\}_{i=1}^N$ contain independent patient trajectories. Here $V_{it}$ denotes the covariates available before treatment at time $t$, including baseline components and, when present, time-varying components; $X_{it}$ is the observed treatment; and $Y_{it}$ is the associated post-treatment outcome. Suppressing $i$, the observed history before treatment at time $t$ is
\begin{equation}
 \mathcal H_t=(X_{<t},Y_{<t},V_{\leq t}).
 \label{eq:history}
\end{equation}
Here $\mathcal H_t$ is the observed history, whereas $h_{t-1}$ is the representation of the preceding trajectory through time $t-1$. Appendix~\ref{app:notation} summarizes the notation and temporal alignment.

Following the longitudinal potential-outcomes framework \citep{robins1986new,robins2008estimation}, let $Y_t(x)$ be the outcome under treatment $x$ at time $t$ with prior treatments observed. For history $h$ and a specified sequence $\bar x=x_{t:t+\tau-1}$, the targets are
\begin{equation}
\left\{
\begin{aligned}
\mu_t(x,h)
&= \mathbb E\!\left[Y_t(x)\mid\mathcal H_t=h\right],\\
\mu_t^{(\tau)}(\bar x,h)
&= \mathbb E\!\left[Y_{t+\tau-1}(\bar x)\mid\mathcal H_t=h\right].
\end{aligned}
\right.
\label{eq:estimands}
\end{equation}
Thus $\tau=1$ reduces to the one-step target. At prediction time only $\mathcal H_t$ and the specified treatment sequence are available, while future outcomes and covariates are not.

Identification from observational data requires longitudinal consistency, positivity, and sequential ignorability \citep{robins2000marginal,robins2008estimation}. For $t\leq s\leq u$ and an intervention sequence $x_{s:u}$, we state these conditions as
\begin{equation}
\left\{
\begin{aligned}
&Y_u(X_{s:u})
= Y_u
&& \text{(Consistency)},\\
&\Pr(X_s=x_s\mid \mathcal H_s=h_s)
>0
&& \text{(Positivity)},\\
&X_s
\perp\!\!\!\perp Y_u(x_{s:u})\mid \mathcal H_s
&& \text{(Sequential ignorability)}.
\end{aligned}
\right.
\label{eq:identification}
\end{equation}
Positivity is required for intervention-relevant treatments at histories that can arise under the intervention, while sequential ignorability assumes no unmeasured treatment-outcome confounding conditional on $\mathcal H_s$. Under these assumptions, the targets in Equation~\ref{eq:estimands} are identified within the standard longitudinal potential-outcome framework.


\section{Methods}\label{sec:method}



\subsection{Covariate and History Encoding}\label{sec:twolevel}

At each decision time $t$, HiSTOR encodes the pre-treatment information in $\mathcal H_t$ through two complementary paths: a covariate encoder for the current covariates $V_t$,
and a causally masked temporal network for the preceding trajectory.

The covariate encoder $e_V$ maps the current covariates to a low-dimensional representation $c_t=e_V(V_t)$. The temporal network $\mathcal T$ summarizes the trajectory preceding treatment $X_t$. At sequence position $t$, it receives the treatment, outcome,
and covariate encoding from the previous time step,
which are embedded and summed into an input token. $\mathcal T$ is a stack of $B$ Transformer decoder blocks and produces the embedding $h_{t-1}=\mathcal T(X_{<t},Y_{<t},V_{<t})$. The attention masks ensure that $h_{t-1}$ uses only information available through time $t-1$.
The current covariate encoding $c_t$ and preceding trajectory representation $h_{t-1}$ are subsequently fused and mapped into the structured latent space.

\subsection{Structured Latent Representation}

Treatment assignment and outcomes may depend on overlapping but distinct aspects of the pre-treatment history. A single entangled representation does not distinguish such structure. To consider this, HiSTOR fuses the current covariate encoding and preceding trajectory representation, then maps their concatenation into a partitioned latent space:

\begin{equation}
 \Phi_t=[h_{t-1},c_t],\qquad
 Z_t=Q(\Phi_t)=[z_t^0,z_t^1,z_t^2,z_t^3].
 \label{eq:structured}
\end{equation}

$z_t^0$ is provided to both the treatment and outcome prediction heads. This encourages $z_t^0$ to retain information in the observed history that is predictive of both treatment assignment and outcomes, and is therefore most directly relevant to confounding adjustment.
In contrast, $z_t^1$ is used only for outcome prediction and $z_t^2$ only for treatment prediction, allowing outcome-specific and treatment-specific variation to be learned.
The remaining latent component $z_t^3$ is trained through residual reconstruction, providing additional capacity to preserve irrelevant covariate information.

\subsection{Learning the Structured Representation}\label{sec:structured}

The partition in equation~\ref{eq:structured} does not by itself determine the functional role of each latent component. HiSTOR induces this structure through task-specific supervision that controls which components are available to the treatment and outcome prediction heads. 

For a specified treatment $x_t$, the outcome and auxiliary treatment
predictions are
\begin{equation}
\widehat y_t(x_t)
=f_Y([z_t^0,z_t^1,x_t]),\qquad
\widehat{\boldsymbol{\pi}}_t
=f_X([z_t^0,z_t^2]),
\label{eq:heads}
\end{equation}
where $\widehat{\boldsymbol{\pi}}_t$ denotes the predicted treatment-assignment distribution. The treatment head is used only to supervise representation learning. The outcome head always conditions on the observed or externally specified treatment $x_t$, rather than on the auxiliary treatment prediction.

During training, the outcome head is optimized using the factual outcome loss $\mathcal L_Y$,
while the treatment head is optimized using the auxiliary treatment loss $\mathcal L_X$.
The residual component $z_t^3$ is trained through reconstruction.
Together, these objectives learn the structured representation end-to-end,
while the complete loss definitions and regularization terms are given in Section~\ref{sec:optimization}.

\subsection{Training Objectives }\label{sec:optimization}

HiSTOR jointly trains the current-state encoder, temporal network, structured mapping, prediction heads, and reconstruction modules.
The overall objective combines factual outcome prediction, auxiliary treatment prediction, reconstruction, and representation regularization:


\begin{equation}
\begin{aligned}
 \mathcal L_{\mathrm{total}}={}&
 \mathcal L_Y
 +\lambda_X\mathcal L_X
 +\lambda_{rec}\mathcal L_{\mathrm{rec}}
 +\lambda_{\mathrm{adv}}(e)
\mathcal L_{\mathrm{adv}}
+\lambda_{\mathrm{norm}}\mathcal L_{\mathrm{norm}}
+\lambda_{\mathrm{roll}}\mathcal L_{\mathrm{roll}}.
\end{aligned}
\label{eq:objective}
\end{equation}

Here, $\mathcal L_Y$ and
$\mathcal L_X$ are introduced in
Equation~\ref{eq:heads}.
The reconstruction objective $\mathcal L_{\mathrm{rec}}$ encourages the compressed representation to retain information about the observed patient state and provides supervision for the residual component $z_t^3$.
The distribution-matching loss $\mathcal L_{\mathrm{adv}}$ regularizes the learned current-state representation,
while $\mathcal L_{\mathrm{norm}}$ controls its scale. This combination of reconstruction and distribution matching
is also used in bidirectional generative modeling~\citep{liu2021density}.

To reduce train-prediction discrepancy, we optionally introduce a rollout loss $\mathcal L_{\mathrm{roll}}$ that recursively applies the one-step predictor over factual multi-step windows and propagates its predictions through subsequent histories.
Complete definitions of the reconstruction, distribution-matching,
rollout, and weighting objectives are provided in Appendix~\ref{app:architecture}.

\subsection{One-Step and Multi-Step Prediction}\label{sec:prediction}

For a specified treatment $x_t$, the one-step prediction is
\begin{equation}
 \widehat\mu_\theta^{(1)}(x_t,\mathcal H_t)
 =f_Y([z_t^0,z_t^1,x_t]),
 \label{eq:onestep}
\end{equation}
where $\theta$ denotes the fitted model parameters and $\mathcal H_t$ is the observed pre-treatment history. This is the prediction $\widehat y_t(x_t)$ defined in Equation~\ref{eq:heads}. Alternative treatments change $x_t$ while retaining the same pre-treatment information.

\paragraph{Autoregressive prediction.} For a treatment plan $\bar x=x_{t:t+\tau-1}$ of length $\tau$, we recursively predict
\begin{equation}
 \widehat y_s^{\mathrm{AR}}
 =\widehat\mu_\theta^{(1)}(x_s,\widehat{\mathcal H}_s),
 \qquad s=t,\ldots,t+\tau-1,
 \label{eq:rollout}
\end{equation}
where $\widehat{\mathcal H}_s$ combines the observed history up to the prediction origin,
the specified treatment sequence, and previously generated outcomes. For $\widetilde V_t$, static covariates are reused unchanged while time-varying covariate inputs are carried forward from their last observed values at the prediction origin. This autoregressive construction can accommodate arbitrary prediction horizons,
but its recursive dependence on model-generated outcomes allows prediction errors to propagate across future steps.


\paragraph{Parallel decoding.}
To avoid recursive outcome feedback, HiSTOR also supports direct multi-step prediction. Given the structured representation $Z_t$ at the prediction origin and a specified treatment plan $\bar x=x_{t:t+\tau-1}$, a decoder predicts the complete outcome trajectory in a single forward pass:
\begin{equation}
 (\widehat y_t^{\mathrm D},\ldots,\widehat y_{t+\tau-1}^{\mathrm D})
 =D_\tau(Z_t,\bar x).
 \label{eq:direct_prediction}
\end{equation}
The decoder uses a shared representation of the treatment plan together with horizon-specific outcome heads.
It is trained additionally on factual multi-step windows.
Thus, autoregressive and parallel decoding provide complementary strategies for extending the same structured representation to longitudinal treatment-conditioned prediction.

Appendix~\ref{app:forecast} specifies the dataset-specific preparation at the forecast origin, and Appendix~\ref{app:architecture} gives the direct-decoder architecture.

\section{Experiments}\label{sec:experiments}


\subsection{Experimental Setup}\label{sec:experimental_setup}

\paragraph{Datasets.}
We evaluate HiSTOR on two simulated longitudinal benchmarks
with known counterfactual outcomes.
CancerSim models tumor volume under four treatment options:
no treatment, chemotherapy, radiotherapy, and their combination,
with treatment-assignment confounding strength
$\gamma\in\{0,1,2,3,4\}$
\citep{geng2017prediction,bica2020estimating,melnychuk2022causal,hess2026igc}.
We use 10,000 training patients, 1,000 validation patients,
and two separately generated 1,000-patient cohorts
for one-step and multi-step testing.
CVSim augments seven base covariates with
$d_{\mathrm{hd}}\in\{0,10,20,50,100,200,500,1000,2000\}$
correlated noisy covariates while keeping
the underlying trajectory-generating process fixed.
The additional covariates remain fixed within patients
in the Static setting and vary over time in the Temporal setting.
We split 5,000 patient trajectories into training,
validation, and test sets using an 80/10/10 patient-level split.
Appendix~\ref{app:data} provides the data-generation details.
To complement these controlled simulation studies, we also include a
semi-synthetic evaluation based on de-identified MIMIC-III covariates;
its data construction, evaluation protocol, and results are reported in
Appendix~\ref{app:mimic}.

\paragraph{Prediction tasks.}
We evaluate independent one-step prediction at observed histories
and multi-step prediction under fixed treatment plans,
without access to future observed outcomes or covariates.
The multi-step plans prescribe sustained combined therapy
in CancerSim and treatment category~2 in CVSim.
Both benchmarks evaluate five future positions:
CVSim scores horizons $\tau=1$ to $5$, whereas CancerSim scores $\tau=2$ to $6$ after an unscored boundary prediction under the factual treatment.
For HiSTOR, the one-step benchmark uses the outcome head,
and the main multi-step comparisons use the direct decoder.
Autoregressive prediction is evaluated in the ablation study.
Appendix~\ref{app:forecast} specifies the treatment plans,
forecast boundaries, and input preparation.

\paragraph{Baselines.}
We compare HiSTOR with MSM \citep{robins2000marginal},
RMSN \citep{lim2018forecasting},
CRN \citep{bica2020estimating},
CT \citep{melnychuk2022causal},
Causal CPC \citep{bouchattaoui2024causal},
CTD-NKO \citep{wang2025enhancing},
and IGC-Net \citep{hess2026igc}.
Within each experimental condition, all methods use the same
data realization, patient splits, and evaluation protocol.
Baseline configurations are provided in Appendix~\ref{app:baselines}.

\paragraph{Evaluation and reporting.}
For the primary CancerSim and CVSim experiments, we report normalized
root mean squared error (NRMSE), using the tumor-death volume threshold
for CancerSim and the task-specific test-target range for CVSim.
These results are summarized as the mean and sample standard deviation
across five random seeds, $\{3,101,42,8,9\}$.
For multi-step evaluation, we average the five horizon-specific
NRMSE values within each seed before aggregating across seeds.
Independent one-step evaluation and the first multi-step horizon
use distinct evaluation protocols.
HiSTOR's backbone and direct-decoder checkpoints are selected
using their respective factual validation objectives,
with counterfactual test targets excluded from model selection.
Appendices~\ref{app:metrics} and~\ref{app:architecture}
provide the metric definitions and optimization details, respectively.

\subsection{Outcome Prediction under Varying Confounding}
\label{sec:cancersim_results}

We evaluate HiSTOR on CancerSim across confounding strengths
$\gamma=0$ to $4$ for independent one-step prediction
and multi-step prediction under sustained combined therapy.
For each condition, relative improvements are calculated
against the baseline with the lowest mean NRMSE.

HiSTOR achieves the lowest mean one-step NRMSE
at all five confounding strengths,
with relative reductions ranging from $4.2\%$ to $13.4\%$
(Table~\ref{tab:cancersim_one_step}).
The gains are more significant under stronger confounding.
At $\gamma=3$, HiSTOR reduces NRMSE
from $0.7122$ for CTD-NKO to $0.6165$,
corresponding to a $13.4\%$ reduction.
At $\gamma=4$, it reduces NRMSE
from $0.7620$ to $0.6703$ relative to the same strongest baseline,
yielding a $12.0\%$ reduction.
These gains exceed the $4.2\%$--$5.2\%$ reductions
observed at $\gamma=0$ and $1$,
indicating a larger predictive advantage
when treatment assignment is more strongly confounded.

\begin{table}[H]
\vspace{-15pt}
\caption{The results for CancerSim one-step prediction. Mean $\pm$ sample SD of NRMSE over five seeds are shown. Best means are bold, second-best underlined; the final row reports HiSTOR's relative reduction over the best baseline.}
\label{tab:cancersim_one_step}
\centering
\footnotesize
\setlength{\tabcolsep}{3pt}
\resizebox{\linewidth}{!}{%
\begin{tabular}{@{}lccccc@{}}
\toprule
Method & $\gamma$ = 0 & $\gamma$ = 1 & $\gamma$ = 2 & $\gamma$ = 3 & $\gamma$ = 4 \\
\midrule
MSM \citep{robins2000marginal}
& 1.0298 $\pm$ 0.0633
& 1.1312 $\pm$ 0.0633
& 1.3005 $\pm$ 0.1039
& 1.6059 $\pm$ 0.1681
& 2.0113 $\pm$ 0.2446 \\
RMSN \citep{lim2018forecasting}
& 0.6994 $\pm$ 0.0690
& 0.6789 $\pm$ 0.0758
& 0.7446 $\pm$ 0.1153
& 0.7324 $\pm$ 0.1042
& 0.8256 $\pm$ 0.0850 \\
CRN \citep{bica2020estimating}
& 0.5969 $\pm$ 0.0639
& 0.6098 $\pm$ 0.0480
& 0.6492 $\pm$ 0.0585
& 0.7143 $\pm$ 0.0868
& 0.8256 $\pm$ 0.0994 \\
CT \citep{melnychuk2022causal}
& \underline{0.5928 $\pm$ 0.0665}
& \underline{0.5974 $\pm$ 0.0855}
& \underline{0.6390 $\pm$ 0.0612}
& 0.7372 $\pm$ 0.0954
& 0.9335 $\pm$ 0.0748 \\
Causal CPC \citep{bouchattaoui2024causal}
& 0.6142 $\pm$ 0.0535
& 0.6228 $\pm$ 0.0636
& 0.6922 $\pm$ 0.1313
& 0.7235 $\pm$ 0.0767
& 0.8427 $\pm$ 0.0857 \\
CTD-NKO \citep{wang2025enhancing}
& 0.6852 $\pm$ 0.0825
& 0.6454 $\pm$ 0.1011
& 0.6659 $\pm$ 0.1056
& \underline{0.7122 $\pm$ 0.0390}
& \underline{0.7620 $\pm$ 0.0566} \\
IGC-Net \citep{hess2026igc}
& 0.6670 $\pm$ 0.0565
& 0.6880 $\pm$ 0.0374
& 0.7537 $\pm$ 0.0828
& 0.9288 $\pm$ 0.2477
& 1.1310 $\pm$ 0.0911 \\
\midrule
HiSTOR
& \textbf{0.5680 $\pm$ 0.0533}
& \textbf{0.5663 $\pm$ 0.0607}
& \textbf{0.5793 $\pm$ 0.0438}
& \textbf{0.6165 $\pm$ 0.0551}
& \textbf{0.6703 $\pm$ 0.0451} \\
\midrule
Relative improvement
& 4.2\%
& 5.2\%
& 9.3\%
& 13.4\%
& 12.0\% \\
\bottomrule
\end{tabular}%
}
\vspace{-5pt}
\end{table}

For multi-step prediction, HiSTOR achieves
the lowest five-horizon-average NRMSE
at every confounding strength
(Table~\ref{tab:cancersim_multi_step}).
Its relative improvement over the strongest baseline
increases steadily from $6.5\%$ at $\gamma=0$
to $14.9\%$ at $\gamma=4$.
In the strongest-confounding setting,
HiSTOR reduces the five-horizon average
from $1.0701$ for the strongest baseline Causal CPC to $0.9105$.
This advantage of HiSTOR is also consistent across multiple repeated runs
with different random seeds at each confounding strength
for the five-horizon average
(Appendix~\ref{app:cancer_paired}).

\begin{table}[h]
\caption{The results for CancerSim multi-step prediction averaged over $\tau=2$ to $6$. Mean $\pm$ sample SD of NRMSE over five seeds are shown. Best means are bold, second-best underlined; the final row reports HiSTOR's relative reduction over the best baseline.}
\label{tab:cancersim_multi_step}
\centering
\footnotesize
\setlength{\tabcolsep}{3pt}
\resizebox{\linewidth}{!}{%
\begin{tabular}{@{}lccccc@{}}
\toprule
Method & $\gamma$ = 0 & $\gamma$ = 1 & $\gamma$ = 2 & $\gamma$ = 3 & $\gamma$ = 4 \\
\midrule
MSM \citep{robins2000marginal}
& 1.0170 $\pm$ 0.0600
& 0.8856 $\pm$ 0.0410
& 0.8822 $\pm$ 0.0224
& 0.9833 $\pm$ 0.0533
& 1.2124 $\pm$ 0.0645 \\
RMSN \citep{lim2018forecasting}
& 0.5735 $\pm$ 0.0637
& 0.5968 $\pm$ 0.0515
& 0.6584 $\pm$ 0.0373
& 0.8303 $\pm$ 0.0851
& 1.0953 $\pm$ 0.0821 \\
CRN \citep{bica2020estimating}
& 0.5573 $\pm$ 0.0157
& 0.5951 $\pm$ 0.0317
& 0.6946 $\pm$ 0.0353
& 0.8111 $\pm$ 0.0463
& 1.0908 $\pm$ 0.0937 \\
CT \citep{melnychuk2022causal}
& 0.5381 $\pm$ 0.0399
& 0.5776 $\pm$ 0.0315
& 0.6619 $\pm$ 0.0549
& 0.8657 $\pm$ 0.0510
& 1.1267 $\pm$ 0.1961 \\
Causal CPC \citep{bouchattaoui2024causal}
& \underline{0.5374 $\pm$ 0.0366}
& 0.5865 $\pm$ 0.0547
& 0.6494 $\pm$ 0.0218
& \underline{0.7934 $\pm$ 0.0543}
& \underline{1.0701 $\pm$ 0.0575} \\
CTD-NKO \citep{wang2025enhancing}
& 0.5462 $\pm$ 0.0502
& \underline{0.5664 $\pm$ 0.0295}
& \underline{0.6473 $\pm$ 0.0514}
& 0.7997 $\pm$ 0.0993
& 1.2394 $\pm$ 0.1614 \\
IGC-Net \citep{hess2026igc}
& 0.6228 $\pm$ 0.0468
& 0.6918 $\pm$ 0.0765
& 0.7373 $\pm$ 0.0370
& 0.9112 $\pm$ 0.1375
& 1.1077 $\pm$ 0.1115 \\
\midrule
HiSTOR
& \textbf{0.5024 $\pm$ 0.0222}
& \textbf{0.5291 $\pm$ 0.0173}
& \textbf{0.5964 $\pm$ 0.0365}
& \textbf{0.7061 $\pm$ 0.0489}
& \textbf{0.9105 $\pm$ 0.0208} \\
\midrule
Rel. improvement
& 6.5\%
& 6.6\%
& 7.9\%
& 11.0\%
& 14.9\% \\
\bottomrule
\end{tabular}%
}
\end{table}

\begin{figure}[!htbp]
\centering
\includegraphics[width=\linewidth]{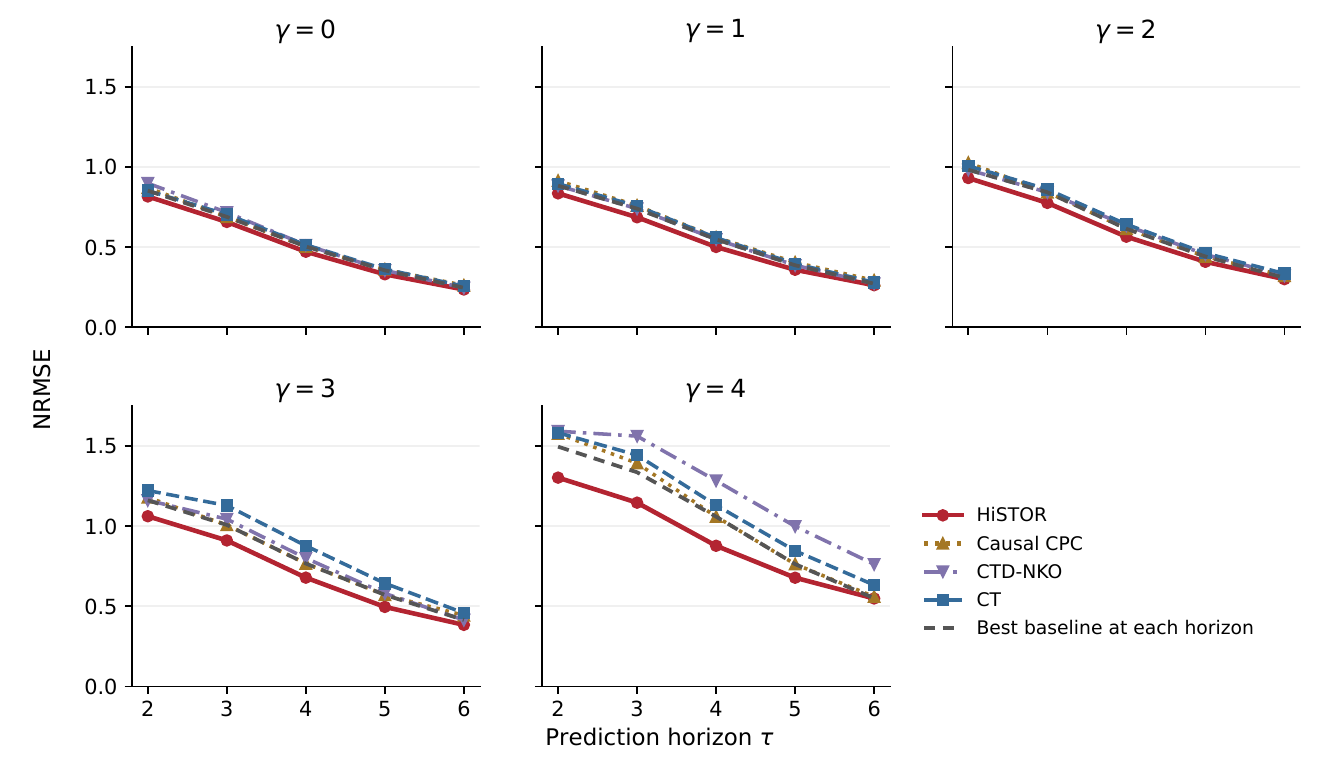}
\caption{CancerSim sustained-treatment NRMSE over $\tau=2,\ldots,6$. Lines are five-seed means; the gray envelope is the lowest mean among all baselines at each horizon and may switch methods. Complete results and SDs are in Appendix~\ref{app:cancer_detailed}.}
\label{fig:cancersim_horizons}
\end{figure}

The horizon-specific results further demonstrate that the improvement
extends across the prediction window.
As shown in Figure~\ref{fig:cancersim_horizons},
HiSTOR achieves the lowest mean NRMSE across all 25 combinations
of confounding strength and prediction horizon.
For example, at $\gamma=4$ and $\tau=4$,
HiSTOR reduces NRMSE from $1.0591$ for Causal CPC to $0.8772$,
a relative reduction of $17.2\%$.
Although the strongest competing method varies across horizons,
HiSTOR consistently achieves the lowest mean NRMSE,
complementing the aggregate multi-step results.
Complete horizon-specific results and paired seed-level comparisons
are provided in Appendix~\ref{app:cancer_detailed}.

\subsection{Outcome Prediction with High-Dimensional Covariates}\label{sec:cvsim_results}

We evaluate HiSTOR on CVSim as the number of correlated noisy covariates increases, while holding the underlying trajectory-generating process fixed. In the Static setting, the additional covariates remain fixed within each patient, assessing prediction in the presence of redundant high-dimensional measurements. In the Temporal setting, these covariates additionally exhibit temporal drift and time-specific noise, assessing prediction when the added measurements also vary over time. Both settings evaluate independent one-step prediction and multi-step prediction under sustained treatment.

\begin{figure}[!htbp]
\centering
\includegraphics[width=\linewidth]{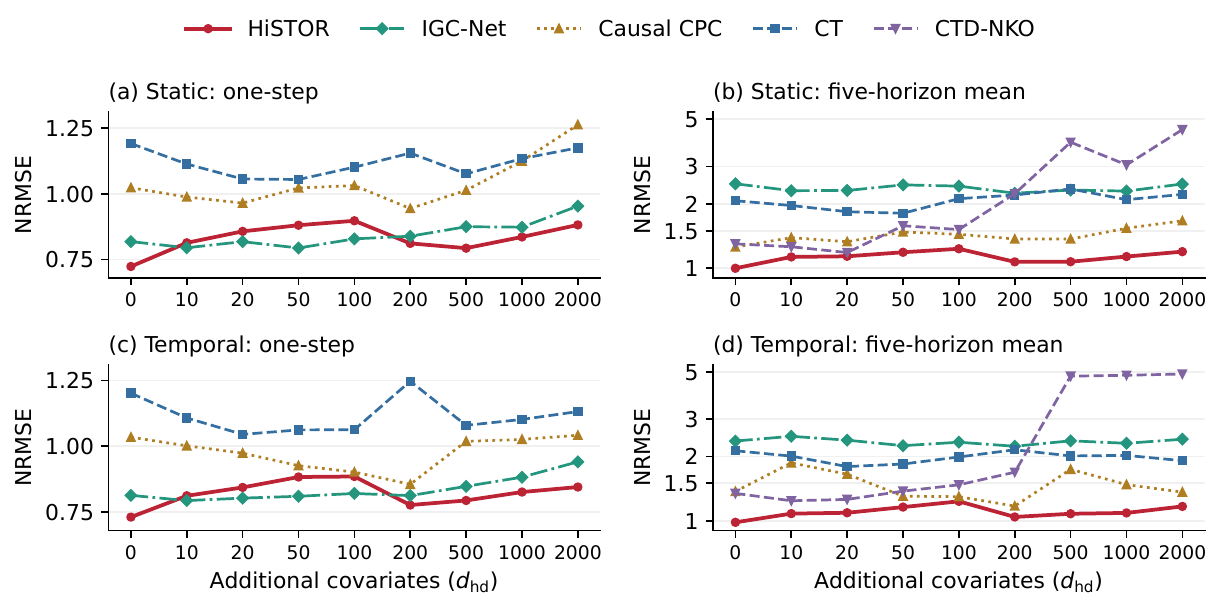}
\caption{CVSim NRMSE across nine $d_{\mathrm{hd}}$ settings. Panels (a,b) show Static and panels (c,d) Temporal results; the left panels report independent one-step error and the right panels report five-horizon means. Lines are five-seed means. Appendix Figure~\ref{fig:cvsim_all_methods} includes all methods.}
\label{fig:cvsim_dimensions}
\end{figure}

For CVSim one-step prediction, HiSTOR achieves the lowest mean NRMSE at all evaluated dimensions from $d_{\mathrm{hd}}=200$ to $2000$ in both settings (Figure~\ref{fig:cvsim_dimensions}(a,c)). At $d_{\mathrm{hd}}=2000$, HiSTOR reduces NRMSE from $0.954$ for the best baseline IGC-Net to $0.882$ in the Static setting and from $0.941$ to $0.845$ in the Temporal setting, corresponding to relative reductions of $7.5\%$ and $10.2\%$. Although IGC-Net has the lowest mean error at $d_{\mathrm{hd}}=10$-$100$, HiSTOR leads at $d_{\mathrm{hd}}=0$ and across all the higher-dimensional settings. Its one-step advantage is therefore most consistent when the observed covariate space is large. These results highlight HiSTOR's predictive advantage in settings with high-dimensional, correlated covariates.

HiSTOR shows a more consistent advantage for multi-step prediction,
achieving the lowest five-horizon-average NRMSE in all 18 combinations of covariate dimension and Static/Temporal setting (Figure~\ref{fig:cvsim_dimensions}(b,d)). At $d_{\mathrm{hd}}=2000$, HiSTOR reduces NRMSE by $28.4\%$ in Static and $14.3\%$ in Temporal relative to baseline Causal CPC, the strongest baseline under this metric in both settings. The advantage also extends across individual prediction horizons: HiSTOR achieves the lowest mean NRMSE in 71 of the 72 conditions with $\tau\geq2$. The only exception is Static $d_{\mathrm{hd}}=20$ at $\tau=2$, where CTD-NKO has a slightly lower mean error.
\begin{table}[h]
\caption{
CVSim NRMSE at $d_{\mathrm{hd}}=2000$, reported as mean $\pm$ sample SD over five seeds.
Independent one-step and multi-step $\tau=1$ are distinct tasks.
Best means are bold, second-best underlined; relative improvements are computed against the lowest-mean baseline in each column.
}
\label{tab:cvsim_highest}
\centering
\footnotesize
\setlength{\tabcolsep}{3pt}

\resizebox{\linewidth}{!}{%
\begin{tabular}{@{}lcccccc@{}}
\toprule
\multirow{2}{*}{Method}
&
\multirow{2}{*}{\shortstack{One-step}}
&
\multicolumn{5}{c}{Multi-step}
\\
\cmidrule(lr){3-7}
&
&
$\tau=1$
&
$\tau=2$
&
$\tau=3$
&
$\tau=4$
&
$\tau=5$
\\
\midrule

\multicolumn{7}{@{}l}{\textit{CVSim Static}} \\
\addlinespace[2pt]

MSM \citep{robins2000marginal}
& 2.797 $\pm$ 1.659
& 3.797 $\pm$ 2.357
& 3.860 $\pm$ 2.392
& 3.914 $\pm$ 2.419
& 4.017 $\pm$ 2.479
& 4.082 $\pm$ 2.515 \\

RMSN \citep{lim2018forecasting}
& 2.857 $\pm$ 0.975
& 3.775 $\pm$ 1.181
& 3.202 $\pm$ 0.735
& 3.147 $\pm$ 0.639
& 3.323 $\pm$ 0.591
& 3.544 $\pm$ 0.553 \\

CRN \citep{bica2020estimating}
& 4.359 $\pm$ 0.909
& 6.076 $\pm$ 1.675
& 7.004 $\pm$ 1.878
& 7.669 $\pm$ 1.911
& 8.433 $\pm$ 1.907
& 9.192 $\pm$ 1.945 \\

CT \citep{melnychuk2022causal}
& 1.175 $\pm$ 0.318
& 1.412 $\pm$ 0.454
& 1.846 $\pm$ 0.472
& 2.234 $\pm$ 0.802
& 2.632 $\pm$ 1.150
& 2.984 $\pm$ 1.342 \\

Causal CPC \citep{bouchattaoui2024causal}
& 1.262 $\pm$ 0.307
& 1.533 $\pm$ 0.624
& \underline{1.586 $\pm$ 0.511}
& \underline{1.640 $\pm$ 0.372}
& \underline{1.741 $\pm$ 0.262}
& \underline{1.858 $\pm$ 0.308} \\

CTD-NKO \citep{wang2025enhancing}
& 3.370 $\pm$ 1.719
& 3.856 $\pm$ 1.449
& 3.865 $\pm$ 2.000
& 4.222 $\pm$ 2.268
& 4.859 $\pm$ 2.282
& 5.567 $\pm$ 2.135 \\

IGC-Net \citep{hess2026igc}
& \underline{0.954 $\pm$ 0.274}
& \underline{1.054 $\pm$ 0.288}
& 1.699 $\pm$ 0.633
& 2.476 $\pm$ 1.373
& 2.972 $\pm$ 1.447
& 4.226 $\pm$ 1.839 \\

\midrule
HiSTOR
& \textbf{0.882 $\pm$ 0.257}
& \textbf{0.993 $\pm$ 0.322}
& \textbf{1.070 $\pm$ 0.379}
& \textbf{1.180 $\pm$ 0.404}
& \textbf{1.294 $\pm$ 0.406}
& \textbf{1.450 $\pm$ 0.405} \\

Rel. improvement
& 7.5\%
& 5.8\%
& 32.5\%
& 28.0\%
& 25.7\%
& 22.0\% \\

\midrule
\multicolumn{7}{@{}l}{\textit{CVSim Temporal}} \\
\addlinespace[2pt]

MSM \citep{robins2000marginal}
& 2.776 $\pm$ 1.661
& 3.841 $\pm$ 2.425
& 3.905 $\pm$ 2.432
& 3.944 $\pm$ 2.420
& 4.047 $\pm$ 2.466
& 4.082 $\pm$ 2.486 \\

RMSN \citep{lim2018forecasting}
& 2.605 $\pm$ 0.967
& 3.040 $\pm$ 1.321
& 2.412 $\pm$ 0.755
& 2.302 $\pm$ 0.718
& 2.534 $\pm$ 0.699
& 2.826 $\pm$ 0.679 \\

CRN \citep{bica2020estimating}
& 4.617 $\pm$ 2.500
& 7.551 $\pm$ 5.717
& 8.658 $\pm$ 6.450
& 9.187 $\pm$ 6.421
& 9.794 $\pm$ 6.411
& 10.452 $\pm$ 6.461 \\

CT \citep{melnychuk2022causal}
& 1.132 $\pm$ 0.283
& 1.358 $\pm$ 0.451
& 1.595 $\pm$ 0.398
& 1.894 $\pm$ 0.305
& 2.210 $\pm$ 0.615
& 2.510 $\pm$ 0.904 \\

Causal CPC \citep{bouchattaoui2024causal}
& 1.042 $\pm$ 0.185
& 1.257 $\pm$ 0.512
& \underline{1.253 $\pm$ 0.457}
& \underline{1.318 $\pm$ 0.427}
& \underline{1.428 $\pm$ 0.429}
& \underline{1.556 $\pm$ 0.455} \\

CTD-NKO \citep{wang2025enhancing}
& 3.382 $\pm$ 1.799
& 4.454 $\pm$ 2.248
& 4.489 $\pm$ 2.374
& 4.748 $\pm$ 2.562
& 5.140 $\pm$ 2.754
& 5.550 $\pm$ 2.875 \\

IGC-Net \citep{hess2026igc}
& \underline{0.941 $\pm$ 0.248}
& \underline{1.084 $\pm$ 0.307}
& 1.531 $\pm$ 0.717
& 2.315 $\pm$ 1.327
& 3.185 $\pm$ 1.352
& 3.945 $\pm$ 1.992 \\

\midrule
HiSTOR
& \textbf{0.845 $\pm$ 0.254}
& \textbf{0.974 $\pm$ 0.230}
& \textbf{1.041 $\pm$ 0.307}
& \textbf{1.150 $\pm$ 0.361}
& \textbf{1.251 $\pm$ 0.442}
& \textbf{1.421 $\pm$ 0.464} \\

Rel. improvement
& 10.2\%
& 10.2\%
& 16.9\%
& 12.7\%
& 12.4\%
& 8.7\% \\

\bottomrule
\end{tabular}%
}

\end{table}
At the highest evaluated dimension, $d_{\mathrm{hd}}=2000$, HiSTOR achieves the lowest mean error for both one-step prediction and all five multi-step horizons in both settings (Table~\ref{tab:cvsim_highest}). At $\tau=2$, its NRMSE reductions relative to the strongest baseline reach $32.5\%$ in Static and $16.9\%$ in Temporal. These gains persist through the final evaluated horizon, with corresponding reductions of $22.0\%$ and $8.7\%$ at $\tau=5$. Together, these results highlight HiSTOR's advantage in forecasting treatment-conditioned trajectories with high-dimensional covariates, whether the additional measurements remain fixed or vary over time. Complete results are provided in Appendix~\ref{app:cvsim_detailed}.

\subsection{Ablation Study}\label{sec:ablation}

\begin{table}[h]
\caption{
CancerSim ablation NRMSE, reported as mean $\pm$ sample SD
over five seeds.
Bold values indicate the lowest mean in each column.
One-step results for autoregressive prediction are omitted
because it uses the same frozen backbone as HiSTOR.
}
\label{tab:ablation}
\centering
\footnotesize
\setlength{\tabcolsep}{2.5pt}
\renewcommand{\arraystretch}{1.12}
\begin{tabular*}{\linewidth}{@{\extracolsep{\fill}}lcccc@{}}
\toprule
\multirow{2}{*}{Variant}
& \multicolumn{2}{c}{$\gamma=1$}
& \multicolumn{2}{c}{$\gamma=4$} \\
\cmidrule(lr){2-3}\cmidrule(lr){4-5}
& One-step & Five-horizon mean
& One-step & Five-horizon mean \\
\midrule
HiSTOR
& $\mathbf{0.5663\pm0.0607}$
& $\mathbf{0.5291\pm0.0173}$
& $\mathbf{0.6703\pm0.0451}$
& $\mathbf{0.9105\pm0.0208}$ \\

w/o Covariate Bottleneck
& $0.5928\pm0.0500$
& $0.5664\pm0.0495$
& $0.6797\pm0.0356$
& $0.9555\pm0.0349$ \\

w/o Distribution Matching
& $0.5777\pm0.0622$
& $0.5516\pm0.0205$
& $0.6805\pm0.0336$
& $0.9189\pm0.0498$ \\

w/o Reconstruction
& $0.5771\pm0.0524$
& $0.5492\pm0.0282$
& $0.6813\pm0.0554$
& $0.9151\pm0.0570$ \\

Autoregressive Prediction
& ---
& $0.5502\pm0.0210$
& ---
& $0.9193\pm0.0409$ \\
\bottomrule
\end{tabular*}
\end{table}

We assess the contributions of HiSTOR's components using the CancerSim benchmark at confounding strengths $\gamma=1$ and $4$. We separately remove the covariate encoding bottleneck, distribution-matching regularization, and reconstruction objectives. We also replace direct multi-step decoding with autoregressive prediction. Table~\ref{tab:ablation} summarizes the results over five seeds. All relative error increases below are calculated against the full model at the corresponding confounding strength.
The full model achieves the lowest mean error in every reported comparison. Removing the covariate bottleneck produces the largest increases in multi-step errors. The five-horizon-average NRMSE rises from $0.5291$ to $0.5664$ at $\gamma=1$ and from $0.9105$ to $0.9555$ at $\gamma=4$, corresponding to relative increases of $7.0\%$ and $4.9\%$. The corresponding one-step errors increase by $4.7\%$ and $1.4\%$.
These results identify the covariate bottleneck as the most influential component among those evaluated, with benefits across both prediction tasks and confounding strengths.

Distribution matching and reconstruction provide additional improvements whose magnitudes depend on the setting. Removing distribution matching increases the five-horizon-average NRMSE by $4.3\%$ at $\gamma=1$ and $0.9\%$ at $\gamma=4$. Removing the reconstruction objectives increases the same metric by $3.8\%$ and $0.5\%$, respectively. Both modifications also increase the mean one-step error. Thus, these objectives contribute to predictive performance, although their effects are smaller than those of the covariate bottleneck in the evaluated settings.

Finally, replacing parallel decoding with autoregressive prediction
increases the five-horizon-average NRMSE by $4.0\%$ at $\gamma=1$ and $1.0\%$ at $\gamma=4$. This result indicates the benefit of direct multi-step parallel decoding compared to autoregressive predictions where the error may accumulate across multi-steps.

\section{Conclusion}\label{sec:discussion}
We introduced HiSTOR, a structured representation learning framework for longitudinal counterfactual outcome prediction with high-dimensional covariates. HiSTOR combines current-covariate compression with masked history encoding and partitions their fusion into shared, outcome-specific, treatment-specific, and residual components. Joint predictive and reconstruction objectives encourage distinct functional roles and retention of information about the observed patient state. Across CancerSim and CVSim, HiSTOR achieves the lowest mean five-horizon-average NRMSE in every evaluated setting,
with one-step advantages across CancerSim confounding strengths
and at higher CVSim dimensions.

Our evaluation currently focuses on simulated and semi-synthetic benchmarks. Future work will extend evaluation to real-world clinical data and broader treatment regimes, explore separate modeling of future covariate dynamics, and incorporate uncertainty quantification~\citep{liu2026bayesian,liu2026missingness}. We will also investigate when compressed histories retain sufficient information for confounding adjustment.

\subsection*{Acknowledgments}
Q.L. was supported in part by the National Human Genome Research Institute (NHGRI) under Award Number R00HG013661, the Operations Core of the Claude D. Pepper Older Americans Independence Center at Yale School of Medicine (P30AG021342), and the YSPH Transformation Pilot Grant. J.H. was partially supported by US National Science Foundation IIS-2412195, CCF-2400785, the Cancer Prevention and Research Institute of Texas (CPRIT) award (RP230363) and the National Institutes of Health (NIH) R01 awards (1R01AI190103-01 and 1R01LM015497-01). We thank Yixuan Luo for conducting exploratory experiments in the early stage of this project.

\bibliographystyle{iclr2027_conference} \bibliography{paper_references}

@article{liu2021density,
  title={Density estimation using deep generative neural networks},
  author={Liu, Qiao and Xu, Jiaze and Jiang, Rui and Wong, Wing Hung},
  journal={Proceedings of the national academy of sciences},
  volume={118},
  number={15},
  pages={e2101344118},
  year={2021},
  publisher={National Academy of Sciences}
}

@article{liu2026bayesian,
  title={A bayesian generative modeling approach for arbitrary conditional inference},
  author={Liu, Qiao and Wong, Wing Hung},
  journal={arXiv e-prints},
  pages={arXiv--2601},
  year={2026}
}

@inproceedings{johansson2016learning,
  title={Learning representations for counterfactual inference},
  author={Johansson, Fredrik and Shalit, Uri and Sontag, David},
  booktitle={International conference on machine learning},
  pages={3020--3029},
  year={2016},
  organization={PMLR}
}

@inproceedings{alaa2018limits,
  title={Limits of estimating heterogeneous treatment effects: Guidelines for practical algorithm design},
  author={Alaa, Ahmed and Schaar, Mihaela},
  booktitle={International conference on machine learning},
  pages={129--138},
  year={2018},
  organization={PMLR}
}

@article{wager2018estimation,
  title={Estimation and inference of heterogeneous treatment effects using random forests},
  author={Wager, Stefan and Athey, Susan},
  journal={Journal of the American Statistical Association},
  volume={113},
  number={523},
  pages={1228--1242},
  year={2018},
  publisher={Taylor \& Francis}
}

@inproceedings{yoon2018ganite,
  title={GANITE: Estimation of individualized treatment effects using generative adversarial nets},
  author={Yoon, Jinsung and Jordon, James and Van Der Schaar, Mihaela},
  booktitle={International conference on learning representations},
  year={2018}
}

@inproceedings{curth2021nonparametric,
  title={Nonparametric estimation of heterogeneous treatment effects: From theory to learning algorithms},
  author={Curth, Alicia and Van der Schaar, Mihaela},
  booktitle={International conference on artificial intelligence and statistics},
  pages={1810--1818},
  year={2021},
  organization={PMLR}
}

@article{robins1986new,
  title={A new approach to causal inference in mortality studies with a sustained exposure period—application to control of the healthy worker survivor effect},
  author={Robins, James},
  journal={Mathematical modelling},
  volume={7},
  number={9-12},
  pages={1393--1512},
  year={1986},
  publisher={Elsevier}
}

@article{bica2020estimating,
  title={Estimating counterfactual treatment outcomes over time through adversarially balanced representations},
  author={Bica, Ioana and Alaa, Ahmed M and Jordon, James and Van Der Schaar, Mihaela},
  journal={arXiv preprint arXiv:2002.04083},
  year={2020}
}

@inproceedings{melnychuk2022causal,
  title={Causal transformer for estimating counterfactual outcomes},
  author={Melnychuk, Valentyn and Frauen, Dennis and Feuerriegel, Stefan},
  booktitle={International conference on machine learning},
  pages={15293--15329},
  year={2022},
  organization={PMLR}
}

@inproceedings{hess2026igc,
  title={Igc-net for conditional average potential outcome estimation over time},
  author={Hess, Konstantin and Frauen, Dennis and Melnychuk, Valentyn and Feuerriegel, Stefan},
  booktitle={International Conference on Learning Representations},
  volume={2026},
  pages={74263--74298},
  year={2026}
}

@article{liu2024encoding,
  title={An encoding generative modeling approach to dimension reduction and covariate adjustment in causal inference with observational studies},
  author={Liu, Qiao and Chen, Zhongren and Wong, Wing Hung},
  journal={Proceedings of the National Academy of Sciences},
  volume={121},
  number={23},
  pages={e2322376121},
  year={2024},
  publisher={National Academy of Sciences}
}

@article{lim2018forecasting,
  title={Forecasting treatment responses over time using recurrent marginal structural networks},
  author={Lim, Bryan},
  journal={Advances in neural information processing systems},
  volume={31},
  year={2018}
}

@inproceedings{li2021g,
  title={G-net: a recurrent network approach to g-computation for counterfactual prediction under a dynamic treatment regime},
  author={Li, Rui and Hu, Stephanie and Lu, Mingyu and Utsumi, Yuria and Chakraborty, Prithwish and Sow, Daby M and Madan, Piyush and Li, Jun and Ghalwash, Mohamed and Shahn, Zach and others},
  booktitle={Machine Learning for Health},
  pages={282--299},
  year={2021},
  organization={PMLR}
}

@article{allam2021analyzing,
  title={Analyzing patient trajectories with artificial intelligence},
  author={Allam, Ahmed and Feuerriegel, Stefan and Rebhan, Michael and Krauthammer, Michael},
  journal={Journal of medical internet research},
  volume={23},
  number={12},
  pages={e29812},
  year={2021},
  publisher={JMIR Publications Toronto, Canada}
}

@article{bica2021real,
  title={From real-world patient data to individualized treatment effects using machine learning: current and future methods to address underlying challenges},
  author={Bica, Ioana and Alaa, Ahmed M and Lambert, Craig and Van Der Schaar, Mihaela},
  journal={Clinical Pharmacology \& Therapeutics},
  volume={109},
  number={1},
  pages={87--100},
  year={2021},
  publisher={Wiley Online Library}
}

@article{battalio2021sense2stop,
  title={Sense2Stop: a micro-randomized trial using wearable sensors to optimize a just-in-time-adaptive stress management intervention for smoking relapse prevention},
  author={Battalio, Samuel L and Conroy, David E and Dempsey, Walter and Liao, Peng and Menictas, Marianne and Murphy, Susan and Nahum-Shani, Inbal and Qian, Tianchen and Kumar, Santosh and Spring, Bonnie},
  journal={Contemporary Clinical Trials},
  volume={109},
  pages={106534},
  year={2021},
  publisher={Elsevier}
}

@misc{murray2016evaluating,
  title={Evaluating digital health interventions: key questions and approaches},
  author={Murray, Elizabeth and Hekler, Eric B and Andersson, Gerhard and Collins, Linda M and Doherty, Aiden and Hollis, Chris and Rivera, Daniel E and West, Robert and Wyatt, Jeremy C},
  journal={American journal of preventive medicine},
  volume={51},
  number={5},
  pages={843--851},
  year={2016},
  publisher={Elsevier}
}

@article{xiong2024g,
  title={G-transformer: Counterfactual outcome prediction under dynamic and time-varying treatment regimes},
  author={Xiong, Hong and Wu, Feng and Deng, Leon and Su, Megan and Shahn, Zach and Lehman, Li-wei H},
  journal={Proceedings of machine learning research},
  volume={252},
  pages={https--proceedings},
  year={2024}
}

@article{robins2000marginal,
  title={Marginal structural models and causal inference in epidemiology},
  author={Robins, James M and Hernan, Miguel Angel and Brumback, Babette},
  journal={Epidemiology},
  volume={11},
  number={5},
  pages={550--560},
  year={2000}
}

@article{robins2008estimation,
  title={Estimation of the causal effects of time-varying exposures},
  author={Robins, James and Hernan, Miguel},
  journal={Chapman \& Hall/CRC Handbooks of Modern Statistical Methods},
  pages={553--599},
  year={2008},
  publisher={Chapman and Hall/CRC}
}

@article{gulrajani2017improved,
  title={Improved training of wasserstein gans},
  author={Gulrajani, Ishaan and Ahmed, Faruk and Arjovsky, Martin and Dumoulin, Vincent and Courville, Aaron},
  journal={Advances in neural information processing systems},
  volume={30},
  year={2017}
}

@article{bouchattaoui2024causal,
  title={Causal contrastive learning for counterfactual regression over time},
  author={Bouchattaoui, Mouad E and Tami, Myriam and Lepetit, Benoit and Courn{\`e}de, Paul-Henry},
  journal={Advances in Neural Information Processing Systems},
  volume={37},
  pages={1333--1369},
  year={2024}
}

@inproceedings{wang2025enhancing,
  title={Enhancing counterfactual estimation: a focus on temporal treatments},
  author={Wang, Xin and Lyu, Shengfei and Luo, Kangyang and Yang, Lishan and Chen, Huanhuan and Miao, Chunyan},
  booktitle={Proceedings of the Thirty-Fourth International Joint Conference on Artificial Intelligence},
  pages={6442--6450},
  year={2025}
}

@article{geng2017prediction,
  title={Prediction of treatment response for combined chemo-and radiation therapy for non-small cell lung cancer patients using a bio-mathematical model},
  author={Geng, Changran and Paganetti, Harald and Grassberger, Clemens},
  journal={Scientific reports},
  volume={7},
  number={1},
  pages={13542},
  year={2017},
  publisher={Nature Publishing Group UK London}
}

@inproceedings{hassanpour2019learning,
  title={Learning disentangled representations for counterfactual regression},
  author={Hassanpour, Negar and Greiner, Russell},
  booktitle={International Conference on Learning Representations},
  year={2019}
}

@article{liu2026ai,
  title={An AI-powered Bayesian generative modeling approach for causal inference in observational studies},
  author={Liu, Qiao and Wong, Wing Hung},
  journal={Journal of the American Statistical Association},
  pages={1--12},
  year={2026},
  publisher={Taylor \& Francis}
}

@inproceedings{zhang2021treatment,
  title={Treatment effect estimation with disentangled latent factors},
  author={Zhang, Weijia and Liu, Lin and Li, Jiuyong},
  booktitle={Proceedings of the AAAI Conference on Artificial Intelligence},
  volume={35},
  number={12},
  pages={10923--10930},
  year={2021}
}

@article{luo2026bgm,
  title={BGM-IV: an AI-powered Bayesian generative modeling approach for instrumental variable analysis},
  author={Luo, Guyue and Liu, Qiao},
  journal={arXiv preprint arXiv:2605.07029},
  year={2026}
}

@article{johnson2016mimic,
  title     = {{MIMIC-III}, a freely accessible critical care database},
  author    = {Johnson, Alistair E. W. and Pollard, Tom J. and Shen, Lu and
               Lehman, Li-Wei H. and Feng, Mengling and Ghassemi, Mohammad and
               Moody, Benjamin and Szolovits, Peter and Celi, Leo Anthony and
               Mark, Roger G.},
  journal   = {Scientific Data},
  volume    = {3},
  number    = {1},
  pages     = {160035},
  year      = {2016},
  publisher = {Nature Publishing Group}
}

@inproceedings{wang2020mimic,
  title     = {{MIMIC-Extract}: A data extraction, preprocessing, and
               representation pipeline for {MIMIC-III}},
  author    = {Wang, Shirly and McDermott, Matthew B. A. and Chauhan, Geeticka and
               Ghassemi, Marzyeh and Hughes, Michael C. and Naumann, Tristan},
  booktitle = {Proceedings of the ACM Conference on Health, Inference, and Learning},
  pages     = {222--235},
  year      = {2020}
}

@article{liu2026missingness,
  title={Missingness-aware Data Imputation via AI-powered Bayesian Generative Modeling},
  author={Liu, Qiao},
  journal={arXiv preprint arXiv:2605.01676},
  year={2026}
}

\clearpage
\appendix
\section{Notation and Temporal Alignment}\label{app:notation}

\begin{table}[h]
\caption{Core notation.}
\centering
\small
\begin{tabularx}{\linewidth}{@{}lX@{}}
\toprule
Symbol & Meaning \\
\midrule

$X_t$
& Treatment assigned at time $t$ \\

$Y_t$
& Post-treatment outcome associated with $X_t$ \\

$V_t$
& Covariates available before treatment at time $t$, including baseline and, when present, time-varying components \\

$\mathcal H_t$
& Observed information available before treatment at time $t$ \\

$W_t$
& Implementation-only notation for the dataset-specific raw input supplied to the current-state projection $P$ \\

$S$
& Static CancerSim patient-type feature; the baseline component of $V_t$ used through the static-feature channel \\
$s_i^{\mathrm{stat}}$
& Dataset-specific static feature vector supplied to the direct decoder \\
$P,E_V$
& Dataset-specific projection of $W_t$ and the fixed-width current-state encoder \\ 

$u_t,c_t$
& Intermediate current-state embedding and fixed-width current-state encoding \\

$\mathcal T,Q$
& Complete temporal history network and structured representation mapping \\

$e_X,e_Y$
& Learned treatment and outcome embedding networks \\

$\operatorname{LN}_{X,Y,C}$
& Separately parameterized layer-normalization operators for the treatment, outcome, and current-state inputs \\

$s_t$
& Fused history token aligned with the observations available through time $t-1$ \\

$\operatorname{PE}$
& Addition of fixed sinusoidal positional encodings to the history-token sequence \\

$\operatorname{TD}$
& Stack of temporal Transformer decoder blocks used as the masked history encoder \\

$W_c,W_m$
& Learned projections into the fused-token and cross-attention-memory spaces, respectively; bias terms are omitted from the notation \\

$T,[\cdot]_t$
& Padded sequence length and selection of the output at sequence position $t$ \\

$M_{tj},M_{\mathrm{pad}}$
& Causal attention mask for query position $t$ and key position $j$, and the key-padding mask \\

$h_{t-1}$
& Masked representation of the observations available through time $t-1$ \\

$\Phi_t,Z_t$
& Fused patient representation and its structured latent representation \\

$z_t^0,z_t^1,z_t^2,z_t^3$
& Shared, outcome-specific, treatment-specific, and residual-reconstruction branches \\

$G_V,D_{\mathrm{raw}},R$
& Embedding decoder, standardized raw-input decoder, and residual-reconstruction predictor \\

$D_c,D_u$
& Critics operating in the current-state encoding and intermediate embedding spaces \\

$f_Y,f_X$
& One-step outcome head and auxiliary treatment head \\

$\widehat y_t(x_t)$
& Predicted outcome under the observed or externally specified treatment $x_t$ \\

$\ell_t^X$
& Unnormalized logits produced by the auxiliary treatment head \\

$p_t^X$
& Predicted treatment-probability vector, defined as $p_t^X=\operatorname{softmax}(\ell_t^X)$ \\

$\widehat x_t$
& Hard auxiliary treatment prediction, defined as $\widehat x_t=\arg\max_k p_{t,k}^X$ \\

$\mathcal I,\mathcal I_b$
& Set of valid patient--time observations in the dataset and in a minibatch, respectively \\

$\beta_t,\eta_k$
& Direct-decoder boundary encoding and treatment-GRU state at horizon offset $k$ \\

$E_X,g_k$
& Treatment embedding used by the direct decoder and its outcome head for horizon offset $k$ \\

$\widetilde y_{t-1}$
& Previous-outcome input after dataset-specific forecast-boundary preparation \\

$w_k$
& Training-derived weight for treatment category $k$ \\

$\mathcal W,\tau$
& Minibatch of complete factual windows and their prediction horizon \\

$\operatorname{sg}$
& Identity in the forward pass with zero gradient through its argument \\

$d_W,d_u,d_c$
& Dimensions of the dataset-specific raw input, intermediate embedding, and fixed-width current-state encoding \\

$d_h,d_Z$
& Dimensions of the historical representation and structured latent representation \\

$d_0,d_1,d_2,d_3$
& Widths of the four structured latent branches \\

$d_y$
& Outcome dimension \\

$m_{it}$
& Indicator of an active patient--time entry in the evaluation metric \\

\bottomrule
\end{tabularx}
\end{table}
For one patient, the information available before treatment at time $t$ is
\[
\mathcal H_t=(X_{<t},Y_{<t},V_{\leq t}).
\]
The model processes this information through two complementary paths. The temporal path constructs $h_{t-1}$ from the shifted sequence of preceding treatments, outcomes, and current-state encodings. The current-state path separately constructs $c_t$ from the dataset-specific raw input $W_t$. These representations are fused as
\[
\Phi_t=[h_{t-1},c_t],
\qquad
Z_t=Q(\Phi_t).
\]

The symbol $W_t$ is used only for implementation details in the appendix. Its construction differs across datasets and is specified in Appendix~\ref{app:data}. In all cases, it contains only information observed before treatment $X_t$.

At sequence position $t$, the shifted inputs $X_{t-1}$, $Y_{t-1}$, and $c_{t-1}$ are combined into the fused history token $s_t$. Fixed positional encodings are added only to the token sequence. The projected sequence $W_m c_{0:T-1}$ forms the cross-attention memory and receives no positional encoding. The causal mask permits position $t$ to attend only to positions $j\leq t$, while $M_{\mathrm{pad}}$ excludes inactive padded positions. Because the inputs are shifted by one position, the resulting output $h_{t-1}$ summarizes observations only through time $t-1$.

The auxiliary treatment head outputs logits $\ell_t^X$, which are converted into categorical probabilities by
\[
p_t^X=\operatorname{softmax}(\ell_t^X).
\]
If a hard auxiliary prediction is needed, it is obtained as
\[
\widehat x_t=\arg\max_k p_{t,k}^X.
\]
This auxiliary prediction does not provide the treatment input to the outcome model. One-step outcome prediction instead conditions on the observed or externally specified treatment $x_t$.

For one-step prediction, the model uses the information available at the prediction origin and predicts $Y_t$ under the observed or specified treatment $x_t$. For multi-step prediction, the model receives a specified treatment sequence and does not use future outcomes or future covariates. The first output of a multi-step evaluation may use different patient origins from the aggregate independent one-step benchmark and therefore need not reproduce the same statistic. Dataset-specific input construction is described in Appendix~\ref{app:data}, while sequence initialization and forecast-boundary preparation are described in Appendices~\ref{app:architecture} and~\ref{app:forecast}.

\section{Architecture and Objectives}\label{app:architecture}
\paragraph{Structured branch configuration.}
We set $(d_0,d_1,d_2,d_3)=(24,24,48,64)$, giving $d_Z=160$. The same allocation is used for all datasets. During backbone training, dropout is applied to $[z_t^0,z_t^2]$ before the auxiliary treatment head. Both simulation datasets use categorical treatment prediction. The direct multi-step decoder uses all four branches as its representation context.

\paragraph{Representation and temporal configuration.}
Both simulation adapters use $d_u=128$, $d_c=16$, $d_h=128$, and $d_Z=160$, so that $Q$ maps $\mathbb R^{144}$ to $\mathbb R^{160}$. The raw input width $d_W$ depends on the dataset and input configuration. The treatment and outcome embedding MLPs each have two hidden layers of width 256 and output width 128. The structured mapping likewise has two hidden layers of width 256. The temporal backbone has four Transformer decoder blocks, four attention heads per block, a feedforward width of 256, and dropout 0.15. Fixed sinusoidal positional encodings are added only to the fused tokens; the projected current-state memory has no added positional encoding. Both attention paths apply the same causal and key-padding masks. The direct prediction decoder uses a boundary and GRU hidden width of 32, a treatment embedding width of 8, and a separate outcome head with hidden width 32 for each horizon.

\paragraph{Current-state projection and reconstruction objectives.}
The main text writes the current-state encoding at a high level as
$c_t=e_V(V_t)$. In the implementation, this encoding is computed from the
dataset-specific input $W_t$ using a projection followed by the encoder:
\[
u_t=P(W_t),\qquad c_t=E_V(u_t).
\]
Thus, $e_V$ in the main text is shorthand for the implementation-level
mapping. The input $W_t$ is dataset-specific: CVSim uses $W_t=V_t$,
whereas CancerSim uses $W_{it}=[Y_{i,t-1},S_i]$, as defined in
Appendix~\ref{app:data}.

They are connected by two first-level reconstruction objectives:
\begin{equation}
 \widehat u_t=G_V(c_t),\qquad
 \widehat W_t^{\mathrm{std}}=D_{\mathrm{raw}}(c_t),\qquad
 \mathcal L_{\mathrm{emb}}=\operatorname{MSE}(\widehat u,u),\quad
 \mathcal L_{\mathrm{raw}}
 =\operatorname{MSE}(\widehat W^{\mathrm{std}},W^{\mathrm{std}}).
 \label{eq:reconstruction}
\end{equation}
Here $G_V$ is the embedding decoder and $D_{\mathrm{raw}}$ reconstructs the standardized raw input. Both losses are averaged over valid patient--time pairs. The raw reconstruction targets are standardized using training-set statistics that remain fixed during validation and testing.

The residual-reconstruction target is the embedding information left unreconstructed by the first-level decoder:
\begin{equation}
 r_t=\operatorname{sg}\!\left[u_t-G_V(c_t)\right],
 \qquad
 \mathcal L_{\mathrm{res}}
 =\operatorname{MSE}\!\left(R(z_t^3),r_t\right),
 \label{eq:residual}
\end{equation}
where $\operatorname{sg}$ blocks gradients through the residual target. The complete reconstruction objective is
\begin{equation}
 \mathcal L_{\mathrm{rec}}
 =
 \mathcal L_{\mathrm{emb}}
 +\mathcal L_{\mathrm{raw}}
 +\mathcal L_{\mathrm{res}}.
 \label{eq:complete_reconstruction}
\end{equation}
The embedding decoder $G_V$ also maps standard-normal samples into the embedding space for the adversarial distributional objective in Equation~\ref{eq:adv}.

\paragraph{Projection branches.}
Both projection branches are used in the experiments. CancerSim and the five CVSim settings with $d_{\mathrm{hd}}\leq100$ use fixed sinusoidal features followed by an MLP. The feature expansion concatenates $W_t$ with $\sin(2^k W_t)$ and $\cos(2^k W_t)$ for $k=0,\ldots,7$, and each of the two hidden layers has width $\max(\lfloor17d_W/2\rfloor,128)$. The four CVSim settings with $d_{\mathrm{hd}}\geq200$ use an MLP directly on $W_t$, with two hidden layers of width $\max(d_W,128)$. Thus, the embedding width remains 128, but the projection network's capacity varies with the input dimension.

\paragraph{Sequence initialization.}
The initial treatment input is the all-zero vector, used as a sequence-start placeholder rather than a treatment category. In particular, it differs from the CancerSim no-treatment category, encoded as $(1,0,0,0)$. Dataset-specific initialization of the outcome and raw current-state inputs is described in Appendix~\ref{app:data}. The first historical current-state slot is set directly to zero in the $d_c$-dimensional encoding space, before the $W_c$ and $W_m$ projections; it is not obtained by encoding a zero-valued raw input.
\subsection*{Temporal Network and Alignment Details}

Let $d_h$ denote the temporal hidden dimension. The learned multilayer-perceptron embeddings $e_X$ and $e_Y$ map the preceding treatment and outcome into $\mathbb R^{d_h}$, while the learned projection $W_c$ maps the preceding current-state encoding into the same space. The history token at sequence position $t$ is
\begin{equation}
 s_t=
 \operatorname{LN}_X(e_X(X_{t-1}))
 +\operatorname{LN}_Y(e_Y(Y_{t-1}))
 +\operatorname{LN}_C(W_c c_{t-1}),
 \label{eq:token}
\end{equation}
where $\operatorname{LN}_X$, $\operatorname{LN}_Y$, and $\operatorname{LN}_C$ are separately parameterized layer-normalization operators.

Let $\operatorname{PE}$ denote the addition of fixed sinusoidal positional encodings to the token sequence. A second learned projection, $W_m$, maps the historical current-state encodings to the cross-attention memory. The temporal network produces
\begin{equation}
 h_{t-1}=
 \left[
 \operatorname{TD}\!\left(
 \operatorname{PE}(s_{1:T}),
 W_m c_{0:T-1};
 M,M_{\mathrm{pad}}
 \right)
 \right]_t,
 \label{eq:backbone}
\end{equation}
where $T$ is the padded sequence length, $\operatorname{TD}$ is the stack of temporal Transformer decoder blocks, and $[\cdot]_t$ selects its output at position $t$. The first input supplies the position-encoded tokens, and the second supplies the historical memory without added positional encoding. These decoder blocks implement the historical encoder $\mathcal T$; the direct treatment-plan decoder is a separate module.

Both attention paths use the causal mask $M$, with $M_{tj}=0$ for $j\leq t$ and $M_{tj}=-\infty$ otherwise. Here $t$ indexes query positions and $j$ indexes key positions. The key-padding mask $M_{\mathrm{pad}}$ excludes inactive positions. Since position $j$ contains inputs aligned with time $j-1$, the output at position $t$ depends only on observations available through time $t-1$.

At the first sequence position, $X_0$ is an all-zero start placeholder. The initial outcome input $Y_0$ is zero for CVSim and the standardized observed initial tumor volume for CancerSim. The initial historical encoding $c_0$ is set directly to zero in the encoding space before applying $W_c$ and $W_m$. The current encoding $c_t$ remains outside this shifted stream and is fused with $h_{t-1}$ after temporal processing.
\subsection*{HiSTOR Optimization Details}

\paragraph{Auxiliary treatment prediction.}
Let \(\mathcal K\) index the treatment categories, and let
\[
p_{it}^X=\operatorname{softmax}(\ell_{it}^X)
\]
denote the auxiliary treatment-probability vector for patient \(i\) at time \(t\). Its \(k\)-th element, \(p_{it,k}^X\), is the predicted probability of treatment category \(k\). For each minibatch, let $\mathcal I_b$ denote the valid patient--time
observations. The auxiliary treatment loss is
\begin{equation}
 \mathcal L_X
 =-\frac{1}{|\mathcal I_b|}
 \sum_{(i,t)\in\mathcal I_b}
 \sum_{k\in\mathcal K}
 \mathbf 1\{X_{it}=k\}
 \log p_{it,k}^X,
 \label{eq:treatmentloss}
\end{equation}
Both simulation datasets encode each treatment choice as one categorical label; in CancerSim, the category representing both therapies is treated as a single treatment class. The auxiliary treatment loss is not weighted by treatment frequency, and its contribution to the backbone objective is scaled by \(\lambda_X\).

The complete backbone objective is given in
Equation~\ref{eq:objective}, where $e$ is the training epoch
and the $\lambda$ coefficients weight the corresponding objectives.

\paragraph{Regularization and updates.}
The backbone uses $(\lambda_X,\lambda_{\mathrm{roll}})=(0.01,0.25)$ for CVSim and $(1,0)$ for CancerSim. The latent-norm coefficient is $\lambda_{\mathrm{norm}}=0.01$.

The encoding critic $D_c$ compares the current-state encoding $c_t$ with standard-normal samples $\epsilon\sim\mathcal N(0,I_{d_c})$, where $I_{d_c}$ is the $d_c$-dimensional identity matrix. The embedding critic $D_u$ compares generated embeddings $G_V(\epsilon)$ with observed embeddings $u_t$. The encoder and generator minimize
\begin{equation}
 \mathcal L_{E,\mathrm{adv}}
 =-\mathbb E[D_c(c_t)],
 \qquad
 \mathcal L_{G,\mathrm{adv}}
 =-\mathbb E[D_u(G_V(\epsilon))].
 \label{eq:adv}
\end{equation}

The adversarial term in Equation~\ref{eq:objective} is the sum of the
encoder and generator objectives:
\begin{equation}
 \mathcal L_{\mathrm{adv}}
 =
 \mathcal L_{E,\mathrm{adv}}
 +
 \mathcal L_{G,\mathrm{adv}}.
\end{equation}

These distributional objectives act on the first-level encoding and embedding spaces. The critics use Wasserstein losses with gradient penalties \citep{gulrajani2017improved} and jointly minimize
\begin{equation}
\begin{aligned}
 \mathcal L_D={}&
 \mathbb E_{\mathcal I}[D_c(c_t)]
 -\mathbb E_\epsilon[D_c(\epsilon)]\\
 &+\mathbb E_\epsilon[D_u(G_V(\epsilon))]
 -\mathbb E_{\mathcal I}[D_u(u_t)]\\
 &+\lambda_{\mathrm{gp}}
 \left(\mathcal R_{\mathrm{gp}}^c+\mathcal R_{\mathrm{gp}}^u\right).
\end{aligned}
\label{eq:critic}
\end{equation}
During critic updates, the current-state representations and generated samples are detached from the remaining network. For each critic, the gradient penalty is
\begin{equation}
 \mathcal R_{\mathrm{gp}}(D)
 =
 \mathbb E_{\widetilde v}
 \left[
 \left(
 \|\nabla_{\widetilde v}D(\widetilde v)\|_2-1
 \right)^2
 \right],
 \label{eq:gp}
\end{equation}
where
\[
\widetilde v
=
\alpha v_{\mathrm{ref}}
+(1-\alpha)v_{\mathrm{model}},
\qquad
\alpha\sim\mathcal U(0,1).
\]
The paired reference and model inputs are \((\epsilon,c_t)\) for \(D_c\) and \((u_t,G_V(\epsilon))\) for \(D_u\). We use \(\lambda_{\mathrm{gp}}=10\) and perform three critic updates for each backbone update.

The latent-norm penalty is
\begin{equation}
 \mathcal L_{\mathrm{norm}}
 =
 \mathbb E_{\mathcal I}
 \left[
 \left(
 \frac{\|c_t\|_2^2}{d_c}-1
 \right)^2
 \right],
 \label{eq:norm}
\end{equation}
with \(\lambda_{\mathrm{norm}}=0.01\). This term regulates the normalized latent scale around one, while distribution matching is handled by the adversarial objective. With zero-based epoch index \(e\), the adversarial coefficient is
\begin{equation}
 \lambda_{\mathrm{adv}}(e)
 =
 0.01
 \min\!\left(
 1,
 \max\!\left(
 0,\frac{e-10}{20}
 \right)
 \right).
\end{equation}

The CVSim rollout coefficient remains \(0.25\) throughout training, without an additional warm-up schedule. The CVSim backbone rollout horizon is five steps. Training samples one origin uniformly from the complete valid five-step windows of each eligible trajectory. At rollout offset \(j\), the current- and previous-treatment input arrays are masked at positions later than \(t+j\), so later treatment entries are not exposed to earlier predictions. Generated outcomes remain connected to the computational graph when they are inserted into subsequent history inputs. Validation uses one deterministically selected middle valid origin for each eligible trajectory.

For CVSim, let \(\rho_k\) denote the empirical proportion of treatment category \(k\) in the training data. The treatment weight for category \(k\) is computed once as
\begin{equation}
 w_k
 =
 \frac{\rho_k^{-1/2}}
 {\sum_{\ell\in\mathcal K}\rho_\ell\rho_\ell^{-1/2}}.
\end{equation}
These weights are used for the backbone one-step outcome loss and for the direct decoder training and validation losses. They are not used for the auxiliary treatment loss, the backbone rollout loss, or the backbone validation criterion. CancerSim uses unit weights throughout.

\paragraph{Outcome and rollout losses.}
Let $w_{X_{it}}$ be the weight of the observed treatment category. The factual one-step outcome loss is
\begin{equation}
 \mathcal L_Y
 =
 \frac{
 \sum_{(i,t)\in\mathcal I_b}
 w_{X_{it}}
 \left\|\widehat y_{it}(X_{it})-Y_{it}\right\|_2^2
 }{
 d_y\sum_{(i,t)\in\mathcal I_b}w_{X_{it}}
 },
     \label{eq:outcomeloss}
    \end{equation}
where $d_y$ is the outcome dimension. CVSim uses the training-derived
treatment weights defined above, whereas CancerSim uses unit weights.

For CVSim, let $\mathcal W_b^{(5)}$ be the sampled valid five-step factual
windows in a minibatch. The rollout prediction
$\widehat y^{\mathrm{roll}}_{i,t+j}$ is generated recursively, with each
prediction supplied to the subsequent history. The rollout loss is
\begin{equation}
 \mathcal L_{\mathrm{roll}}
 =
 \frac{
 \sum_{(i,t)\in\mathcal W_b^{(5)}}
 \sum_{j=0}^{4}
 \left\|\widehat y^{\mathrm{roll}}_{i,t+j}-Y_{i,t+j}\right\|_2^2
 }{
 5d_y\,|\mathcal W_b^{(5)}|
 },
 \label{eq:rolloutloss}
\end{equation}
and does not use treatment weights. CancerSim sets
$\lambda_{\mathrm{roll}}=0$.

\paragraph{Direct-decoder architecture.}
At the prediction origin, the decoder forms
\[
 \beta_t=\tanh\!\left(W_b[Z_t,s_i^{\mathrm{stat}},\widetilde y_{t-1}]\right).
\]
Here $W_b$ is learned and $\widetilde y_{t-1}$ is the dataset-specific
previous-outcome input. Here $s_i^{\mathrm{stat}}$ denotes the static
decoder input: patient-factor features in CVSim and the patient-type
feature in CancerSim. The main-text expression
$D_\tau(Z_t,\bar x)$ is a high-level shorthand; the implementation also
supplies $s_i^{\mathrm{stat}}$ and the prepared previous-outcome input
$\widetilde y_{t-1}$ at the prediction boundary.
A shared treatment-plan GRU and horizon-specific heads then produce
\[
 \eta_k=\operatorname{GRU}(E_X(x_{t+k}),\eta_{k-1}),\qquad
 \widehat y_{t+k}^{\mathrm D}=g_k([\beta_t,\eta_k]),
 \quad \eta_{-1}=\beta_t,
\]
for $k=0,\ldots,\tau-1$. Here $E_X$ is the decoder treatment embedding and $g_k$ is the outcome head at offset $k$. Generated outcomes are not fed between heads.
\paragraph{Direct-decoder objective.}
After backbone selection, the exponential-moving-average backbone is frozen and used to extract decoder contexts without gradients. Let $\mathcal W$ be a minibatch of complete factual windows indexed by patient $i$ and prediction origin $t$, let $\tau$ be the number of target steps, and let $d_y$ be the outcome dimension. The direct decoder minimizes
\begin{equation}
 \mathcal L_{\mathrm{dir}}
 =
 \frac{
 \sum_{(i,t)\in\mathcal W}\sum_{j=0}^{\tau-1}
 w_{X_{i,t+j}}
 \|\widehat y^{\mathrm D}_{i,t+j}-Y_{i,t+j}\|_2^2
 }{
 d_y\sum_{(i,t)\in\mathcal W}\sum_{j=0}^{\tau-1}
 w_{X_{i,t+j}}
 }.
 \label{eq:directloss}
\end{equation}
Here $\widehat y^{\mathrm D}_{i,t+j}$ is the direct prediction at horizon offset $j$, and $w_{X_{i,t+j}}$ is the weight assigned to the observed treatment category at that step. CVSim uses the training-derived treatment weights defined above, whereas CancerSim uses unit weights. Decoder checkpoints are selected using the corresponding factual validation objective.

Backbone training uses AdamW with separate learning-rate groups for the current-state autoencoder and the remaining non-critic modules, together with a separate optimizer for the critics. Gradients are clipped to norm \(1\). An exponential moving average with decay \(0.99\) is maintained for the backbone and is used for validation, one-step prediction, and decoder-context extraction. Decoder contexts are extracted without gradients after backbone selection.

The direct decoder uses AdamW with learning rate \(10^{-3}\), weight decay
\(10^{-3}\), gradient clipping at norm \(1\), no parameter averaging,
hidden width \(32\), batch size \(4096\), early-stopping patience \(20\),
and a maximum training budget of \(200\) epochs. Decoder windows contain only valid targets and exclude the first sequence position, which lacks a
genuine preceding outcome. The decoder uses all four latent branches of
\(Z_t\); its treatment GRU shares parameters across horizons, while each
horizon has an independent outcome head.

\section{Data and Evaluation Protocols}\label{app:data}
\subsection*{CVSim}
CVSim is a stochastic longitudinal dynamical system in which a
six-dimensional state evolves with the preceding categorical treatment and a three-dimensional observed patient-factor vector. Four noisy state projections form the base time-varying covariates. Treatment is sampled from a three-category softmax mechanism depending on the covariates and patient factor. The scalar outcome combines state and covariate contributions with treatment main effects and state-dependent treatment interactions. The observed configuration supplies the patient factor as static input.

For the current-state projection, CVSim directly uses the observed covariate vector. Under the implementation-only notation introduced in Appendix~\ref{app:notation},
\[
W_{it}=V_{it}.
\]
Thus, $d_W=7+d_{\mathrm{hd}}$, consisting of the four base time-varying covariates, the three observed patient factors supplied as static inputs, and any additional dimensions. The preceding-outcome channel is initialized to zero at the sequence start.

The additional dimensions are not pure-noise features or separate latent
patient factors. Each dimension is generated as a noisy linear measurement
of the same three-dimensional observed patient-factor vector. These
measurements are added after the base trajectories have been generated and
therefore do not directly alter the base treatment-assignment or outcome
equations.

Writing $q_i\in\mathbb R^3$ for the observed patient-factor vector and
$W_{\mathrm{hd}}\in\mathbb R^{d_{\mathrm{hd}}\times3}$ for a projection
matrix sampled once for each simulated dataset, the extensions are
\[
 V_i^{\mathrm{hd}}
 =W_{\mathrm{hd}}q_i+\epsilon_i
 \quad\text{(Static)},\qquad
 V_{i,t}^{\mathrm{hd}}
 =W_{\mathrm{hd}}q_i
 +\frac{t}{T-1}\mathbf 1_{d_{\mathrm{hd}}}
 +\epsilon_{i,t}
 \quad\text{(Temporal)}.
\]
The entries of $W_{\mathrm{hd}}$ are drawn independently from
$\mathcal N(0,0.5^2)$, and the entries of the measurement-noise vectors are
drawn independently from $\mathcal N(0,0.3^2)$. Thus, conditional on $q_i$,
the noise components are independent, but the complete added measurements
are not generally independent because they share the signal
$W_{\mathrm{hd}}q_i$. In the Static setting, $\epsilon_i$ is sampled once
per patient and the resulting measurements are held fixed over time. In the
Temporal setting, $\epsilon_{i,t}$ is resampled independently for each time
point, and a common linear drift is added.

Consequently, increasing $d_{\mathrm{hd}}$ increases the number of noisy,
redundant observed measurements rather than the number of underlying patient
factors; the patient-factor dimension remains three. The two extension modes
coincide when $d_{\mathrm{hd}}=0$. The simulator generates potential outcomes
for all three treatments at every time point. For multi-step evaluation, it
starts from the factual state at the forecast origin and reuses the exogenous
noise under the fixed treatment plan. The resulting future covariates are not
supplied to the prediction models.

\begin{table}[htbp]
\caption{CVSim simulation and evaluation parameters used in the experiments.}
\label{tab:cvsim_parameters}
\centering\small
\begin{tabularx}{\linewidth}{@{}lX@{}}
\toprule
Parameter & Value \\
\midrule
Patients; train/validation/test & 5,000; 4,000/500/500 \\
Trajectory length; state dimension & 50; 6 \\
Base time-varying/static covariates & 4/3 \\
Treatment categories & 0, 1, 2, encoded as three-way one-hot vectors \\
Integration step; confounding coefficient & 0.1; 1.0 \\
State/covariate/outcome noise SD & 0.05/0.05/0.05 \\
Treatment main/interaction coefficient SD & 0.20/0.10 \\
Added dimensions $d_{\mathrm{hd}}$ & 0, 10, 20, 50, 100, 200, 500, 1,000, 2,000 \\
Extension modes & Static and temporal, both with observed patient factors \\
Extension noise SD; temporal drift & 0.3; linear drift from 0 to 1 \\
Base-covariate temporal drift & Disabled; separate from the temporal extension \\
Total raw covariate width & $7+d_{\mathrm{hd}}$, including static inputs \\
Forecast origin (zero-based array index) & $s=50-5=45$ \\
Fixed treatment; scored horizons & Arm 2, $(0,0,1)$; $\tau=1,\ldots,5$ \\
\bottomrule
\end{tabularx}
\end{table}

\subsection*{CancerSim}
CancerSim simulates tumor volume $Y_{it}$ using Gompertz-type growth, chemotherapy concentration, linear and quadratic radiotherapy effects, and multiplicative noise. The simulator has no separate time-varying covariate channel beyond the tumor-volume trajectory. The patient-type feature $S_i$ is a baseline covariate that modifies the treatment-response coefficients. Accordingly, the CancerSim covariate variable in the general formulation contains the baseline component $S_i$, while tumor volume remains the outcome $Y_{it}$.

Treatment probabilities at time $t$ depend on the recent mean tumor diameter computed from the observed outcome history $Y_{i,<t}$. Past outcomes therefore belong to the observed pre-treatment history and provide time-varying confounding information, while retaining their definition as outcomes.

For the implemented current-state projection, the raw input is constructed as
\[
W_{it}=[Y_{i,t-1},S_i].
\]
Here $W_{it}$ is an implementation-only input tensor; this construction does not redefine the outcome $Y_{i,t-1}$ as a covariate. The first input uses the observed initial tumor volume after dataset-level standardization, and the static patient-type feature is repeated across time. Hence $d_W=2$, and the projection expands this two-dimensional input to the 16-dimensional current-state encoding $c_t$. The initial historical outcome input is the standardized baseline tumor volume.

The training, validation, one-step counterfactual, and multi-step counterfactual cohorts are generated separately. Counterfactual expansion produces multiple evaluation entries per patient; these entries must not be counted as independent patients. The one-step test includes factual prefixes and alternative-treatment branches, and its metric scores all active entries rather than only the final counterfactual position of each branch.

\begin{table}[htbp]
\caption{CancerSim simulation and evaluation parameters used in the experiments. Sampling means and standard deviations refer to the parameter distributions before truncation.}
\label{tab:cancer_parameters}
\centering\small
\begin{tabularx}{\linewidth}{@{}lX@{}}
\toprule
Parameter & Value \\
\midrule
Training/validation patients & 10,000/1,000 \\
Patients per counterfactual test cohort & 1,000 for each separately generated test \\
Maximum simulation length & 60; at most 59 shifted factual prediction slots \\
Treatment categories & None, chemotherapy, radiotherapy, both \\
Static covariate & Patient type, sampled uniformly from $\{1,2,3\}$ \\
Chemo/radio assignment coefficients & Both equal $\gamma\in\{0,1,2,3,4\}$ \\
Assignment window; lag & 15; 0 \\
Logistic slope; diameter midpoint & $\gamma/13$; 6.5 \\
Chemotherapy dose; half-life & 5; 1 \\
Radiotherapy dose; noise SD & 2; 0.01 \\
Carrying capacity; death threshold & $\pi30^3/6$; $D_C=\pi13^3/6$ \\
Growth-rate sampling mean/SD & $7\times10^{-5}$/$7.23\times10^{-3}$ \\
Radiotherapy $\alpha$ sampling mean/SD & 0.0398/0.168 \\
Chemotherapy coefficient mean/SD & 0.028/0.0007 \\
$\alpha$--growth correlation; $\alpha/\beta$ & 0.87; 10 \\
Patient-type adjustments & Type 1: add 0.00398 to $\alpha$; type 3: add 0.0028 to the chemotherapy coefficient \\
Recovery probability & $\exp(-5.8\times10^8\,\mathrm{volume})$ \\
Fixed treatment; scored horizons & Both therapies, $(0,0,0,1)$; $\tau=2,\ldots,6$ \\
\bottomrule
\end{tabularx}
\end{table}

Initial diameters follow a truncated log-normal stage mixture. For stages I, II, IIIA, IIIB, and IV, the log-scale $(\mu,\sigma)$ pairs are $(1.72,4.70)$, $(1.96,1.63)$, $(1.91,9.40)$, $(2.76,6.87)$, and $(3.86,8.82)$, with sampling weights 1432, 128, 1306, 7248, and 12840. The lower diameter bound is 0.3; the upper bound is 5 for stage I and 13 otherwise. Diameters are converted to volumes. Factual trajectories stop at the death threshold, stochastic recovery, or the maximum length. Outcome and patient-type scaling statistics are fitted on training data; treatment indicators retain their categorical encoding.

\section{Forecast Boundary and Model Adaptations}\label{app:forecast}
For fixed-regime multi-step evaluation in CVSim, $Y_{s-1}$ and $V_s$ remain observed, later previous-outcome slots are cleared, and future covariate placeholders repeat $V_s$. Models may mask, replace, or bypass these placeholders. In CancerSim, the boundary outcome is first predicted under the factual boundary treatment and inserted into the subsequent history. This bridge is not scored; the five fixed-plan targets correspond to horizons two through six. Scoring uses each model's multi-step output directly, without replacing its first scored entry with a separate one-step prediction.

\begin{table}[htbp]
\caption{Future-covariate handling within the common forecast boundary.}
\label{tab:future_covariates}
\centering\small
\begin{tabularx}{\linewidth}{@{}lX@{}}
\toprule
Method & Multi-step inputs \\
\midrule
HiSTOR & Direct decoding uses the boundary representation, static covariates, previous-outcome input, and treatment plan, without future covariates or outcome feedback between horizon heads. CancerSim first supplies a predicted bridge outcome at the boundary. \\
CT \citep{melnychuk2022causal} & Masks access to post-origin covariates and recursively updates outcomes. \\
CRN, RMSN & Decode from the encoded history using treatment, static information, and predicted outcomes; decoders do not consume future covariates. \\
MSM \citep{robins2000marginal} & Uses origin-history features and cumulative treatment-plan features, without future-covariate simulation. \\
Causal CPC \citep{bouchattaoui2024causal} & Decodes from origin context and the treatment plan, without future covariates. \\
CTD-NKO \citep{wang2025enhancing} & Recursively predicts outcomes and subsequent covariates, replacing placeholders. \\
IGC-Net \citep{hess2026igc} & Uses origin history and separately fitted fixed-prefix regressions, without sampling future covariate distributions. \\
\bottomrule
\end{tabularx}
\end{table}

\section{Metric Definitions and Aggregation}\label{app:metrics}
We report normalized root mean squared error (NRMSE); lower values are better. The normalization constants are defined without using model predictions: CancerSim uses the simulator's fixed tumor-death volume threshold, whereas CVSim uses ranges computed from the corresponding test targets. For CancerSim one-step prediction, let $m_{it}$ mark an active expanded test entry and let $\mathcal T_+=\{t:\sum_i m_{it}>0\}$. The reported error is
\begin{equation}
E_{C,1}=\frac{100}{D_C}\sqrt{\frac{1}{|\mathcal T_+|}
\sum_{t\in\mathcal T_+}\frac{\sum_i m_{it}(\widehat y_{it}-y_{it})^2}{\sum_i m_{it}}}.
\end{equation}
Here $i$ indexes counterfactually expanded test entries and $D_C=\pi13^3/6\approx1150.3465$ is the simulator's tumor-death threshold. Thus CancerSim first averages squared error across active entries at each time and then averages across valid times; it does not pool all active entries before taking the square root.

For CVSim one-step prediction, let $\mathcal I_1$ contain all active patient--time--treatment entries. With $D_{V,1}$ equal to the range of their target potential outcomes,
\begin{equation}
E_{V,1}=\frac{100}{\max(D_{V,1},10^{-8})}
\sqrt{\frac{1}{|\mathcal I_1|}\sum_{(i,t,a)\in\mathcal I_1}
(\widehat y_{it}(a)-y_{it}(a))^2}.
\end{equation}
This definition pools the active test units, time points, and all three current treatment arms.

For scored multi-step position $h$, let $\mathcal I_h$ contain the entries with finite predictions and targets. The horizon-specific error is
\begin{equation}
E_h=\frac{100}{\max(D,10^{-8})}\sqrt{\frac{1}{|\mathcal I_h|}
\sum_{i\in\mathcal I_h}(\widehat y_{i,h}-y_{i,h})^2}.
\end{equation}
CancerSim uses $D=D_C$ and reports horizons $2$ through $6$ after the unscored bridge prediction. CVSim uses the range of all fixed-regime multi-step targets across patients and the five scored positions, shared across horizons within the same seed and condition, and reports horizons $1$ through $5$. This rollout denominator differs from $D_{V,1}$ because the one-step and multi-step evaluations use different target arrays. With the same simulator seed and patient split, adding high-dimensional measurements leaves the outcome trajectories unchanged, so the CVSim rollout denominator is constant across dimensions within a seed but may differ across seeds.

Within seed $r$, the horizon average is $\overline E^{(r)}=\frac15\sum_{h=1}^5 E_h^{(r)}$. For either a one-step score, a horizon-specific score, or this average, we report the mean and sample standard deviation across seeds $\{3,101,42,8,9\}$, using denominator $R-1$ with $R=5$. No errors are pooled across datasets.

\section{Baselines and Training Configurations}\label{app:baselines}
The comparison uses HiSTOR and the seven baselines listed in Section~\ref{sec:experimental_setup}. All methods receive the same simulated data realization, split, intervention regime, forecast boundary, and seed within each experimental condition. CancerSim uses 10,000 training and 1,000 validation patients, together with separate 1,000-patient one-step and multi-step test cohorts. CVSim uses an 80/10/10 patient split of 5,000 trajectories. Counterfactual test targets are used only for evaluation.

CRN uses an encoder LSTM with hidden width 24, balancing width 12, and head width 36, followed by a one-layer decoder with balancing width 18 and head width 36. Its encoder and decoder use Adam with learning rates $10^{-2}$ and $10^{-3}$ and batch sizes 128 and 1,024, respectively. CT uses one Transformer layer, two heads, dropout 0.1, and EMA decay 0.99; its CancerSim sequence/balancing/head widths vary across dataset-specific configurations, while CVSim uses widths 16/12/24. RMSN retains its four propensity, history, encoder, and decoder stages with dataset-specific widths, learning rates, batch sizes, and clipping thresholds.

Causal CPC uses dataset-specific local, context, and head widths with six-step contrastive prediction and an autoregressive treatment-plan decoder; its representation and outcome modules use AdamW and its treatment adversary uses SGD. CTD-NKO uses condition-specific CancerSim widths and a 24-dimensional CVSim model, with Adam, EMA, and validation-based selection. IGC-Net uses a 16/8/4 CancerSim hidden/history/head configuration and a 48/88/44 CVSim configuration; multi-step prediction fits a separate fixed-regime model for each required treatment-plan prefix. MSM uses unregularized multinomial logistic propensity models, stabilized weights clipped at the first and 99th percentiles, and weighted linear outcome regressions.

HiSTOR uses a 16-dimensional current-covariate encoding, a 128-dimensional four-layer, four-head Transformer with feed-forward width 256, and a 160-dimensional structured representation split into widths 24/24/48/64. The backbone uses AdamW with learning rates $10^{-3}$ for the main generator, $5\times10^{-4}$ for the autoencoder modules, and $10^{-4}$ for the critics, together with gradient-penalty weight 10, three critic updates, and EMA decay 0.99. The direct treatment-plan decoder has hidden width 32, uses all four representation branches, and is trained with AdamW at learning rate $10^{-3}$, weight decay $10^{-3}$, batch size 4,096, and gradient clipping at 1. The selected backbone remains frozen during decoder training.

\begin{table}[htbp]
\caption{Training budgets and validation criteria.}
\label{tab:training_rules}
\centering\small
\begin{tabularx}{\linewidth}{@{}>{\raggedright\arraybackslash}p{0.19\linewidth}>{\raggedright\arraybackslash}X>{\raggedright\arraybackslash}X@{}}
\toprule
Method & Maximum budget & Validation and checkpoint selection \\
\midrule
CRN \citep{bica2020estimating} & 100 epochs per encoder/decoder stage & Fixed budget; no best-validation restoration \\
CT \citep{melnychuk2022causal} & 150 epochs & Fixed budget; final EMA parameters \\
RMSN \citep{lim2018forecasting} & 100 epochs per propensity/prediction stage & Fixed budget; no best-validation restoration \\
HiSTOR (backbone) & CancerSim: 50 epochs; CVSim: 120 epochs & EMA validation. CancerSim: one-step MSE; CVSim: one-step MSE plus $0.25$ times rollout MSE. Patience: 12/25, respectively. \\
HiSTOR (decoder) & 200 epochs & Factual multi-step loss in Equation~\ref{eq:directloss}; no EMA; patience 20. \\
MSM \citep{robins2000marginal} & Weighted regression & Training-set fit; no validation checkpoint \\
Causal CPC \citep{bouchattaoui2024causal} & 2,000 epochs per stage & Contrastive pretraining and outcome fine-tuning objectives; patience 30 \\
CTD-NKO \citep{wang2025enhancing} & CancerSim: 150; CVSim: 500 & Weighted outcome, covariate, and Koopman validation losses with EMA; patience 30 \\
IGC-Net \citep{hess2026igc} & CancerSim: 50; CVSim: 100 & Factual MSE for the one-step head; pseudo-outcome validation for each fixed prefix; patience 30 \\
\bottomrule
\end{tabularx}
\end{table}

CRN, CT, and RMSN use fixed training budgets, while HiSTOR, Causal CPC,
CTD-NKO, and IGC-Net use the validation and checkpoint-selection criteria
reported in Table~\ref{tab:training_rules}. All stochastic experiments use
the five reported evaluation seeds.

\section{Detailed CancerSim Results}\label{app:cancer_detailed}

Values are NRMSE, mean $\pm$ sample SD across five randomly selected seeds $\{3,101,42,8,9\}$. Lower is better; the lowest mean is bold and the second-lowest underlined. Figure~\ref{fig:cancersim_horizons} summarizes the horizon-specific mean results, followed by the corresponding mean and sample-SD tables and paired seed-level comparisons.

\begin{table}[h]
\caption{CancerSim: $\tau$=2. Values are NRMSE, mean $\pm$ sample SD across five randomly selected seeds $\{3,101,42,8,9\}$. Lower is better; the lowest mean is bold and the second-lowest underlined.}
\label{tab:cancersim_tau2_0}
\centering\footnotesize
\setlength{\tabcolsep}{3pt}
\resizebox{\linewidth}{!}{%
\begin{tabular}{@{}lccccc@{}}
\toprule
Method & $\gamma$=0 & $\gamma$=1 & $\gamma$=2 & $\gamma$=3 & $\gamma$=4 \\
\midrule
MSM \citep{robins2000marginal} & 1.2741 $\pm$ 0.1100 & 1.1635 $\pm$ 0.0900 & 1.2066 $\pm$ 0.0585 & 1.4065 $\pm$ 0.1020 & 1.8154 $\pm$ 0.1062 \\
RMSN \citep{lim2018forecasting} & 0.9211 $\pm$ 0.1186 & 0.9329 $\pm$ 0.0732 & 1.0331 $\pm$ 0.0750 & 1.1923 $\pm$ 0.1337 & 1.5835 $\pm$ 0.1700 \\
CRN \citep{bica2020estimating} & 0.9276 $\pm$ 0.0552 & 0.9578 $\pm$ 0.0771 & 1.0822 $\pm$ 0.0609 & 1.1791 $\pm$ 0.0744 & \underline{1.4956 $\pm$ 0.1417} \\
CT \citep{melnychuk2022causal} & \underline{0.8526 $\pm$ 0.0666} & 0.8956 $\pm$ 0.0510 & 1.0073 $\pm$ 0.0781 & 1.2217 $\pm$ 0.1078 & 1.5822 $\pm$ 0.1954 \\
Causal CPC \citep{bouchattaoui2024causal} & 0.8707 $\pm$ 0.0642 & 0.9144 $\pm$ 0.0661 & 1.0265 $\pm$ 0.0297 & 1.1786 $\pm$ 0.0823 & 1.5764 $\pm$ 0.0844 \\
CTD-NKO \citep{wang2025enhancing} & 0.8988 $\pm$ 0.0903 & \underline{0.8836 $\pm$ 0.0466} & \underline{0.9827 $\pm$ 0.0878} & \underline{1.1608 $\pm$ 0.1459} & 1.5919 $\pm$ 0.1812 \\
IGC-Net \citep{hess2026igc} & 0.9028 $\pm$ 0.0810 & 0.9437 $\pm$ 0.0800 & 1.0616 $\pm$ 0.0712 & 1.2969 $\pm$ 0.1968 & 1.5079 $\pm$ 0.1704 \\
\midrule
HiSTOR & \textbf{0.8164 $\pm$ 0.0471} & \textbf{0.8359 $\pm$ 0.0455} & \textbf{0.9313 $\pm$ 0.0678} & \textbf{1.0616 $\pm$ 0.0696} & \textbf{1.3023 $\pm$ 0.0403} \\
\bottomrule
\end{tabular}}
\end{table}

\begin{table}[h]
\caption{CancerSim: $\tau$=3. Values are NRMSE, mean $\pm$ sample SD across five randomly selected seeds $\{3,101,42,8,9\}$. Lower is better; the lowest mean is bold and the second-lowest underlined.}
\label{tab:cancersim_tau3_0}
\centering\footnotesize
\setlength{\tabcolsep}{3pt}
\resizebox{\linewidth}{!}{%
\begin{tabular}{@{}lccccc@{}}
\toprule
Method & $\gamma$=0 & $\gamma$=1 & $\gamma$=2 & $\gamma$=3 & $\gamma$=4 \\
\midrule
MSM \citep{robins2000marginal} & 1.2151 $\pm$ 0.0802 & 1.0676 $\pm$ 0.0612 & 1.0760 $\pm$ 0.0378 & 1.2109 $\pm$ 0.0858 & 1.4938 $\pm$ 0.0915 \\
RMSN \citep{lim2018forecasting} & 0.7531 $\pm$ 0.0863 & 0.7798 $\pm$ 0.0677 & 0.8688 $\pm$ 0.0547 & 1.0709 $\pm$ 0.0985 & 1.4446 $\pm$ 0.1342 \\
CRN \citep{bica2020estimating} & 0.7207 $\pm$ 0.0200 & 0.7605 $\pm$ 0.0470 & 0.8918 $\pm$ 0.0493 & 1.0124 $\pm$ 0.0468 & \underline{1.3358 $\pm$ 0.1304} \\
CT \citep{melnychuk2022causal} & 0.7039 $\pm$ 0.0512 & 0.7565 $\pm$ 0.0371 & 0.8613 $\pm$ 0.0584 & 1.1286 $\pm$ 0.0820 & 1.4427 $\pm$ 0.1822 \\
Causal CPC \citep{bouchattaoui2024causal} & \underline{0.6892 $\pm$ 0.0442} & 0.7593 $\pm$ 0.0648 & 0.8426 $\pm$ 0.0366 & \underline{1.0072 $\pm$ 0.0809} & 1.3924 $\pm$ 0.0735 \\
CTD-NKO \citep{wang2025enhancing} & 0.7168 $\pm$ 0.0679 & \underline{0.7402 $\pm$ 0.0398} & \underline{0.8424 $\pm$ 0.0726} & 1.0433 $\pm$ 0.1289 & 1.5613 $\pm$ 0.1790 \\
IGC-Net \citep{hess2026igc} & 0.7690 $\pm$ 0.0796 & 0.8233 $\pm$ 0.0411 & 0.8910 $\pm$ 0.0592 & 1.0994 $\pm$ 0.1694 & 1.3834 $\pm$ 0.1194 \\
\midrule
HiSTOR & \textbf{0.6573 $\pm$ 0.0302} & \textbf{0.6864 $\pm$ 0.0209} & \textbf{0.7762 $\pm$ 0.0521} & \textbf{0.9105 $\pm$ 0.0667} & \textbf{1.1462 $\pm$ 0.0310} \\
\bottomrule
\end{tabular}}
\end{table}

\begin{table}[h]
\caption{CancerSim: $\tau$=4. Values are NRMSE, mean $\pm$ sample SD across five randomly selected seeds $\{3,101,42,8,9\}$. Lower is better; the lowest mean is bold and the second-lowest underlined.}
\label{tab:cancersim_tau4_0}
\centering\footnotesize
\setlength{\tabcolsep}{3pt}
\resizebox{\linewidth}{!}{%
\begin{tabular}{@{}lccccc@{}}
\toprule
Method & $\gamma$=0 & $\gamma$=1 & $\gamma$=2 & $\gamma$=3 & $\gamma$=4 \\
\midrule
MSM \citep{robins2000marginal} & 1.0396 $\pm$ 0.0558 & 0.8940 $\pm$ 0.0395 & 0.8815 $\pm$ 0.0251 & 0.9675 $\pm$ 0.0506 & 1.1679 $\pm$ 0.0738 \\
RMSN \citep{lim2018forecasting} & 0.5467 $\pm$ 0.0640 & 0.5764 $\pm$ 0.0653 & 0.6356 $\pm$ 0.0360 & 0.8279 $\pm$ 0.0876 & 1.1093 $\pm$ 0.0867 \\
CRN \citep{bica2020estimating} & 0.5204 $\pm$ 0.0256 & 0.5637 $\pm$ 0.0280 & 0.6723 $\pm$ 0.0483 & 0.7932 $\pm$ 0.0311 & 1.0614 $\pm$ 0.0893 \\
CT \citep{melnychuk2022causal} & 0.5144 $\pm$ 0.0404 & 0.5589 $\pm$ 0.0314 & 0.6427 $\pm$ 0.0474 & 0.8771 $\pm$ 0.0610 & 1.1302 $\pm$ 0.2003 \\
Causal CPC \citep{bouchattaoui2024causal} & \underline{0.5041 $\pm$ 0.0368} & 0.5608 $\pm$ 0.0545 & \underline{0.6149 $\pm$ 0.0279} & \underline{0.7665 $\pm$ 0.0606} & \underline{1.0591 $\pm$ 0.0822} \\
CTD-NKO \citep{wang2025enhancing} & 0.5135 $\pm$ 0.0504 & \underline{0.5486 $\pm$ 0.0352} & 0.6347 $\pm$ 0.0528 & 0.7999 $\pm$ 0.1004 & 1.2832 $\pm$ 0.1673 \\
IGC-Net \citep{hess2026igc} & 0.6253 $\pm$ 0.0590 & 0.6709 $\pm$ 0.0632 & 0.7317 $\pm$ 0.0957 & 0.9435 $\pm$ 0.1485 & 1.1810 $\pm$ 0.1571 \\
\midrule
HiSTOR & \textbf{0.4716 $\pm$ 0.0209} & \textbf{0.5013 $\pm$ 0.0207} & \textbf{0.5654 $\pm$ 0.0343} & \textbf{0.6784 $\pm$ 0.0509} & \textbf{0.8772 $\pm$ 0.0401} \\
\bottomrule
\end{tabular}}
\end{table}

\begin{table}[h]
\caption{CancerSim: $\tau$=5. Values are NRMSE, mean $\pm$ sample SD across five randomly selected seeds $\{3,101,42,8,9\}$. Lower is better; the lowest mean is bold and the second-lowest underlined.}
\label{tab:cancersim_tau5_0}
\centering\footnotesize
\setlength{\tabcolsep}{3pt}
\resizebox{\linewidth}{!}{%
\begin{tabular}{@{}lccccc@{}}
\toprule
Method & $\gamma$=0 & $\gamma$=1 & $\gamma$=2 & $\gamma$=3 & $\gamma$=4 \\
\midrule
MSM \citep{robins2000marginal} & 0.8566 $\pm$ 0.0448 & 0.7227 $\pm$ 0.0299 & 0.6987 $\pm$ 0.0291 & 0.7520 $\pm$ 0.0293 & 0.8948 $\pm$ 0.0575 \\
RMSN \citep{lim2018forecasting} & 0.3820 $\pm$ 0.0486 & 0.4087 $\pm$ 0.0528 & 0.4453 $\pm$ 0.0248 & 0.6126 $\pm$ 0.0932 & 0.7896 $\pm$ 0.0523 \\
CRN \citep{bica2020estimating} & 0.3643 $\pm$ 0.0262 & 0.4064 $\pm$ 0.0229 & 0.4838 $\pm$ 0.0341 & 0.6091 $\pm$ 0.0470 & 0.8571 $\pm$ 0.0685 \\
CT \citep{melnychuk2022causal} & 0.3630 $\pm$ 0.0327 & 0.3966 $\pm$ 0.0269 & 0.4629 $\pm$ 0.0467 & 0.6433 $\pm$ 0.0502 & 0.8458 $\pm$ 0.2130 \\
Causal CPC \citep{bouchattaoui2024causal} & 0.3610 $\pm$ 0.0287 & 0.4055 $\pm$ 0.0513 & \underline{0.4418 $\pm$ 0.0222} & \underline{0.5710 $\pm$ 0.0431} & \underline{0.7627 $\pm$ 0.0573} \\
CTD-NKO \citep{wang2025enhancing} & \underline{0.3553 $\pm$ 0.0348} & \underline{0.3872 $\pm$ 0.0264} & 0.4547 $\pm$ 0.0322 & 0.5801 $\pm$ 0.0754 & 0.9987 $\pm$ 0.1647 \\
IGC-Net \citep{hess2026igc} & 0.4390 $\pm$ 0.0532 & 0.5397 $\pm$ 0.1107 & 0.5732 $\pm$ 0.0468 & 0.7030 $\pm$ 0.0830 & 0.8287 $\pm$ 0.1329 \\
\midrule
HiSTOR & \textbf{0.3307 $\pm$ 0.0165} & \textbf{0.3590 $\pm$ 0.0182} & \textbf{0.4088 $\pm$ 0.0249} & \textbf{0.4959 $\pm$ 0.0408} & \textbf{0.6780 $\pm$ 0.0294} \\
\bottomrule
\end{tabular}}
\end{table}

\begin{table}[H]
\caption{CancerSim: $\tau$=6. Values are NRMSE, mean $\pm$ sample SD across five randomly selected seeds $\{3,101,42,8,9\}$. Lower is better; the lowest mean is bold and the second-lowest underlined.}
\label{tab:cancersim_tau6_0}
\centering\footnotesize
\setlength{\tabcolsep}{3pt}
\resizebox{\linewidth}{!}{%
\begin{tabular}{@{}lccccc@{}}
\toprule
Method & $\gamma$=0 & $\gamma$=1 & $\gamma$=2 & $\gamma$=3 & $\gamma$=4 \\
\midrule
MSM \citep{robins2000marginal} & 0.6995 $\pm$ 0.0387 & 0.5805 $\pm$ 0.0272 & 0.5482 $\pm$ 0.0285 & 0.5792 $\pm$ 0.0194 & 0.6899 $\pm$ 0.0397 \\
RMSN \citep{lim2018forecasting} & 0.2648 $\pm$ 0.0380 & 0.2863 $\pm$ 0.0380 & \underline{0.3091 $\pm$ 0.0188} & 0.4479 $\pm$ 0.1021 & \underline{0.5498 $\pm$ 0.0400} \\
CRN \citep{bica2020estimating} & 0.2533 $\pm$ 0.0236 & 0.2874 $\pm$ 0.0165 & 0.3428 $\pm$ 0.0225 & 0.4616 $\pm$ 0.0873 & 0.7041 $\pm$ 0.0592 \\
CT \citep{melnychuk2022causal} & 0.2566 $\pm$ 0.0286 & 0.2806 $\pm$ 0.0225 & 0.3352 $\pm$ 0.0513 & 0.4578 $\pm$ 0.0380 & 0.6323 $\pm$ 0.2171 \\
Causal CPC \citep{bouchattaoui2024causal} & 0.2622 $\pm$ 0.0182 & 0.2925 $\pm$ 0.0479 & 0.3213 $\pm$ 0.0170 & 0.4434 $\pm$ 0.0416 & 0.5598 $\pm$ 0.0440 \\
CTD-NKO \citep{wang2025enhancing} & \underline{0.2466 $\pm$ 0.0237} & \underline{0.2723 $\pm$ 0.0190} & 0.3219 $\pm$ 0.0182 & \underline{0.4144 $\pm$ 0.0566} & 0.7617 $\pm$ 0.1655 \\
IGC-Net \citep{hess2026igc} & 0.3777 $\pm$ 0.0316 & 0.4813 $\pm$ 0.2088 & 0.4291 $\pm$ 0.0802 & 0.5134 $\pm$ 0.1244 & 0.6374 $\pm$ 0.0859 \\
\midrule
HiSTOR & \textbf{0.2359 $\pm$ 0.0171} & \textbf{0.2627 $\pm$ 0.0178} & \textbf{0.3004 $\pm$ 0.0181} & \textbf{0.3840 $\pm$ 0.0471} & \textbf{0.5485 $\pm$ 0.0536} \\
\bottomrule
\end{tabular}}
\end{table}

\subsection{Paired Comparisons}\label{app:cancer_paired}
We compare HiSTOR with the strongest external baseline identified by the cross-seed mean in each condition. HiSTOR achieves a lower mean within-seed NRMSE difference in every CancerSim condition. It wins in three or more of the five seeds for every independent one-step comparison and in all five seeds for every five-horizon-average comparison. The tables report the mean paired difference, defined as HiSTOR minus the comparator, together with approximate 95\% t intervals across the five seeds.

\begin{table}[H]
\caption{CancerSim: paired comparisons for independent one-step.}
\label{tab:paired_CancerSim_one}
\centering\footnotesize
\setlength{\tabcolsep}{3pt}
\resizebox{\linewidth}{!}{%
\begin{tabular}{@{}lcccc@{}}
\toprule
Condition & Comparator & Wins / 5 & Mean difference & Approx. 95\% interval \\
\midrule
0 & CT \citep{melnychuk2022causal} & 3/5 & -0.0248 & [-0.0693, 0.0198] \\
1 & CT \citep{melnychuk2022causal} & 4/5 & -0.0311 & [-0.0811, 0.0188] \\
2 & CT \citep{melnychuk2022causal} & 5/5 & -0.0597 & [-0.1243, 0.0049] \\
3 & CTD-NKO \citep{wang2025enhancing} & 5/5 & -0.0957 & [-0.1435, -0.0478] \\
4 & CTD-NKO \citep{wang2025enhancing} & 5/5 & -0.0917 & [-0.1622, -0.0212] \\
\bottomrule
\end{tabular}}
\end{table}

\begin{table}[H]
\caption{CancerSim: paired comparisons for five-horizon averages.}
\label{tab:paired_CancerSim_H5mean}
\centering\footnotesize
\setlength{\tabcolsep}{3pt}
\resizebox{\linewidth}{!}{%
\begin{tabular}{@{}lcccc@{}}
\toprule
Condition & Comparator & Wins / 5 & Mean difference & Approx. 95\% interval \\
\midrule
0 & Causal CPC \citep{bouchattaoui2024causal} & 5/5 & -0.0351 & [-0.0633, -0.0068] \\
1 & CTD-NKO \citep{wang2025enhancing} & 5/5 & -0.0373 & [-0.0650, -0.0096] \\
2 & CTD-NKO \citep{wang2025enhancing} & 5/5 & -0.0509 & [-0.1024, 0.0007] \\
3 & Causal CPC \citep{bouchattaoui2024causal} & 5/5 & -0.0873 & [-0.1167, -0.0579] \\
4 & Causal CPC \citep{bouchattaoui2024causal} & 5/5 & -0.1596 & [-0.2107, -0.1085] \\
\bottomrule
\end{tabular}}
\end{table}
\section{Detailed CVSim Results}\label{app:cvsim_detailed}

Values are NRMSE, mean $\pm$ sample SD across five randomly selected seeds $\{3,101,42,8,9\}$. Lower is better; the lowest mean is bold and the second-lowest underlined.

\begin{table}[H]
\caption{CVSim Static: independent one-step. Values are NRMSE, mean $\pm$ sample SD across five randomly selected seeds $\{3,101,42,8,9\}$. Lower is better; the lowest mean is bold and the second-lowest underlined.}
\label{tab:cvsimstatic_one_0}
\centering\footnotesize
\setlength{\tabcolsep}{3pt}
\resizebox{\linewidth}{!}{%
\begin{tabular}{@{}lccccc@{}}
\toprule
Method & d\_hd=0 & d\_hd=10 & d\_hd=20 & d\_hd=50 & d\_hd=100 \\
\midrule
MSM \citep{robins2000marginal} & 2.7450 $\pm$ 1.6322 & 2.7454 $\pm$ 1.6333 & 2.7450 $\pm$ 1.6337 & 2.7442 $\pm$ 1.6323 & 2.7449 $\pm$ 1.6336 \\
RMSN \citep{lim2018forecasting} & 2.6288 $\pm$ 1.4077 & 2.5587 $\pm$ 1.3994 & 2.5495 $\pm$ 1.3605 & 2.5482 $\pm$ 1.3358 & 2.5416 $\pm$ 1.3063 \\
CRN \citep{bica2020estimating} & 4.3197 $\pm$ 3.1278 & 3.4662 $\pm$ 1.5178 & 3.4285 $\pm$ 1.5090 & 3.0455 $\pm$ 1.3404 & 2.9070 $\pm$ 1.3528 \\
CT \citep{melnychuk2022causal} & 1.1909 $\pm$ 0.2306 & 1.1130 $\pm$ 0.2309 & 1.0561 $\pm$ 0.2623 & 1.0544 $\pm$ 0.2537 & 1.1011 $\pm$ 0.2754 \\
Causal CPC \citep{bouchattaoui2024causal} & 1.0229 $\pm$ 0.2806 & 0.9878 $\pm$ 0.1462 & 0.9647 $\pm$ 0.2157 & 1.0222 $\pm$ 0.1746 & 1.0315 $\pm$ 0.1334 \\
CTD-NKO \citep{wang2025enhancing} & 2.6155 $\pm$ 1.3320 & 2.6028 $\pm$ 1.2879 & 2.6830 $\pm$ 1.3248 & 2.7466 $\pm$ 1.3316 & 2.6789 $\pm$ 1.3221 \\
IGC-Net \citep{hess2026igc} & \underline{0.8185 $\pm$ 0.1649} & \textbf{0.7952 $\pm$ 0.1937} & \textbf{0.8182 $\pm$ 0.2002} & \textbf{0.7942 $\pm$ 0.1962} & \textbf{0.8289 $\pm$ 0.1985} \\
\midrule
HiSTOR & \textbf{0.7240 $\pm$ 0.2137} & \underline{0.8139 $\pm$ 0.2282} & \underline{0.8574 $\pm$ 0.2685} & \underline{0.8807 $\pm$ 0.2587} & \underline{0.8976 $\pm$ 0.2576} \\
\bottomrule
\end{tabular}}
\end{table}

\begin{table}[H]
\caption{CVSim Static: independent one-step. Values are NRMSE, mean $\pm$ sample SD across five randomly selected seeds $\{3,101,42,8,9\}$. Lower is better; the lowest mean is bold and the second-lowest underlined.}
\label{tab:cvsimstatic_one_1}
\centering\footnotesize
\setlength{\tabcolsep}{3pt}
\resizebox{\linewidth}{!}{%
\begin{tabular}{@{}lcccc@{}}
\toprule
Method & d\_hd=200 & d\_hd=500 & d\_hd=1000 & d\_hd=2000 \\
\midrule
MSM \citep{robins2000marginal} & 2.7480 $\pm$ 1.6364 & 2.7470 $\pm$ 1.6352 & 2.7533 $\pm$ 1.6372 & 2.7967 $\pm$ 1.6590 \\
RMSN \citep{lim2018forecasting} & 2.5427 $\pm$ 1.2593 & 2.4430 $\pm$ 1.1556 & 2.4987 $\pm$ 1.0909 & 2.8571 $\pm$ 0.9745 \\
CRN \citep{bica2020estimating} & 2.9344 $\pm$ 1.2817 & 3.4432 $\pm$ 1.2395 & 5.0170 $\pm$ 2.5278 & 4.3593 $\pm$ 0.9090 \\
CT \citep{melnychuk2022causal} & 1.1546 $\pm$ 0.2004 & 1.0759 $\pm$ 0.2817 & 1.1336 $\pm$ 0.3511 & 1.1746 $\pm$ 0.3178 \\
Causal CPC \citep{bouchattaoui2024causal} & 0.9441 $\pm$ 0.1871 & 1.0135 $\pm$ 0.1543 & 1.1242 $\pm$ 0.2058 & 1.2622 $\pm$ 0.3074 \\
CTD-NKO \citep{wang2025enhancing} & 2.8848 $\pm$ 1.5157 & 3.1853 $\pm$ 1.6830 & 3.1071 $\pm$ 1.5156 & 3.3701 $\pm$ 1.7194 \\
IGC-Net \citep{hess2026igc} & \underline{0.8390 $\pm$ 0.2010} & \underline{0.8750 $\pm$ 0.2725} & \underline{0.8733 $\pm$ 0.2340} & \underline{0.9535 $\pm$ 0.2743} \\
\midrule
HiSTOR & \textbf{0.8118 $\pm$ 0.2348} & \textbf{0.7936 $\pm$ 0.2279} & \textbf{0.8362 $\pm$ 0.2442} & \textbf{0.8819 $\pm$ 0.2571} \\
\bottomrule
\end{tabular}}
\end{table}

\begin{table}[H]
\caption{CVSim Static: H=5 mean. Values are NRMSE, mean $\pm$ sample SD across five randomly selected seeds $\{3,101,42,8,9\}$. Lower is better; the lowest mean is bold and the second-lowest underlined.}
\label{tab:cvsimstatic_H=5mean_0}
\centering\footnotesize
\setlength{\tabcolsep}{3pt}
\resizebox{\linewidth}{!}{%
\begin{tabular}{@{}lccccc@{}}
\toprule
Method & d\_hd=0 & d\_hd=10 & d\_hd=20 & d\_hd=50 & d\_hd=100 \\
\midrule
MSM \citep{robins2000marginal} & 3.8995 $\pm$ 2.3936 & 3.9023 $\pm$ 2.3966 & 3.9011 $\pm$ 2.4019 & 3.8970 $\pm$ 2.3968 & 3.9030 $\pm$ 2.4009 \\
RMSN \citep{lim2018forecasting} & 1.4049 $\pm$ 0.2490 & 1.3870 $\pm$ 0.2611 & 1.3362 $\pm$ 0.3567 & 1.5987 $\pm$ 0.4111 & 1.5739 $\pm$ 0.2969 \\
CRN \citep{bica2020estimating} & 9.0673 $\pm$ 7.7702 & 8.9171 $\pm$ 7.0460 & 8.8059 $\pm$ 6.8482 & 8.5169 $\pm$ 6.0701 & 6.9780 $\pm$ 4.4332 \\
CT \citep{melnychuk2022causal} & 2.0750 $\pm$ 0.4091 & 1.9703 $\pm$ 0.5882 & 1.8430 $\pm$ 0.4586 & 1.8120 $\pm$ 0.4352 & 2.1216 $\pm$ 0.6573 \\
Causal CPC \citep{bouchattaoui2024causal} & \underline{1.2583 $\pm$ 0.2243} & 1.3926 $\pm$ 0.2146 & 1.3327 $\pm$ 0.3708 & \underline{1.4799 $\pm$ 0.2834} & \underline{1.4441 $\pm$ 0.1880} \\
CTD-NKO \citep{wang2025enhancing} & 1.3025 $\pm$ 0.3854 & \underline{1.2620 $\pm$ 0.4424} & \underline{1.1858 $\pm$ 0.5011} & 1.5834 $\pm$ 0.6597 & 1.5202 $\pm$ 0.6123 \\
IGC-Net \citep{hess2026igc} & 2.4918 $\pm$ 1.0791 & 2.3128 $\pm$ 0.8874 & 2.3176 $\pm$ 0.8236 & 2.4622 $\pm$ 0.9580 & 2.4297 $\pm$ 1.2103 \\
\midrule
HiSTOR & \textbf{1.0005 $\pm$ 0.2796} & \textbf{1.1309 $\pm$ 0.2932} & \textbf{1.1389 $\pm$ 0.3556} & \textbf{1.1894 $\pm$ 0.2853} & \textbf{1.2350 $\pm$ 0.3525} \\
\bottomrule
\end{tabular}}
\end{table}

\begin{table}[H]
\caption{CVSim Static: H=5 mean. Values are NRMSE, mean $\pm$ sample SD across five randomly selected seeds $\{3,101,42,8,9\}$. Lower is better; the lowest mean is bold and the second-lowest underlined.}
\label{tab:cvsimstatic_H=5mean_1}
\centering\footnotesize
\setlength{\tabcolsep}{3pt}
\resizebox{\linewidth}{!}{%
\begin{tabular}{@{}lcccc@{}}
\toprule
Method & d\_hd=200 & d\_hd=500 & d\_hd=1000 & d\_hd=2000 \\
\midrule
MSM \citep{robins2000marginal} & 3.9029 $\pm$ 2.3989 & 3.8979 $\pm$ 2.4004 & 3.8804 $\pm$ 2.3867 & 3.9338 $\pm$ 2.4313 \\
RMSN \citep{lim2018forecasting} & 1.6446 $\pm$ 0.3517 & 1.8023 $\pm$ 0.4819 & 2.2660 $\pm$ 0.6327 & 3.3983 $\pm$ 0.6877 \\
CRN \citep{bica2020estimating} & 7.4061 $\pm$ 4.4954 & 6.6797 $\pm$ 3.4750 & 10.1151 $\pm$ 5.9721 & 7.6749 $\pm$ 1.7319 \\
CT \citep{melnychuk2022causal} & 2.2070 $\pm$ 0.5525 & 2.3559 $\pm$ 0.6702 & 2.0990 $\pm$ 0.4857 & 2.2217 $\pm$ 0.7037 \\
Causal CPC \citep{bouchattaoui2024causal} & \underline{1.3728 $\pm$ 0.3327} & \underline{1.3718 $\pm$ 0.3357} & \underline{1.5421 $\pm$ 0.6982} & \underline{1.6714 $\pm$ 0.3533} \\
CTD-NKO \citep{wang2025enhancing} & 2.2378 $\pm$ 1.1573 & 3.9033 $\pm$ 2.6208 & 3.0568 $\pm$ 1.5703 & 4.4738 $\pm$ 2.0124 \\
IGC-Net \citep{hess2026igc} & 2.2502 $\pm$ 0.7940 & 2.3289 $\pm$ 1.0383 & 2.3028 $\pm$ 1.1119 & 2.4854 $\pm$ 1.0472 \\
\midrule
HiSTOR & \textbf{1.0722 $\pm$ 0.3035} & \textbf{1.0740 $\pm$ 0.2879} & \textbf{1.1355 $\pm$ 0.3909} & \textbf{1.1973 $\pm$ 0.3572} \\
\bottomrule
\end{tabular}}
\end{table}

\begin{table}[H]
\caption{CVSim Static: $\tau$=1. Values are NRMSE, mean $\pm$ sample SD across five randomly selected seeds $\{3,101,42,8,9\}$. Lower is better; the lowest mean is bold and the second-lowest underlined.}
\label{tab:cvsimstatic_tau1_0}
\centering\footnotesize
\setlength{\tabcolsep}{3pt}
\resizebox{\linewidth}{!}{%
\begin{tabular}{@{}lccccc@{}}
\toprule
Method & d\_hd=0 & d\_hd=10 & d\_hd=20 & d\_hd=50 & d\_hd=100 \\
\midrule
MSM \citep{robins2000marginal} & 3.7967 $\pm$ 2.3417 & 3.7980 $\pm$ 2.3431 & 3.7982 $\pm$ 2.3480 & 3.7950 $\pm$ 2.3471 & 3.7989 $\pm$ 2.3487 \\
RMSN \citep{lim2018forecasting} & 1.1531 $\pm$ 0.2963 & 1.0590 $\pm$ 0.2679 & 1.0448 $\pm$ 0.1897 & 1.2750 $\pm$ 0.4344 & 1.3775 $\pm$ 0.4149 \\
CRN \citep{bica2020estimating} & 6.8521 $\pm$ 6.2222 & 4.8815 $\pm$ 3.3568 & 4.8326 $\pm$ 3.4881 & 4.0465 $\pm$ 3.2664 & 3.8699 $\pm$ 3.3554 \\
CT \citep{melnychuk2022causal} & 1.4014 $\pm$ 0.3895 & 1.3001 $\pm$ 0.4927 & 1.2009 $\pm$ 0.3307 & 1.1995 $\pm$ 0.3709 & 1.2679 $\pm$ 0.3902 \\
Causal CPC \citep{bouchattaoui2024causal} & 1.1443 $\pm$ 0.2723 & 1.1734 $\pm$ 0.3190 & 1.2380 $\pm$ 0.6390 & 1.2519 $\pm$ 0.3046 & 1.2238 $\pm$ 0.4068 \\
CTD-NKO \citep{wang2025enhancing} & 1.0182 $\pm$ 0.3781 & \textbf{0.9375 $\pm$ 0.1359} & \underline{0.8845 $\pm$ 0.2905} & 1.2035 $\pm$ 0.1932 & 1.3646 $\pm$ 0.2754 \\
IGC-Net \citep{hess2026igc} & \underline{0.9955 $\pm$ 0.1839} & \underline{0.9447 $\pm$ 0.1875} & \textbf{0.8769 $\pm$ 0.1344} & \textbf{0.9057 $\pm$ 0.1571} & \textbf{0.9766 $\pm$ 0.1804} \\
\midrule
HiSTOR & \textbf{0.8319 $\pm$ 0.1968} & 0.9592 $\pm$ 0.2238 & 1.0034 $\pm$ 0.2195 & \underline{1.0010 $\pm$ 0.2726} & \underline{1.0198 $\pm$ 0.2554} \\
\bottomrule
\end{tabular}}
\end{table}

\begin{table}[H]
\caption{CVSim Static: $\tau$=1. Values are NRMSE, mean $\pm$ sample SD across five randomly selected seeds $\{3,101,42,8,9\}$. Lower is better; the lowest mean is bold and the second-lowest underlined.}
\label{tab:cvsimstatic_tau1_1}
\centering\footnotesize
\setlength{\tabcolsep}{3pt}
\resizebox{\linewidth}{!}{%
\begin{tabular}{@{}lcccc@{}}
\toprule
Method & d\_hd=200 & d\_hd=500 & d\_hd=1000 & d\_hd=2000 \\
\midrule
MSM \citep{robins2000marginal} & 3.8005 $\pm$ 2.3501 & 3.7934 $\pm$ 2.3489 & 3.7636 $\pm$ 2.3257 & 3.7965 $\pm$ 2.3568 \\
RMSN \citep{lim2018forecasting} & 1.4634 $\pm$ 0.3948 & 1.6731 $\pm$ 0.2485 & 2.3968 $\pm$ 0.7055 & 3.7753 $\pm$ 1.1812 \\
CRN \citep{bica2020estimating} & 4.0863 $\pm$ 3.4004 & 4.5051 $\pm$ 3.0349 & 8.0181 $\pm$ 5.7618 & 6.0755 $\pm$ 1.6745 \\
CT \citep{melnychuk2022causal} & 1.4265 $\pm$ 0.3914 & 1.4002 $\pm$ 0.5101 & 1.3635 $\pm$ 0.4614 & 1.4122 $\pm$ 0.4542 \\
Causal CPC \citep{bouchattaoui2024causal} & 1.2106 $\pm$ 0.4059 & 1.2280 $\pm$ 0.4627 & 1.4765 $\pm$ 0.9703 & 1.5326 $\pm$ 0.6243 \\
CTD-NKO \citep{wang2025enhancing} & 1.8459 $\pm$ 1.1898 & 2.8919 $\pm$ 1.5128 & 2.6533 $\pm$ 1.1288 & 3.8557 $\pm$ 1.4492 \\
IGC-Net \citep{hess2026igc} & \underline{0.9469 $\pm$ 0.2745} & \underline{1.0119 $\pm$ 0.2902} & \textbf{0.9624 $\pm$ 0.2520} & \underline{1.0543 $\pm$ 0.2876} \\
\midrule
HiSTOR & \textbf{0.9313 $\pm$ 0.2844} & \textbf{0.9059 $\pm$ 0.2721} & \underline{0.9752 $\pm$ 0.3340} & \textbf{0.9928 $\pm$ 0.3224} \\
\bottomrule
\end{tabular}}
\end{table}

\begin{table}[H]
\caption{CVSim Static: $\tau$=2. Values are NRMSE, mean $\pm$ sample SD across five randomly selected seeds $\{3,101,42,8,9\}$. Lower is better; the lowest mean is bold and the second-lowest underlined.}
\label{tab:cvsimstatic_tau2_0}
\centering\footnotesize
\setlength{\tabcolsep}{3pt}
\resizebox{\linewidth}{!}{%
\begin{tabular}{@{}lccccc@{}}
\toprule
Method & d\_hd=0 & d\_hd=10 & d\_hd=20 & d\_hd=50 & d\_hd=100 \\
\midrule
MSM \citep{robins2000marginal} & 3.8421 $\pm$ 2.3669 & 3.8444 $\pm$ 2.3688 & 3.8435 $\pm$ 2.3744 & 3.8398 $\pm$ 2.3717 & 3.8436 $\pm$ 2.3733 \\
RMSN \citep{lim2018forecasting} & 1.2818 $\pm$ 0.3601 & 1.3209 $\pm$ 0.3758 & 1.2316 $\pm$ 0.4053 & 1.5392 $\pm$ 0.5251 & 1.4706 $\pm$ 0.3845 \\
CRN \citep{bica2020estimating} & 8.7891 $\pm$ 7.7591 & 9.0417 $\pm$ 8.4202 & 9.2318 $\pm$ 8.6043 & 8.8772 $\pm$ 7.9215 & 6.3405 $\pm$ 5.1047 \\
CT \citep{melnychuk2022causal} & 1.7815 $\pm$ 0.5207 & 1.6361 $\pm$ 0.5089 & 1.5811 $\pm$ 0.2875 & 1.5177 $\pm$ 0.3577 & 1.7857 $\pm$ 0.4054 \\
Causal CPC \citep{bouchattaoui2024causal} & 1.2125 $\pm$ 0.1918 & 1.3220 $\pm$ 0.2847 & 1.2781 $\pm$ 0.5236 & 1.4102 $\pm$ 0.3295 & 1.3547 $\pm$ 0.2720 \\
CTD-NKO \citep{wang2025enhancing} & \underline{1.1625 $\pm$ 0.3745} & \underline{1.1364 $\pm$ 0.2813} & \textbf{1.0632 $\pm$ 0.3760} & \underline{1.2666 $\pm$ 0.4762} & \underline{1.2273 $\pm$ 0.5151} \\
IGC-Net \citep{hess2026igc} & 1.4297 $\pm$ 0.6540 & 1.5615 $\pm$ 0.5378 & 1.4286 $\pm$ 0.6138 & 1.5837 $\pm$ 0.6295 & 1.3760 $\pm$ 0.6378 \\
\midrule
HiSTOR & \textbf{0.9089 $\pm$ 0.2846} & \textbf{1.0375 $\pm$ 0.3101} & \underline{1.0669 $\pm$ 0.2943} & \textbf{1.0878 $\pm$ 0.3019} & \textbf{1.1359 $\pm$ 0.3119} \\
\bottomrule
\end{tabular}}
\end{table}

\begin{table}[H]
\caption{CVSim Static: $\tau$=2. Values are NRMSE, mean $\pm$ sample SD across five randomly selected seeds $\{3,101,42,8,9\}$. Lower is better; the lowest mean is bold and the second-lowest underlined.}
\label{tab:cvsimstatic_tau2_1}
\centering\footnotesize
\setlength{\tabcolsep}{3pt}
\resizebox{\linewidth}{!}{%
\begin{tabular}{@{}lcccc@{}}
\toprule
Method & d\_hd=200 & d\_hd=500 & d\_hd=1000 & d\_hd=2000 \\
\midrule
MSM \citep{robins2000marginal} & 3.8473 $\pm$ 2.3735 & 3.8390 $\pm$ 2.3698 & 3.8171 $\pm$ 2.3576 & 3.8602 $\pm$ 2.3920 \\
RMSN \citep{lim2018forecasting} & 1.5285 $\pm$ 0.4599 & 1.6396 $\pm$ 0.6311 & 2.0595 $\pm$ 0.7100 & 3.2020 $\pm$ 0.7347 \\
CRN \citep{bica2020estimating} & 7.0946 $\pm$ 5.1845 & 5.9010 $\pm$ 3.9557 & 9.5498 $\pm$ 6.1773 & 7.0041 $\pm$ 1.8783 \\
CT \citep{melnychuk2022causal} & 1.8717 $\pm$ 0.3589 & 1.9612 $\pm$ 0.6664 & 1.7431 $\pm$ 0.5346 & 1.8464 $\pm$ 0.4721 \\
Causal CPC \citep{bouchattaoui2024causal} & \underline{1.2941 $\pm$ 0.3640} & \underline{1.3024 $\pm$ 0.4044} & 1.5344 $\pm$ 0.8348 & \underline{1.5859 $\pm$ 0.5107} \\
CTD-NKO \citep{wang2025enhancing} & 1.7462 $\pm$ 1.0320 & 3.0734 $\pm$ 2.2712 & 2.5635 $\pm$ 1.4733 & 3.8654 $\pm$ 1.9996 \\
IGC-Net \citep{hess2026igc} & 1.3990 $\pm$ 0.5394 & 1.5334 $\pm$ 0.7683 & \underline{1.4196 $\pm$ 0.7057} & 1.6986 $\pm$ 0.6330 \\
\midrule
HiSTOR & \textbf{1.0511 $\pm$ 0.2511} & \textbf{0.9992 $\pm$ 0.2571} & \textbf{1.0669 $\pm$ 0.3857} & \textbf{1.0699 $\pm$ 0.3791} \\
\bottomrule
\end{tabular}}
\end{table}

\begin{table}[H]
\caption{CVSim Static: $\tau$=3. Values are NRMSE, mean $\pm$ sample SD across five randomly selected seeds $\{3,101,42,8,9\}$. Lower is better; the lowest mean is bold and the second-lowest underlined.}
\label{tab:cvsimstatic_tau3_0}
\centering\footnotesize
\setlength{\tabcolsep}{3pt}
\resizebox{\linewidth}{!}{%
\begin{tabular}{@{}lccccc@{}}
\toprule
Method & d\_hd=0 & d\_hd=10 & d\_hd=20 & d\_hd=50 & d\_hd=100 \\
\midrule
MSM \citep{robins2000marginal} & 3.8858 $\pm$ 2.3833 & 3.8889 $\pm$ 2.3869 & 3.8878 $\pm$ 2.3928 & 3.8827 $\pm$ 2.3867 & 3.8893 $\pm$ 2.3900 \\
RMSN \citep{lim2018forecasting} & 1.3476 $\pm$ 0.3118 & 1.3437 $\pm$ 0.2847 & 1.2530 $\pm$ 0.4230 & 1.5526 $\pm$ 0.4808 & 1.5011 $\pm$ 0.3281 \\
CRN \citep{bica2020estimating} & 9.1894 $\pm$ 8.0217 & 9.4800 $\pm$ 8.2640 & 9.4051 $\pm$ 8.1390 & 9.3331 $\pm$ 7.6721 & 7.1943 $\pm$ 4.7858 \\
CT \citep{melnychuk2022causal} & 2.0565 $\pm$ 0.4461 & 1.9468 $\pm$ 0.6155 & 1.8455 $\pm$ 0.4525 & 1.7779 $\pm$ 0.4644 & 2.1193 $\pm$ 0.6504 \\
Causal CPC \citep{bouchattaoui2024causal} & \underline{1.2392 $\pm$ 0.1970} & 1.3901 $\pm$ 0.2311 & 1.3172 $\pm$ 0.3776 & 1.4882 $\pm$ 0.3049 & 1.4384 $\pm$ 0.1717 \\
CTD-NKO \citep{wang2025enhancing} & 1.2602 $\pm$ 0.3715 & \underline{1.2056 $\pm$ 0.4227} & \underline{1.1264 $\pm$ 0.4706} & \underline{1.4315 $\pm$ 0.6483} & \underline{1.3177 $\pm$ 0.6717} \\
IGC-Net \citep{hess2026igc} & 2.2344 $\pm$ 1.2891 & 2.2210 $\pm$ 0.9903 & 2.4918 $\pm$ 1.2630 & 2.7502 $\pm$ 1.6769 & 2.2083 $\pm$ 1.2575 \\
\midrule
HiSTOR & \textbf{0.9900 $\pm$ 0.3023} & \textbf{1.1090 $\pm$ 0.3407} & \textbf{1.1037 $\pm$ 0.3987} & \textbf{1.2015 $\pm$ 0.2967} & \textbf{1.2166 $\pm$ 0.3975} \\
\bottomrule
\end{tabular}}
\end{table}

\begin{table}[H]
\caption{CVSim Static: $\tau$=3. Values are NRMSE, mean $\pm$ sample SD across five randomly selected seeds $\{3,101,42,8,9\}$. Lower is better; the lowest mean is bold and the second-lowest underlined.}
\label{tab:cvsimstatic_tau3_1}
\centering\footnotesize
\setlength{\tabcolsep}{3pt}
\resizebox{\linewidth}{!}{%
\begin{tabular}{@{}lcccc@{}}
\toprule
Method & d\_hd=200 & d\_hd=500 & d\_hd=1000 & d\_hd=2000 \\
\midrule
MSM \citep{robins2000marginal} & 3.8885 $\pm$ 2.3890 & 3.8831 $\pm$ 2.3888 & 3.8620 $\pm$ 2.3730 & 3.9135 $\pm$ 2.4187 \\
RMSN \citep{lim2018forecasting} & 1.5693 $\pm$ 0.4167 & 1.6966 $\pm$ 0.5911 & 2.0625 $\pm$ 0.6452 & 3.1471 $\pm$ 0.6391 \\
CRN \citep{bica2020estimating} & 7.7512 $\pm$ 5.0555 & 6.7121 $\pm$ 3.7592 & 10.1678 $\pm$ 6.1497 & 7.6694 $\pm$ 1.9110 \\
CT \citep{melnychuk2022causal} & 2.2215 $\pm$ 0.5228 & 2.4012 $\pm$ 0.7244 & 2.0839 $\pm$ 0.4917 & 2.2336 $\pm$ 0.8017 \\
Causal CPC \citep{bouchattaoui2024causal} & \underline{1.3440 $\pm$ 0.3049} & \underline{1.3528 $\pm$ 0.3272} & \underline{1.5384 $\pm$ 0.6873} & \underline{1.6396 $\pm$ 0.3721} \\
CTD-NKO \citep{wang2025enhancing} & 2.0075 $\pm$ 1.1139 & 3.6483 $\pm$ 2.8250 & 2.8035 $\pm$ 1.7030 & 4.2220 $\pm$ 2.2681 \\
IGC-Net \citep{hess2026igc} & 1.9483 $\pm$ 0.8715 & 2.1269 $\pm$ 1.0623 & 2.2383 $\pm$ 1.2719 & 2.4755 $\pm$ 1.3731 \\
\midrule
HiSTOR & \textbf{1.0594 $\pm$ 0.3136} & \textbf{1.0587 $\pm$ 0.3167} & \textbf{1.1272 $\pm$ 0.4279} & \textbf{1.1799 $\pm$ 0.4036} \\
\bottomrule
\end{tabular}}
\end{table}

\begin{table}[H]
\caption{CVSim Static: $\tau$=4. Values are NRMSE, mean $\pm$ sample SD across five randomly selected seeds $\{3,101,42,8,9\}$. Lower is better; the lowest mean is bold and the second-lowest underlined.}
\label{tab:cvsimstatic_tau4_0}
\centering\footnotesize
\setlength{\tabcolsep}{3pt}
\resizebox{\linewidth}{!}{%
\begin{tabular}{@{}lccccc@{}}
\toprule
Method & d\_hd=0 & d\_hd=10 & d\_hd=20 & d\_hd=50 & d\_hd=100 \\
\midrule
MSM \citep{robins2000marginal} & 3.9651 $\pm$ 2.4204 & 3.9685 $\pm$ 2.4244 & 3.9672 $\pm$ 2.4295 & 3.9625 $\pm$ 2.4224 & 3.9699 $\pm$ 2.4303 \\
RMSN \citep{lim2018forecasting} & 1.5065 $\pm$ 0.3061 & 1.4922 $\pm$ 0.2924 & 1.4398 $\pm$ 0.4848 & 1.7078 $\pm$ 0.5304 & 1.6583 $\pm$ 0.3616 \\
CRN \citep{bica2020estimating} & 9.8884 $\pm$ 8.3587 & 10.2238 $\pm$ 8.2640 & 9.9621 $\pm$ 7.8215 & 9.8489 $\pm$ 7.2351 & 8.2260 $\pm$ 4.7884 \\
CT \citep{melnychuk2022causal} & 2.4058 $\pm$ 0.4334 & 2.3078 $\pm$ 0.8270 & 2.1282 $\pm$ 0.7222 & 2.0975 $\pm$ 0.6891 & 2.5231 $\pm$ 1.0198 \\
Causal CPC \citep{bouchattaoui2024causal} & \underline{1.2969 $\pm$ 0.3309} & 1.4873 $\pm$ 0.2119 & 1.3800 $\pm$ 0.2703 & \underline{1.5829 $\pm$ 0.3423} & \underline{1.5449 $\pm$ 0.1957} \\
CTD-NKO \citep{wang2025enhancing} & 1.4428 $\pm$ 0.4619 & \underline{1.4238 $\pm$ 0.6436} & \underline{1.3305 $\pm$ 0.6556} & 1.8057 $\pm$ 0.9354 & 1.6440 $\pm$ 0.7936 \\
IGC-Net \citep{hess2026igc} & 3.6215 $\pm$ 1.6127 & 2.9644 $\pm$ 1.3032 & 3.2620 $\pm$ 1.2855 & 2.5970 $\pm$ 1.5192 & 3.1975 $\pm$ 1.5491 \\
\midrule
HiSTOR & \textbf{1.0798 $\pm$ 0.3425} & \textbf{1.2175 $\pm$ 0.3529} & \textbf{1.2034 $\pm$ 0.4397} & \textbf{1.2776 $\pm$ 0.3202} & \textbf{1.3276 $\pm$ 0.4217} \\
\bottomrule
\end{tabular}}
\end{table}

\begin{table}[H]
\caption{CVSim Static: $\tau$=4. Values are NRMSE, mean $\pm$ sample SD across five randomly selected seeds $\{3,101,42,8,9\}$. Lower is better; the lowest mean is bold and the second-lowest underlined.}
\label{tab:cvsimstatic_tau4_1}
\centering\footnotesize
\setlength{\tabcolsep}{3pt}
\resizebox{\linewidth}{!}{%
\begin{tabular}{@{}lcccc@{}}
\toprule
Method & d\_hd=200 & d\_hd=500 & d\_hd=1000 & d\_hd=2000 \\
\midrule
MSM \citep{robins2000marginal} & 3.9677 $\pm$ 2.4229 & 3.9630 $\pm$ 2.4252 & 3.9557 $\pm$ 2.4191 & 4.0165 $\pm$ 2.4791 \\
RMSN \citep{lim2018forecasting} & 1.7159 $\pm$ 0.4682 & 1.8696 $\pm$ 0.5355 & 2.2626 $\pm$ 0.6166 & 3.3231 $\pm$ 0.5906 \\
CRN \citep{bica2020estimating} & 8.6021 $\pm$ 5.1523 & 7.6696 $\pm$ 3.5421 & 11.0051 $\pm$ 6.0536 & 8.4334 $\pm$ 1.9068 \\
CT \citep{melnychuk2022causal} & 2.5971 $\pm$ 0.8808 & 2.8316 $\pm$ 0.8829 & 2.4722 $\pm$ 0.6265 & 2.6322 $\pm$ 1.1500 \\
Causal CPC \citep{bouchattaoui2024causal} & \underline{1.4506 $\pm$ 0.3698} & \underline{1.4478 $\pm$ 0.3257} & \underline{1.5702 $\pm$ 0.5720} & \underline{1.7409 $\pm$ 0.2619} \\
CTD-NKO \citep{wang2025enhancing} & 2.5020 $\pm$ 1.2278 & 4.4983 $\pm$ 3.1942 & 3.3130 $\pm$ 1.8087 & 4.8590 $\pm$ 2.2819 \\
IGC-Net \citep{hess2026igc} & 2.6430 $\pm$ 1.1436 & 3.1506 $\pm$ 1.8658 & 2.8266 $\pm$ 1.5578 & 2.9721 $\pm$ 1.4467 \\
\midrule
HiSTOR & \textbf{1.1243 $\pm$ 0.3496} & \textbf{1.1428 $\pm$ 0.3719} & \textbf{1.2018 $\pm$ 0.4650} & \textbf{1.2942 $\pm$ 0.4064} \\
\bottomrule
\end{tabular}}
\end{table}

\begin{table}[H]
\caption{CVSim Static: $\tau$=5. Values are NRMSE, mean $\pm$ sample SD across five randomly selected seeds $\{3,101,42,8,9\}$. Lower is better; the lowest mean is bold and the second-lowest underlined.}
\label{tab:cvsimstatic_tau5_0}
\centering\footnotesize
\setlength{\tabcolsep}{3pt}
\resizebox{\linewidth}{!}{%
\begin{tabular}{@{}lccccc@{}}
\toprule
Method & d\_hd=0 & d\_hd=10 & d\_hd=20 & d\_hd=50 & d\_hd=100 \\
\midrule
MSM \citep{robins2000marginal} & 4.0077 $\pm$ 2.4594 & 4.0114 $\pm$ 2.4640 & 4.0090 $\pm$ 2.4684 & 4.0053 $\pm$ 2.4597 & 4.0131 $\pm$ 2.4662 \\
RMSN \citep{lim2018forecasting} & 1.7355 $\pm$ 0.2844 & 1.7191 $\pm$ 0.3702 & 1.7118 $\pm$ 0.5419 & 1.9189 $\pm$ 0.6002 & 1.8623 $\pm$ 0.4085 \\
CRN \citep{bica2020estimating} & 10.6173 $\pm$ 8.6792 & 10.9587 $\pm$ 8.2702 & 10.5977 $\pm$ 7.6693 & 10.4788 $\pm$ 7.0183 & 9.2596 $\pm$ 4.9251 \\
CT \citep{melnychuk2022causal} & 2.7297 $\pm$ 0.5142 & 2.6609 $\pm$ 1.0704 & 2.4593 $\pm$ 1.0107 & 2.4672 $\pm$ 0.9946 & 2.9122 $\pm$ 1.3927 \\
Causal CPC \citep{bouchattaoui2024causal} & \underline{1.3985 $\pm$ 0.4466} & \underline{1.5903 $\pm$ 0.3243} & \underline{1.4500 $\pm$ 0.2503} & \underline{1.6662 $\pm$ 0.4207} & \underline{1.6587 $\pm$ 0.3049} \\
CTD-NKO \citep{wang2025enhancing} & 1.6286 $\pm$ 0.5091 & 1.6069 $\pm$ 0.8057 & 1.5242 $\pm$ 0.7642 & 2.2097 $\pm$ 1.1367 & 2.0473 $\pm$ 0.8866 \\
IGC-Net \citep{hess2026igc} & 4.1778 $\pm$ 2.0647 & 3.8722 $\pm$ 1.7267 & 3.5286 $\pm$ 1.3180 & 4.4741 $\pm$ 2.1373 & 4.3902 $\pm$ 2.6522 \\
\midrule
HiSTOR & \textbf{1.1918 $\pm$ 0.3160} & \textbf{1.3311 $\pm$ 0.3199} & \textbf{1.3172 $\pm$ 0.4643} & \textbf{1.3794 $\pm$ 0.2877} & \textbf{1.4751 $\pm$ 0.4031} \\
\bottomrule
\end{tabular}}
\end{table}

\begin{table}[H]
\caption{CVSim Static: $\tau$=5. Values are NRMSE, mean $\pm$ sample SD across five randomly selected seeds $\{3,101,42,8,9\}$. Lower is better; the lowest mean is bold and the second-lowest underlined.}
\label{tab:cvsimstatic_tau5_1}
\centering\footnotesize
\setlength{\tabcolsep}{3pt}
\resizebox{\linewidth}{!}{%
\begin{tabular}{@{}lcccc@{}}
\toprule
Method & d\_hd=200 & d\_hd=500 & d\_hd=1000 & d\_hd=2000 \\
\midrule
MSM \citep{robins2000marginal} & 4.0103 $\pm$ 2.4627 & 4.0108 $\pm$ 2.4733 & 4.0036 $\pm$ 2.4620 & 4.0823 $\pm$ 2.5153 \\
RMSN \citep{lim2018forecasting} & 1.9458 $\pm$ 0.5239 & 2.1325 $\pm$ 0.5349 & 2.5487 $\pm$ 0.5972 & 3.5440 $\pm$ 0.5534 \\
CRN \citep{bica2020estimating} & 9.4963 $\pm$ 5.3358 & 8.6107 $\pm$ 3.3439 & 11.8347 $\pm$ 5.9853 & 9.1921 $\pm$ 1.9449 \\
CT \citep{melnychuk2022causal} & 2.9184 $\pm$ 1.2140 & 3.1855 $\pm$ 1.1446 & 2.8321 $\pm$ 0.8347 & 2.9842 $\pm$ 1.3423 \\
Causal CPC \citep{bouchattaoui2024causal} & \underline{1.5645 $\pm$ 0.4738} & \underline{1.5278 $\pm$ 0.3469} & \underline{1.5910 $\pm$ 0.4851} & \underline{1.8580 $\pm$ 0.3081} \\
CTD-NKO \citep{wang2025enhancing} & 3.0872 $\pm$ 1.3923 & 5.4045 $\pm$ 3.4026 & 3.9509 $\pm$ 1.8503 & 5.5671 $\pm$ 2.1347 \\
IGC-Net \citep{hess2026igc} & 4.3137 $\pm$ 1.9070 & 3.8217 $\pm$ 1.3745 & 4.0671 $\pm$ 1.9654 & 4.2264 $\pm$ 1.8394 \\
\midrule
HiSTOR & \textbf{1.1950 $\pm$ 0.3910} & \textbf{1.2633 $\pm$ 0.3451} & \textbf{1.3065 $\pm$ 0.4236} & \textbf{1.4497 $\pm$ 0.4051} \\
\bottomrule
\end{tabular}}
\end{table}

\begin{table}[H]
\caption{CVSim Temporal: independent one-step. Values are NRMSE, mean $\pm$ sample SD across five randomly selected seeds $\{3,101,42,8,9\}$. Lower is better; the lowest mean is bold and the second-lowest underlined.}
\label{tab:cvsimtemporal_one_0}
\centering\footnotesize
\setlength{\tabcolsep}{3pt}
\resizebox{\linewidth}{!}{%
\begin{tabular}{@{}lccccc@{}}
\toprule
Method & d\_hd=0 & d\_hd=10 & d\_hd=20 & d\_hd=50 & d\_hd=100 \\
\midrule
MSM \citep{robins2000marginal} & 2.7450 $\pm$ 1.6322 & 2.7454 $\pm$ 1.6333 & 2.7452 $\pm$ 1.6325 & 2.7459 $\pm$ 1.6326 & 2.7468 $\pm$ 1.6339 \\
RMSN \citep{lim2018forecasting} & 2.6294 $\pm$ 1.4075 & 2.5317 $\pm$ 1.3379 & 2.4788 $\pm$ 1.3331 & 2.4691 $\pm$ 1.3105 & 2.4451 $\pm$ 1.1880 \\
CRN \citep{bica2020estimating} & 4.4422 $\pm$ 3.0405 & 3.0922 $\pm$ 0.9791 & 2.8961 $\pm$ 1.2966 & 2.8790 $\pm$ 1.6229 & 2.9901 $\pm$ 1.1956 \\
CT \citep{melnychuk2022causal} & 1.2015 $\pm$ 0.2318 & 1.1074 $\pm$ 0.2483 & 1.0451 $\pm$ 0.2390 & 1.0620 $\pm$ 0.2633 & 1.0627 $\pm$ 0.2857 \\
Causal CPC \citep{bouchattaoui2024causal} & 1.0343 $\pm$ 0.2127 & 1.0016 $\pm$ 0.1462 & 0.9737 $\pm$ 0.2364 & 0.9262 $\pm$ 0.1202 & 0.9020 $\pm$ 0.2280 \\
CTD-NKO \citep{wang2025enhancing} & 2.6337 $\pm$ 1.3748 & 2.6658 $\pm$ 1.4017 & 2.6573 $\pm$ 1.3121 & 2.7368 $\pm$ 1.3708 & 2.7254 $\pm$ 1.4472 \\
IGC-Net \citep{hess2026igc} & \underline{0.8135 $\pm$ 0.1684} & \textbf{0.7936 $\pm$ 0.1816} & \textbf{0.8030 $\pm$ 0.1764} & \textbf{0.8100 $\pm$ 0.1962} & \textbf{0.8206 $\pm$ 0.1942} \\
\midrule
HiSTOR & \textbf{0.7309 $\pm$ 0.2110} & \underline{0.8119 $\pm$ 0.2581} & \underline{0.8436 $\pm$ 0.2756} & \underline{0.8832 $\pm$ 0.2926} & \underline{0.8851 $\pm$ 0.2763} \\
\bottomrule
\end{tabular}}
\end{table}

\begin{table}[H]
\caption{CVSim Temporal: independent one-step. Values are NRMSE, mean $\pm$ sample SD across five randomly selected seeds $\{3,101,42,8,9\}$. Lower is better; the lowest mean is bold and the second-lowest underlined.}
\label{tab:cvsimtemporal_one_1}
\centering\footnotesize
\setlength{\tabcolsep}{3pt}
\resizebox{\linewidth}{!}{%
\begin{tabular}{@{}lcccc@{}}
\toprule
Method & d\_hd=200 & d\_hd=500 & d\_hd=1000 & d\_hd=2000 \\
\midrule
MSM \citep{robins2000marginal} & 2.7473 $\pm$ 1.6328 & 2.7503 $\pm$ 1.6370 & 2.7513 $\pm$ 1.6350 & 2.7764 $\pm$ 1.6611 \\
RMSN \citep{lim2018forecasting} & 2.4139 $\pm$ 1.1749 & 2.4714 $\pm$ 1.1313 & 2.4826 $\pm$ 1.0474 & 2.6053 $\pm$ 0.9674 \\
CRN \citep{bica2020estimating} & 2.7721 $\pm$ 1.1417 & 4.2994 $\pm$ 3.0813 & 3.8956 $\pm$ 2.0132 & 4.6170 $\pm$ 2.4998 \\
CT \citep{melnychuk2022causal} & 1.2468 $\pm$ 0.2041 & 1.0792 $\pm$ 0.3036 & 1.1018 $\pm$ 0.3079 & 1.1316 $\pm$ 0.2828 \\
Causal CPC \citep{bouchattaoui2024causal} & 0.8554 $\pm$ 0.2100 & 1.0186 $\pm$ 0.1984 & 1.0259 $\pm$ 0.2398 & 1.0417 $\pm$ 0.1852 \\
CTD-NKO \citep{wang2025enhancing} & 2.6362 $\pm$ 1.3536 & 3.3635 $\pm$ 1.7934 & 3.3747 $\pm$ 1.7954 & 3.3819 $\pm$ 1.7992 \\
IGC-Net \citep{hess2026igc} & \underline{0.8125 $\pm$ 0.1901} & \underline{0.8477 $\pm$ 0.2057} & \underline{0.8826 $\pm$ 0.2565} & \underline{0.9409 $\pm$ 0.2475} \\
\midrule
HiSTOR & \textbf{0.7768 $\pm$ 0.2086} & \textbf{0.7941 $\pm$ 0.2179} & \textbf{0.8261 $\pm$ 0.2387} & \textbf{0.8451 $\pm$ 0.2535} \\
\bottomrule
\end{tabular}}
\end{table}

\begin{table}[H]
\caption{CVSim Temporal: H=5 mean. Values are NRMSE, mean $\pm$ sample SD across five randomly selected seeds $\{3,101,42,8,9\}$. Lower is better; the lowest mean is bold and the second-lowest underlined.}
\label{tab:cvsimtemporal_H=5mean_0}
\centering\footnotesize
\setlength{\tabcolsep}{3pt}
\resizebox{\linewidth}{!}{%
\begin{tabular}{@{}lccccc@{}}
\toprule
Method & d\_hd=0 & d\_hd=10 & d\_hd=20 & d\_hd=50 & d\_hd=100 \\
\midrule
MSM \citep{robins2000marginal} & 3.8995 $\pm$ 2.3936 & 3.8986 $\pm$ 2.3975 & 3.8989 $\pm$ 2.3969 & 3.9016 $\pm$ 2.3974 & 3.9049 $\pm$ 2.3950 \\
RMSN \citep{lim2018forecasting} & 1.4081 $\pm$ 0.2477 & 1.4633 $\pm$ 0.3766 & 1.3780 $\pm$ 0.3220 & 1.5407 $\pm$ 0.3899 & 1.6851 $\pm$ 0.4357 \\
CRN \citep{bica2020estimating} & 9.1398 $\pm$ 7.6974 & 8.5451 $\pm$ 6.8844 & 8.6751 $\pm$ 6.9349 & 8.8647 $\pm$ 7.0063 & 7.2686 $\pm$ 4.7127 \\
CT \citep{melnychuk2022causal} & 2.1296 $\pm$ 0.5086 & 2.0102 $\pm$ 0.6012 & 1.7952 $\pm$ 0.3714 & 1.8452 $\pm$ 0.4301 & 1.9914 $\pm$ 0.4479 \\
Causal CPC \citep{bouchattaoui2024causal} & 1.3727 $\pm$ 0.3482 & 1.8688 $\pm$ 0.9381 & 1.6531 $\pm$ 0.5848 & \underline{1.3039 $\pm$ 0.3076} & \underline{1.2967 $\pm$ 0.2788} \\
CTD-NKO \citep{wang2025enhancing} & \underline{1.3442 $\pm$ 0.2660} & \underline{1.2410 $\pm$ 0.4004} & \underline{1.2593 $\pm$ 0.5008} & 1.3771 $\pm$ 0.4523 & 1.4741 $\pm$ 0.5179 \\
IGC-Net \citep{hess2026igc} & 2.3652 $\pm$ 1.1921 & 2.4896 $\pm$ 0.8284 & 2.3876 $\pm$ 0.9372 & 2.2500 $\pm$ 0.9989 & 2.3353 $\pm$ 1.1627 \\
\midrule
HiSTOR & \textbf{0.9831 $\pm$ 0.2840} & \textbf{1.0802 $\pm$ 0.3154} & \textbf{1.0901 $\pm$ 0.3135} & \textbf{1.1589 $\pm$ 0.4043} & \textbf{1.2333 $\pm$ 0.3158} \\
\bottomrule
\end{tabular}}
\end{table}

\begin{table}[H]
\caption{CVSim Temporal: H=5 mean. Values are NRMSE, mean $\pm$ sample SD across five randomly selected seeds $\{3,101,42,8,9\}$. Lower is better; the lowest mean is bold and the second-lowest underlined.}
\label{tab:cvsimtemporal_H=5mean_1}
\centering\footnotesize
\setlength{\tabcolsep}{3pt}
\resizebox{\linewidth}{!}{%
\begin{tabular}{@{}lcccc@{}}
\toprule
Method & d\_hd=200 & d\_hd=500 & d\_hd=1000 & d\_hd=2000 \\
\midrule
MSM \citep{robins2000marginal} & 3.9050 $\pm$ 2.4022 & 3.9088 $\pm$ 2.4049 & 3.9184 $\pm$ 2.3946 & 3.9638 $\pm$ 2.4453 \\
RMSN \citep{lim2018forecasting} & 1.6362 $\pm$ 0.4110 & 1.7675 $\pm$ 0.4534 & 2.0422 $\pm$ 0.5464 & 2.6230 $\pm$ 0.8039 \\
CRN \citep{bica2020estimating} & 8.5728 $\pm$ 7.5055 & 9.2303 $\pm$ 8.2751 & 8.4620 $\pm$ 6.3207 & 9.1284 $\pm$ 6.2666 \\
CT \citep{melnychuk2022causal} & 2.1523 $\pm$ 0.7620 & 2.0136 $\pm$ 0.5755 & 2.0252 $\pm$ 0.4911 & 1.9134 $\pm$ 0.3355 \\
Causal CPC \citep{bouchattaoui2024causal} & \underline{1.1721 $\pm$ 0.2252} & \underline{1.7428 $\pm$ 0.4955} & \underline{1.4770 $\pm$ 0.8099} & \underline{1.3624 $\pm$ 0.4074} \\
CTD-NKO \citep{wang2025enhancing} & 1.6892 $\pm$ 0.6190 & 4.7669 $\pm$ 2.4956 & 4.8129 $\pm$ 2.5124 & 4.8762 $\pm$ 2.5523 \\
IGC-Net \citep{hess2026igc} & 2.2353 $\pm$ 0.9896 & 2.3707 $\pm$ 1.0259 & 2.3084 $\pm$ 1.0909 & 2.4119 $\pm$ 1.0914 \\
\midrule
HiSTOR & \textbf{1.0420 $\pm$ 0.2763} & \textbf{1.0786 $\pm$ 0.3004} & \textbf{1.0883 $\pm$ 0.3161} & \textbf{1.1674 $\pm$ 0.3325} \\
\bottomrule
\end{tabular}}
\end{table}

\begin{table}[H]
\caption{CVSim Temporal: $\tau$=1. Values are NRMSE, mean $\pm$ sample SD across five randomly selected seeds $\{3,101,42,8,9\}$. Lower is better; the lowest mean is bold and the second-lowest underlined.}
\label{tab:cvsimtemporal_tau1_0}
\centering\footnotesize
\setlength{\tabcolsep}{3pt}
\resizebox{\linewidth}{!}{%
\begin{tabular}{@{}lccccc@{}}
\toprule
Method & d\_hd=0 & d\_hd=10 & d\_hd=20 & d\_hd=50 & d\_hd=100 \\
\midrule
MSM \citep{robins2000marginal} & 3.7967 $\pm$ 2.3417 & 3.7962 $\pm$ 2.3460 & 3.7948 $\pm$ 2.3411 & 3.7981 $\pm$ 2.3436 & 3.8022 $\pm$ 2.3439 \\
RMSN \citep{lim2018forecasting} & 1.1525 $\pm$ 0.2959 & 1.0561 $\pm$ 0.2378 & 1.0494 $\pm$ 0.1727 & 1.0915 $\pm$ 0.1531 & 1.4266 $\pm$ 0.3521 \\
CRN \citep{bica2020estimating} & 7.1539 $\pm$ 6.0200 & 4.2965 $\pm$ 2.6300 & 4.0356 $\pm$ 3.2368 & 4.0400 $\pm$ 3.4593 & 3.9995 $\pm$ 3.0871 \\
CT \citep{melnychuk2022causal} & 1.4235 $\pm$ 0.4274 & 1.3013 $\pm$ 0.4951 & 1.2217 $\pm$ 0.3652 & 1.2406 $\pm$ 0.3828 & 1.2940 $\pm$ 0.4059 \\
Causal CPC \citep{bouchattaoui2024causal} & 1.2283 $\pm$ 0.3749 & 1.3479 $\pm$ 0.5813 & 1.4328 $\pm$ 0.8236 & 1.0898 $\pm$ 0.4290 & 1.1082 $\pm$ 0.3818 \\
CTD-NKO \citep{wang2025enhancing} & 1.0944 $\pm$ 0.3759 & \textbf{0.8241 $\pm$ 0.2657} & \textbf{0.8813 $\pm$ 0.2566} & \underline{1.0029 $\pm$ 0.2229} & 1.1181 $\pm$ 0.2819 \\
IGC-Net \citep{hess2026igc} & \underline{0.9656 $\pm$ 0.1796} & 0.9563 $\pm$ 0.1668 & \underline{0.9159 $\pm$ 0.1606} & \textbf{0.9325 $\pm$ 0.1393} & \textbf{0.9977 $\pm$ 0.1857} \\
\midrule
HiSTOR & \textbf{0.8150 $\pm$ 0.1965} & \underline{0.9094 $\pm$ 0.2042} & 0.9259 $\pm$ 0.2860 & 1.0186 $\pm$ 0.3144 & \underline{1.0264 $\pm$ 0.3026} \\
\bottomrule
\end{tabular}}
\end{table}

\begin{table}[H]
\caption{CVSim Temporal: $\tau$=1. Values are NRMSE, mean $\pm$ sample SD across five randomly selected seeds $\{3,101,42,8,9\}$. Lower is better; the lowest mean is bold and the second-lowest underlined.}
\label{tab:cvsimtemporal_tau1_1}
\centering\footnotesize
\setlength{\tabcolsep}{3pt}
\resizebox{\linewidth}{!}{%
\begin{tabular}{@{}lcccc@{}}
\toprule
Method & d\_hd=200 & d\_hd=500 & d\_hd=1000 & d\_hd=2000 \\
\midrule
MSM \citep{robins2000marginal} & 3.8028 $\pm$ 2.3540 & 3.8076 $\pm$ 2.3731 & 3.8032 $\pm$ 2.3484 & 3.8414 $\pm$ 2.4254 \\
RMSN \citep{lim2018forecasting} & 1.4443 $\pm$ 0.1762 & 1.9093 $\pm$ 0.4264 & 2.3739 $\pm$ 0.7301 & 3.0402 $\pm$ 1.3213 \\
CRN \citep{bica2020estimating} & 3.7383 $\pm$ 2.5857 & 6.9846 $\pm$ 7.1065 & 5.6170 $\pm$ 3.8932 & 7.5510 $\pm$ 5.7171 \\
CT \citep{melnychuk2022causal} & 1.4020 $\pm$ 0.3368 & 1.2786 $\pm$ 0.4567 & 1.3430 $\pm$ 0.4789 & 1.3580 $\pm$ 0.4511 \\
Causal CPC \citep{bouchattaoui2024causal} & 1.0631 $\pm$ 0.3207 & 1.4190 $\pm$ 0.7191 & 1.4839 $\pm$ 1.2221 & 1.2567 $\pm$ 0.5121 \\
CTD-NKO \citep{wang2025enhancing} & 1.4214 $\pm$ 0.3317 & 4.4267 $\pm$ 2.1838 & 4.4615 $\pm$ 2.2147 & 4.4542 $\pm$ 2.2475 \\
IGC-Net \citep{hess2026igc} & \underline{0.9004 $\pm$ 0.1810} & \underline{1.0281 $\pm$ 0.2780} & \underline{1.0373 $\pm$ 0.3260} & \underline{1.0841 $\pm$ 0.3065} \\
\midrule
HiSTOR & \textbf{0.8563 $\pm$ 0.2322} & \textbf{0.9072 $\pm$ 0.2904} & \textbf{0.9507 $\pm$ 0.2519} & \textbf{0.9735 $\pm$ 0.2304} \\
\bottomrule
\end{tabular}}
\end{table}

\begin{table}[H]
\caption{CVSim Temporal: $\tau$=2. Values are NRMSE, mean $\pm$ sample SD across five randomly selected seeds $\{3,101,42,8,9\}$. Lower is better; the lowest mean is bold and the second-lowest underlined.}
\label{tab:cvsimtemporal_tau2_0}
\centering\footnotesize
\setlength{\tabcolsep}{3pt}
\resizebox{\linewidth}{!}{%
\begin{tabular}{@{}lccccc@{}}
\toprule
Method & d\_hd=0 & d\_hd=10 & d\_hd=20 & d\_hd=50 & d\_hd=100 \\
\midrule
MSM \citep{robins2000marginal} & 3.8421 $\pm$ 2.3669 & 3.8411 $\pm$ 2.3713 & 3.8418 $\pm$ 2.3708 & 3.8452 $\pm$ 2.3705 & 3.8503 $\pm$ 2.3724 \\
RMSN \citep{lim2018forecasting} & 1.2856 $\pm$ 0.3597 & 1.4236 $\pm$ 0.4497 & 1.3069 $\pm$ 0.3973 & 1.4910 $\pm$ 0.4588 & 1.5823 $\pm$ 0.4680 \\
CRN \citep{bica2020estimating} & 8.7599 $\pm$ 7.7889 & 9.1807 $\pm$ 8.6508 & 9.1705 $\pm$ 8.6337 & 9.0367 $\pm$ 8.3286 & 6.8978 $\pm$ 5.0452 \\
CT \citep{melnychuk2022causal} & 1.8273 $\pm$ 0.5952 & 1.6233 $\pm$ 0.4889 & 1.5592 $\pm$ 0.3426 & 1.5558 $\pm$ 0.4011 & 1.6731 $\pm$ 0.4366 \\
Causal CPC \citep{bouchattaoui2024causal} & 1.3387 $\pm$ 0.3295 & 1.5632 $\pm$ 0.6230 & 1.5250 $\pm$ 0.6959 & 1.1963 $\pm$ 0.3568 & \underline{1.1807 $\pm$ 0.3038} \\
CTD-NKO \citep{wang2025enhancing} & \underline{1.2266 $\pm$ 0.2930} & \underline{1.0614 $\pm$ 0.3058} & \underline{1.0858 $\pm$ 0.3369} & \underline{1.1267 $\pm$ 0.2747} & 1.2568 $\pm$ 0.3553 \\
IGC-Net \citep{hess2026igc} & 1.4822 $\pm$ 0.6285 & 1.5422 $\pm$ 0.5603 & 1.4291 $\pm$ 0.6542 & 1.5371 $\pm$ 0.7930 & 1.4539 $\pm$ 0.7841 \\
\midrule
HiSTOR & \textbf{0.8958 $\pm$ 0.3056} & \textbf{0.9857 $\pm$ 0.3042} & \textbf{0.9744 $\pm$ 0.3470} & \textbf{1.0792 $\pm$ 0.3631} & \textbf{1.1120 $\pm$ 0.3374} \\
\bottomrule
\end{tabular}}
\end{table}

\begin{table}[H]
\caption{CVSim Temporal: $\tau$=2. Values are NRMSE, mean $\pm$ sample SD across five randomly selected seeds $\{3,101,42,8,9\}$. Lower is better; the lowest mean is bold and the second-lowest underlined.}
\label{tab:cvsimtemporal_tau2_1}
\centering\footnotesize
\setlength{\tabcolsep}{3pt}
\resizebox{\linewidth}{!}{%
\begin{tabular}{@{}lcccc@{}}
\toprule
Method & d\_hd=200 & d\_hd=500 & d\_hd=1000 & d\_hd=2000 \\
\midrule
MSM \citep{robins2000marginal} & 3.8427 $\pm$ 2.3760 & 3.8495 $\pm$ 2.3678 & 3.8605 $\pm$ 2.3732 & 3.9045 $\pm$ 2.4321 \\
RMSN \citep{lim2018forecasting} & 1.4734 $\pm$ 0.4912 & 1.5514 $\pm$ 0.5455 & 1.8450 $\pm$ 0.5639 & 2.4118 $\pm$ 0.7545 \\
CRN \citep{bica2020estimating} & 8.4191 $\pm$ 8.0628 & 8.7964 $\pm$ 8.6260 & 8.1064 $\pm$ 7.0506 & 8.6584 $\pm$ 6.4498 \\
CT \citep{melnychuk2022causal} & 1.8379 $\pm$ 0.3928 & 1.6826 $\pm$ 0.5708 & 1.7027 $\pm$ 0.5107 & 1.5954 $\pm$ 0.3979 \\
Causal CPC \citep{bouchattaoui2024causal} & \underline{1.1090 $\pm$ 0.2561} & \underline{1.5502 $\pm$ 0.6138} & \underline{1.4870 $\pm$ 1.0508} & \underline{1.2529 $\pm$ 0.4568} \\
CTD-NKO \citep{wang2025enhancing} & 1.4866 $\pm$ 0.4620 & 4.4460 $\pm$ 2.3377 & 4.4858 $\pm$ 2.3760 & 4.4887 $\pm$ 2.3739 \\
IGC-Net \citep{hess2026igc} & 1.4547 $\pm$ 0.6507 & 1.5586 $\pm$ 0.8958 & 1.5487 $\pm$ 0.7415 & 1.5305 $\pm$ 0.7174 \\
\midrule
HiSTOR & \textbf{0.9240 $\pm$ 0.2942} & \textbf{1.0344 $\pm$ 0.2680} & \textbf{1.0389 $\pm$ 0.2735} & \textbf{1.0408 $\pm$ 0.3067} \\
\bottomrule
\end{tabular}}
\end{table}

\begin{table}[H]
\caption{CVSim Temporal: $\tau$=3. Values are NRMSE, mean $\pm$ sample SD across five randomly selected seeds $\{3,101,42,8,9\}$. Lower is better; the lowest mean is bold and the second-lowest underlined.}
\label{tab:cvsimtemporal_tau3_0}
\centering\footnotesize
\setlength{\tabcolsep}{3pt}
\resizebox{\linewidth}{!}{%
\begin{tabular}{@{}lccccc@{}}
\toprule
Method & d\_hd=0 & d\_hd=10 & d\_hd=20 & d\_hd=50 & d\_hd=100 \\
\midrule
MSM \citep{robins2000marginal} & 3.8858 $\pm$ 2.3833 & 3.8842 $\pm$ 2.3868 & 3.8865 $\pm$ 2.3897 & 3.8888 $\pm$ 2.3902 & 3.8908 $\pm$ 2.3856 \\
RMSN \citep{lim2018forecasting} & 1.3522 $\pm$ 0.3108 & 1.4307 $\pm$ 0.4132 & 1.3081 $\pm$ 0.3365 & 1.5213 $\pm$ 0.4338 & 1.6321 $\pm$ 0.4570 \\
CRN \citep{bica2020estimating} & 9.2023 $\pm$ 7.9992 & 9.1453 $\pm$ 8.1649 & 9.5191 $\pm$ 8.6350 & 9.8500 $\pm$ 8.8859 & 7.6501 $\pm$ 5.3291 \\
CT \citep{melnychuk2022causal} & 2.1180 $\pm$ 0.5476 & 1.9765 $\pm$ 0.6233 & 1.7822 $\pm$ 0.3857 & 1.8253 $\pm$ 0.5196 & 1.9645 $\pm$ 0.4525 \\
Causal CPC \citep{bouchattaoui2024causal} & 1.3751 $\pm$ 0.3206 & 1.8174 $\pm$ 0.8584 & 1.6337 $\pm$ 0.5918 & \underline{1.2919 $\pm$ 0.2888} & \underline{1.2509 $\pm$ 0.2657} \\
CTD-NKO \citep{wang2025enhancing} & \underline{1.3109 $\pm$ 0.2354} & \underline{1.1867 $\pm$ 0.3959} & \underline{1.1934 $\pm$ 0.4943} & 1.2935 $\pm$ 0.3924 & 1.3953 $\pm$ 0.4537 \\
IGC-Net \citep{hess2026igc} & 2.2265 $\pm$ 1.2850 & 2.1121 $\pm$ 0.7847 & 2.3933 $\pm$ 1.9473 & 1.9576 $\pm$ 0.8403 & 2.0291 $\pm$ 0.9745 \\
\midrule
HiSTOR & \textbf{0.9656 $\pm$ 0.3078} & \textbf{1.0671 $\pm$ 0.3362} & \textbf{1.0473 $\pm$ 0.3628} & \textbf{1.1418 $\pm$ 0.4107} & \textbf{1.2051 $\pm$ 0.3534} \\
\bottomrule
\end{tabular}}
\end{table}

\begin{table}[H]
\caption{CVSim Temporal: $\tau$=3. Values are NRMSE, mean $\pm$ sample SD across five randomly selected seeds $\{3,101,42,8,9\}$. Lower is better; the lowest mean is bold and the second-lowest underlined.}
\label{tab:cvsimtemporal_tau3_1}
\centering\footnotesize
\setlength{\tabcolsep}{3pt}
\resizebox{\linewidth}{!}{%
\begin{tabular}{@{}lcccc@{}}
\toprule
Method & d\_hd=200 & d\_hd=500 & d\_hd=1000 & d\_hd=2000 \\
\midrule
MSM \citep{robins2000marginal} & 3.8903 $\pm$ 2.3920 & 3.8871 $\pm$ 2.3959 & 3.9055 $\pm$ 2.3782 & 3.9438 $\pm$ 2.4199 \\
RMSN \citep{lim2018forecasting} & 1.4991 $\pm$ 0.4947 & \underline{1.5643 $\pm$ 0.4692} & 1.7588 $\pm$ 0.5120 & 2.3022 $\pm$ 0.7176 \\
CRN \citep{bica2020estimating} & 9.6523 $\pm$ 9.7523 & 9.3451 $\pm$ 8.5753 & 8.7543 $\pm$ 6.9779 & 9.1869 $\pm$ 6.4209 \\
CT \citep{melnychuk2022causal} & 2.1612 $\pm$ 0.7570 & 2.0144 $\pm$ 0.5815 & 2.0017 $\pm$ 0.5027 & 1.8937 $\pm$ 0.3048 \\
Causal CPC \citep{bouchattaoui2024causal} & \underline{1.1478 $\pm$ 0.2038} & 1.7079 $\pm$ 0.4869 & \underline{1.4826 $\pm$ 0.8399} & \underline{1.3182 $\pm$ 0.4269} \\
CTD-NKO \citep{wang2025enhancing} & 1.6351 $\pm$ 0.6281 & 4.6420 $\pm$ 2.5112 & 4.6873 $\pm$ 2.5473 & 4.7475 $\pm$ 2.5618 \\
IGC-Net \citep{hess2026igc} & 2.0111 $\pm$ 1.0456 & 2.0744 $\pm$ 1.3022 & 2.2938 $\pm$ 1.1763 & 2.3153 $\pm$ 1.3271 \\
\midrule
HiSTOR & \textbf{1.0217 $\pm$ 0.3082} & \textbf{1.0761 $\pm$ 0.3286} & \textbf{1.0827 $\pm$ 0.3490} & \textbf{1.1502 $\pm$ 0.3614} \\
\bottomrule
\end{tabular}}
\end{table}

\begin{table}[H]
\caption{CVSim Temporal: $\tau$=4. Values are NRMSE, mean $\pm$ sample SD across five randomly selected seeds $\{3,101,42,8,9\}$. Lower is better; the lowest mean is bold and the second-lowest underlined.}
\label{tab:cvsimtemporal_tau4_0}
\centering\footnotesize
\setlength{\tabcolsep}{3pt}
\resizebox{\linewidth}{!}{%
\begin{tabular}{@{}lccccc@{}}
\toprule
Method & d\_hd=0 & d\_hd=10 & d\_hd=20 & d\_hd=50 & d\_hd=100 \\
\midrule
MSM \citep{robins2000marginal} & 3.9651 $\pm$ 2.4204 & 3.9649 $\pm$ 2.4240 & 3.9650 $\pm$ 2.4230 & 3.9674 $\pm$ 2.4246 & 3.9681 $\pm$ 2.4180 \\
RMSN \citep{lim2018forecasting} & 1.5109 $\pm$ 0.3055 & 1.5847 $\pm$ 0.4538 & 1.4825 $\pm$ 0.4160 & 1.6825 $\pm$ 0.4671 & 1.7795 $\pm$ 0.5406 \\
CRN \citep{bica2020estimating} & 9.9220 $\pm$ 8.2963 & 9.7230 $\pm$ 8.2491 & 10.0333 $\pm$ 8.3287 & 10.4043 $\pm$ 8.5707 & 8.4971 $\pm$ 5.6567 \\
CT \citep{melnychuk2022causal} & 2.4662 $\pm$ 0.5405 & 2.3808 $\pm$ 0.8863 & 2.0439 $\pm$ 0.5551 & 2.1270 $\pm$ 0.6799 & 2.3368 $\pm$ 0.5892 \\
Causal CPC \citep{bouchattaoui2024causal} & \underline{1.4283 $\pm$ 0.3899} & 2.1444 $\pm$ 1.2535 & 1.7726 $\pm$ 0.5054 & \underline{1.4170 $\pm$ 0.2990} & \underline{1.3949 $\pm$ 0.3476} \\
CTD-NKO \citep{wang2025enhancing} & 1.4702 $\pm$ 0.3164 & \underline{1.4211 $\pm$ 0.5684} & \underline{1.4438 $\pm$ 0.7029} & 1.5732 $\pm$ 0.6627 & 1.6515 $\pm$ 0.7137 \\
IGC-Net \citep{hess2026igc} & 3.1213 $\pm$ 1.9334 & 3.5015 $\pm$ 1.5944 & 3.0241 $\pm$ 0.8232 & 2.9377 $\pm$ 2.0382 & 3.2641 $\pm$ 1.3258 \\
\midrule
HiSTOR & \textbf{1.0634 $\pm$ 0.3309} & \textbf{1.1589 $\pm$ 0.3945} & \textbf{1.1744 $\pm$ 0.3259} & \textbf{1.2187 $\pm$ 0.4948} & \textbf{1.3402 $\pm$ 0.3351} \\
\bottomrule
\end{tabular}}
\end{table}

\begin{table}[H]
\caption{CVSim Temporal: $\tau$=4. Values are NRMSE, mean $\pm$ sample SD across five randomly selected seeds $\{3,101,42,8,9\}$. Lower is better; the lowest mean is bold and the second-lowest underlined.}
\label{tab:cvsimtemporal_tau4_1}
\centering\footnotesize
\setlength{\tabcolsep}{3pt}
\resizebox{\linewidth}{!}{%
\begin{tabular}{@{}lcccc@{}}
\toprule
Method & d\_hd=200 & d\_hd=500 & d\_hd=1000 & d\_hd=2000 \\
\midrule
MSM \citep{robins2000marginal} & 3.9756 $\pm$ 2.4257 & 3.9841 $\pm$ 2.4437 & 3.9866 $\pm$ 2.4153 & 4.0472 $\pm$ 2.4655 \\
RMSN \citep{lim2018forecasting} & 1.7232 $\pm$ 0.5150 & \underline{1.7660 $\pm$ 0.4591} & 1.9739 $\pm$ 0.5195 & 2.5343 $\pm$ 0.6988 \\
CRN \citep{bica2020estimating} & 10.2041 $\pm$ 9.6734 & 10.1112 $\pm$ 8.5561 & 9.5223 $\pm$ 6.9038 & 9.7942 $\pm$ 6.4114 \\
CT \citep{melnychuk2022causal} & 2.5202 $\pm$ 1.1858 & 2.3838 $\pm$ 0.7752 & 2.3744 $\pm$ 0.6647 & 2.2099 $\pm$ 0.6148 \\
Causal CPC \citep{bouchattaoui2024causal} & \underline{1.2229 $\pm$ 0.2336} & 1.9155 $\pm$ 0.4871 & \underline{1.4785 $\pm$ 0.6079} & \underline{1.4279 $\pm$ 0.4289} \\
CTD-NKO \citep{wang2025enhancing} & 1.8271 $\pm$ 0.8175 & 4.9769 $\pm$ 2.6887 & 5.0271 $\pm$ 2.7036 & 5.1400 $\pm$ 2.7537 \\
IGC-Net \citep{hess2026igc} & 2.9464 $\pm$ 1.6850 & 3.4248 $\pm$ 1.4696 & 3.0458 $\pm$ 1.6178 & 3.1845 $\pm$ 1.3517 \\
\midrule
HiSTOR & \textbf{1.1254 $\pm$ 0.3353} & \textbf{1.1601 $\pm$ 0.3350} & \textbf{1.1320 $\pm$ 0.3930} & \textbf{1.2510 $\pm$ 0.4418} \\
\bottomrule
\end{tabular}}
\end{table}

\begin{table}[H]
\caption{CVSim Temporal: $\tau$=5. Values are NRMSE, mean $\pm$ sample SD across five randomly selected seeds $\{3,101,42,8,9\}$. Lower is better; the lowest mean is bold and the second-lowest underlined.}
\label{tab:cvsimtemporal_tau5_0}
\centering\footnotesize
\setlength{\tabcolsep}{3pt}
\resizebox{\linewidth}{!}{%
\begin{tabular}{@{}lccccc@{}}
\toprule
Method & d\_hd=0 & d\_hd=10 & d\_hd=20 & d\_hd=50 & d\_hd=100 \\
\midrule
MSM \citep{robins2000marginal} & 4.0077 $\pm$ 2.4594 & 4.0064 $\pm$ 2.4631 & 4.0066 $\pm$ 2.4641 & 4.0085 $\pm$ 2.4621 & 4.0131 $\pm$ 2.4587 \\
RMSN \citep{lim2018forecasting} & 1.7393 $\pm$ 0.2836 & 1.8216 $\pm$ 0.5468 & 1.7433 $\pm$ 0.4956 & 1.9173 $\pm$ 0.5048 & 2.0049 $\pm$ 0.6214 \\
CRN \citep{bica2020estimating} & 10.6612 $\pm$ 8.5815 & 10.3802 $\pm$ 8.4132 & 10.6169 $\pm$ 8.1565 & 10.9928 $\pm$ 8.2859 & 9.2986 $\pm$ 5.9158 \\
CT \citep{melnychuk2022causal} & 2.8132 $\pm$ 0.6403 & 2.7692 $\pm$ 1.1529 & 2.3687 $\pm$ 0.8105 & 2.4770 $\pm$ 0.9247 & 2.6887 $\pm$ 0.8144 \\
Causal CPC \citep{bouchattaoui2024causal} & \underline{1.4933 $\pm$ 0.4758} & 2.4711 $\pm$ 1.7521 & 1.9015 $\pm$ 0.4874 & \underline{1.5247 $\pm$ 0.3522} & \underline{1.5489 $\pm$ 0.4284} \\
CTD-NKO \citep{wang2025enhancing} & 1.6191 $\pm$ 0.4061 & \underline{1.7119 $\pm$ 0.5881} & \underline{1.6924 $\pm$ 0.8664} & 1.8892 $\pm$ 0.9017 & 1.9486 $\pm$ 0.9485 \\
IGC-Net \citep{hess2026igc} & 4.0302 $\pm$ 2.1144 & 4.3361 $\pm$ 1.3926 & 4.1755 $\pm$ 1.5627 & 3.8849 $\pm$ 1.5206 & 3.9320 $\pm$ 2.8398 \\
\midrule
HiSTOR & \textbf{1.1757 $\pm$ 0.3035} & \textbf{1.2798 $\pm$ 0.3720} & \textbf{1.3284 $\pm$ 0.3094} & \textbf{1.3363 $\pm$ 0.4991} & \textbf{1.4830 $\pm$ 0.3621} \\
\bottomrule
\end{tabular}}
\end{table}

\begin{table}[H]
\caption{CVSim Temporal: $\tau$=5. Values are NRMSE, mean $\pm$ sample SD across five randomly selected seeds $\{3,101,42,8,9\}$. Lower is better; the lowest mean is bold and the second-lowest underlined.}
\label{tab:cvsimtemporal_tau5_1}
\centering\footnotesize
\setlength{\tabcolsep}{3pt}
\resizebox{\linewidth}{!}{%
\begin{tabular}{@{}lcccc@{}}
\toprule
Method & d\_hd=200 & d\_hd=500 & d\_hd=1000 & d\_hd=2000 \\
\midrule
MSM \citep{robins2000marginal} & 4.0137 $\pm$ 2.4674 & 4.0157 $\pm$ 2.4475 & 4.0362 $\pm$ 2.4616 & 4.0822 $\pm$ 2.4862 \\
RMSN \citep{lim2018forecasting} & 2.0413 $\pm$ 0.5540 & \underline{2.0463 $\pm$ 0.4792} & 2.2591 $\pm$ 0.5369 & 2.8261 $\pm$ 0.6786 \\
CRN \citep{bica2020estimating} & 10.8501 $\pm$ 9.6670 & 10.9141 $\pm$ 8.5901 & 10.3101 $\pm$ 6.9036 & 10.4517 $\pm$ 6.4611 \\
CT \citep{melnychuk2022causal} & 2.8403 $\pm$ 1.5287 & 2.7089 $\pm$ 1.1283 & 2.7040 $\pm$ 0.8821 & 2.5102 $\pm$ 0.9041 \\
Causal CPC \citep{bouchattaoui2024causal} & \underline{1.3179 $\pm$ 0.2902} & 2.1217 $\pm$ 0.6315 & \underline{1.4528 $\pm$ 0.3846} & \underline{1.5564 $\pm$ 0.4550} \\
CTD-NKO \citep{wang2025enhancing} & 2.0755 $\pm$ 0.9644 & 5.3428 $\pm$ 2.7960 & 5.4029 $\pm$ 2.7734 & 5.5504 $\pm$ 2.8749 \\
IGC-Net \citep{hess2026igc} & 3.8640 $\pm$ 1.6429 & 3.7674 $\pm$ 1.4610 & 3.6166 $\pm$ 1.8322 & 3.9451 $\pm$ 1.9916 \\
\midrule
HiSTOR & \textbf{1.2826 $\pm$ 0.3161} & \textbf{1.2151 $\pm$ 0.3834} & \textbf{1.2371 $\pm$ 0.3929} & \textbf{1.4214 $\pm$ 0.4636} \\
\bottomrule
\end{tabular}}
\end{table}

\begin{figure}[!htbp]
\centering
\includegraphics[width=\linewidth]{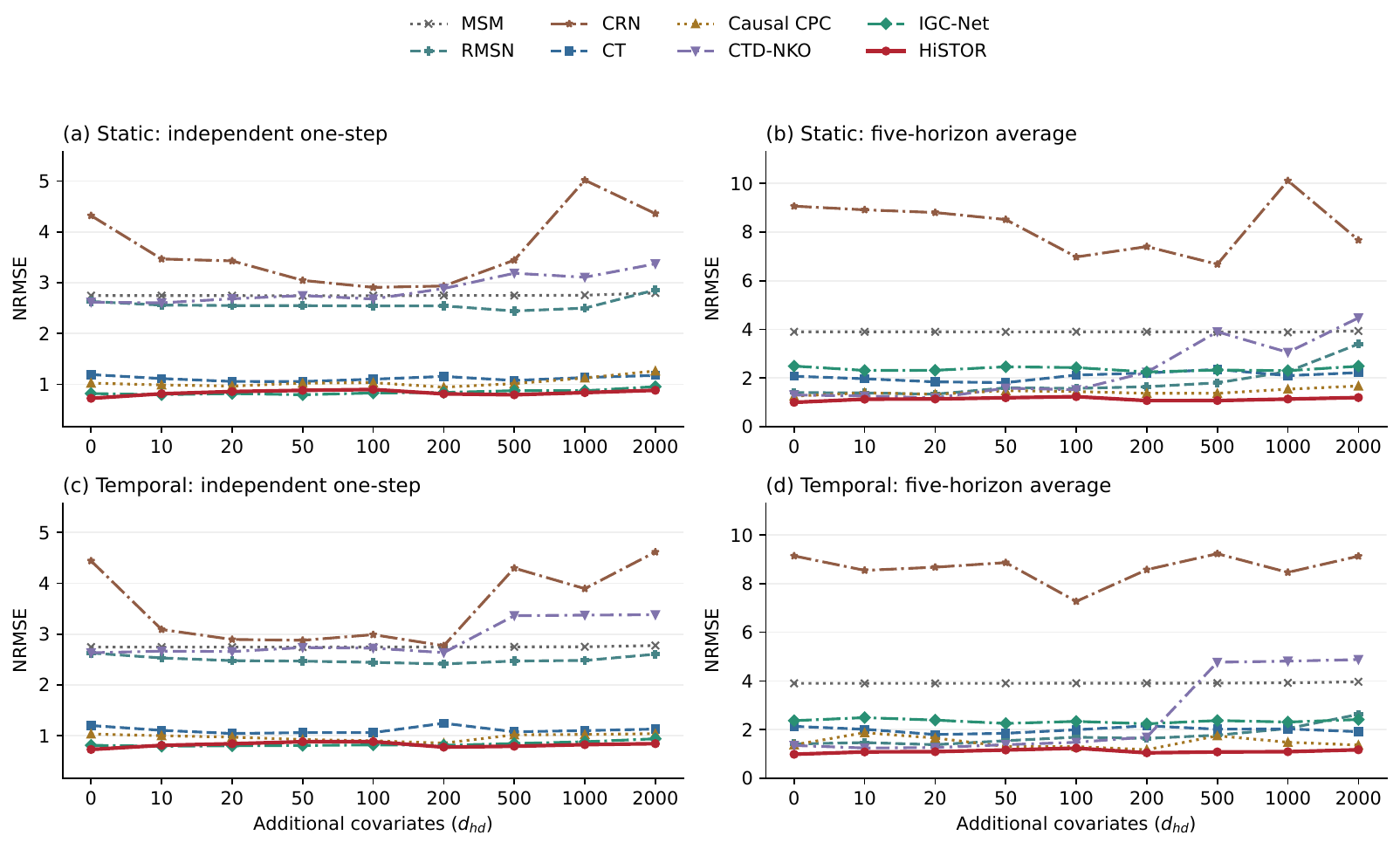}
\caption{All eight methods across the complete CVSim dimension grid. Lines show five-seed mean NRMSE.}
\label{fig:cvsim_all_methods}
\end{figure}
\clearpage

\section{MIMIC-III Semi-Synthetic Evaluation}\label{app:mimic}

\paragraph{Data and protocol.}
To complement the controlled simulation studies, we evaluate HiSTOR on
MIMIC-III ICU covariates processed using the MIMIC-Extract pipeline
\citep{johnson2016mimic,wang2020mimic}, followed by the IGC-Net
semi-synthetic outcome and treatment generator \citep{hess2026igc}.
This is a semi-synthetic counterfactual benchmark, not a real-outcome
clinical prediction experiment. The cohort contains 1,000 patients with a
patient-level split of 600/200/200 for training, validation, and testing.
Each raw trajectory contains 30--50 time points; the upstream indexing shift
makes the effective model sequence one time point shorter. The benchmark has
three binary treatment variables and two synthetic outcomes. We use seeds
$\{3,101,42,8,9\}$. At $\tau=1$, we evaluate one-step counterfactual
prediction. For $\tau=2,\ldots,6$, we evaluate multi-step prediction under
the six-step treatment plan $[0,0,1]$ repeated at every step.

\paragraph{Comparison protocol.}
We include CT \citep{melnychuk2022causal} and IGC-Net
\citep{hess2026igc} as two strong prior methods with released support for
this benchmark. CT and IGC-Net use their released official implementations
and configurations, whereas HiSTOR uses its model-specific configuration.
All three models share the same cached cohort, patient split, seeds, forecast
boundary, and evaluation implementation. The table reports arithmetic means
over the five seeds. 

\begin{table}[H]
\caption{Values are mean scaled RMSE over seeds $\{3,101,42,8,9\}$; lower is better. Standard deviations are omitted from this compact table.}
\label{tab:mimic_semisynthetic}
\centering
\small
\setlength{\tabcolsep}{5pt}
\begin{tabular}{@{}lcccccc@{}}
\toprule
Method & $\tau=1$ & $\tau=2$ & $\tau=3$ & $\tau=4$ & $\tau=5$ & $\tau=6$ \\
\midrule
CT \citep{melnychuk2022causal} & \textbf{0.1269} & 0.3157 & 0.4134 & 0.4717 & 0.5256 & 0.5824 \\
IGC-Net \citep{hess2026igc} & 0.1349 & 0.3537 & 0.4430 & 0.5299 & 0.5599 & 0.6519 \\
HiSTOR & 0.1701 & \textbf{0.3148} & \textbf{0.3951} & \textbf{0.4709} & \textbf{0.4949} & \textbf{0.5446} \\
\bottomrule
\end{tabular}
\end{table}
CT achieves the lowest one-step error at $\tau=1$, whereas HiSTOR achieves
the lowest error among the three evaluated methods at every multi-step
horizon $\tau=2,\ldots,6$. Relative to CT, the strongest competing method
at these horizons, HiSTOR reduces error by approximately $0.3\%$, $4.4\%$,
$0.2\%$, $5.8\%$, and $6.5\%$, respectively. These results provide
complementary evidence for HiSTOR's multi-step prediction performance on
clinically grounded covariates under the semi-synthetic fixed-treatment
protocol.

\end{document}